\documentclass[conference]{IEEEtran}
\usepackage{times}
\usepackage{amsmath}

\usepackage[numbers,sort&compress]{natbib}
\usepackage{multicol}
\usepackage[bookmarks=true]{hyperref}

\usepackage{listings}
\usepackage{xcolor}
\usepackage{graphicx}
\usepackage{algorithm}
\usepackage{algpseudocode}
\usepackage{booktabs}

\usepackage{array}
\usepackage{tcolorbox}
\usepackage{titletoc}

\newcommand{\RQI}[0]{\textit{\textbf{RQ1}}}
\newcommand{\RQII}[0]{\textit{\textbf{RQ2}}}

\makeatletter
\let\labelindent\relax
\makeatother
\usepackage{enumitem}
\usepackage{pdfpages}
\usepackage{hyperref}

\usepackage{graphicx}
\lstdefinestyle{pydata}{
  basicstyle=\ttfamily\footnotesize,
  columns=fixed,
  keepspaces=true,
  showstringspaces=false,
  frame=none,
  numbers=none,
  breaklines=true,
  backgroundcolor=\color{gray!10},
  xleftmargin=2pt,
  xrightmargin=2pt,
  aboveskip=4pt,
  belowskip=4pt,
  emph={id,type,description,content,condition,verification_condition},
  emphstyle=\bfseries,
  emph={[2]add_fact, add_effect},
  emphstyle={[2]\itshape},
  emph={[3]None},
  emphstyle={[3]\color{gray}}
}

\hypersetup{
    colorlinks=true,
    linkcolor=[RGB]{70,130,210},
    citecolor=[RGB]{50, 205, 50},
    urlcolor=[RGB]{70,130,210},
    pdfauthor={Anxing Xiao, Hanbo Zhang, Tianrun Hu, David Hsu},
    pdftitle={Hypothesis-driven Model Expansion under Uncertainty for Open-World Robot Planning},
    pdfsubject={Robotics, Planning},
    pdfkeywords={robotics;mobile manipulation;planning;active perception}
}

\begin{document}



\title{\fontsize{23.2}{29}\selectfont Hypothesis-driven Model Expansion under Uncertainty \\for Open-World Robot Planning}

\author{
\authorblockN{
Anxing Xiao\textsuperscript{1},
Hanbo Zhang\textsuperscript{1,2},
Tianrun Hu\textsuperscript{1,2},
David Hsu\textsuperscript{1,2}
}
\authorblockA{
\textsuperscript{1}School of Computing, National University of Singapore \\
\textsuperscript{2}Smart Systems Institute, National University of Singapore 
}
\vspace{-3.2em}
}

\maketitle

\begin{abstract}
We consider an open-world planning setting in which service robots operate in unknown environments with incomplete knowledge of objects and actions.
Traditional closed-world approaches with pre-programmed knowledge bases fail when robots encounter unexpected situations and tasks, posing a fundamental challenge for autonomous knowledge expansion in human environments.
In this work, we propose an open-world planning framework that enables robots to automatically generate, verify, and update hypotheses about their abstract world models. 
Our key insight is to explicitly maintain uncertainty-aware knowledge expansion and integrate hypothesis verification into goal-reaching planning. 
Our framework, Hypothesis-driven Uncertainty-aware Model Expansion (HUME), leverages foundation models to generate initial hypotheses over states and transitions, and applies automated planning to produce action sequences that jointly address hypothesis verification and task execution.
Through iterative execution and refinement, the robot expands its knowledge by incorporating verification feedback from the foundation models when hypotheses prove incorrect.
Extensive experiments in simulated and real-world environments demonstrate that our framework enables autonomous knowledge expansion and effective operation in open-world settings.
These results indicate that integrating uncertainty-aware model expansion from robot foundation models with planning advances the practical deployment of household service robots. Project website: \href{https://open-world-planning.github.io}{open-world-planning.github.io}


\end{abstract}

\IEEEpeerreviewmaketitle

\section{Introduction} 
\begin{figure}[t]
 \vspace{0.1cm}
  \centering
  \includegraphics[width=0.97\linewidth]{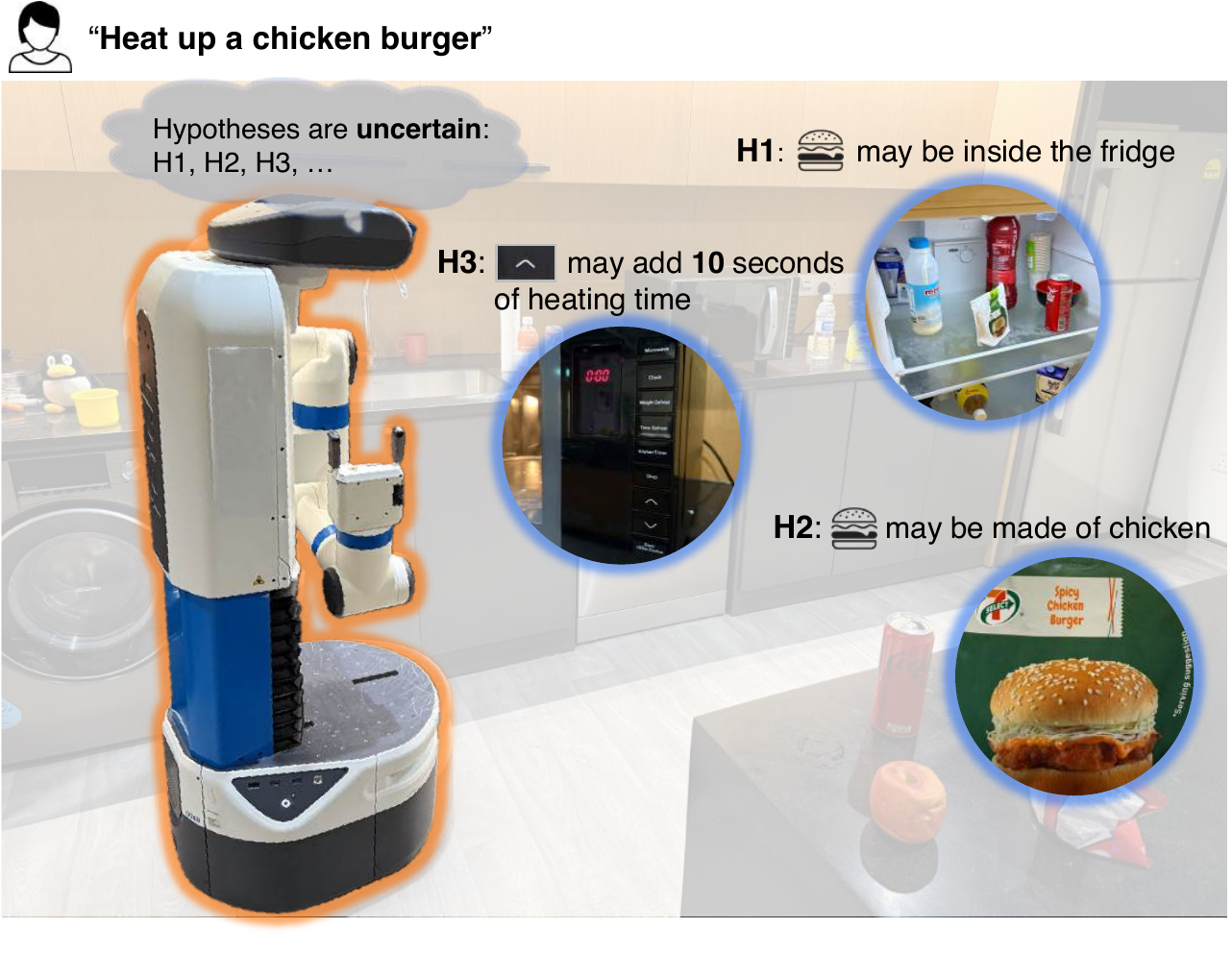}
  \vspace{-0.35cm}
  \caption{Illustration of a service robot operating in an open-world scenario. To fulfill tasks such as \textit{``heat up a chicken burger''}, the robot must expand its abstract world model and account for uncertainty during planning.
  }
  \label{fig:teaser}
\vspace{-0.7cm}
\end{figure}

Home service robots operate under substantial uncertainty in real-world household environments, where objects may be hidden, their attributes unknown, and the effects of actions underspecified.
Consider a daily task such as \textit{``serve a heated chicken burger''}, as shown in Fig.~\ref{fig:teaser}.
To accomplish this task, a robot must infer the burger's location, determine whether it is made of chicken, and, in some cases, reason about appliance-specific action effects, such as how much the heating time increases when pressing the `+' button on a microwave. 
These challenges make household environments inherently open-world. 
Classical automated planning relies on predefined state representations and transition models \cite{fox2003pddl2,brewka2011answer, garrett2020pddlstream}.
While effective in well-specified domains, this reliance on manually designed knowledge and closed-world assumptions limits generalization to open-world settings with underspecified knowledge. In contrast, Large Language Models (LLMs) exhibit remarkable reasoning and generalization to predict plausible plans without explicitly modeling states or transition dynamics \citep{huang2022language, brohan2023can, huang2023inner, liang2023code, driess2023palm}. Yet, in open-world settings with partial information, these properties can blur the boundary between reasoning grounded in known facts and reasoning based on hallucinated knowledge. This motivates a central question of this work: \textit{how can we combine structured reasoning with the unstructured open-world knowledge of foundation models?}

To bridge this gap, recent research has explored using foundation models to construct or fix symbolic representations of the environment \cite{liu2024learning, yeunidomain, ding2023integrating, chen2024language, liang2025visualpredicator, yang2025skillwrapper, ye2026uniplan, athalye2026pixels}. 
These approaches allow models to generate planning domains \cite{liu2024learning, athalye2026pixels, yeunidomain, liang2025visualpredicator, yang2025skillwrapper, ye2026uniplan}, infer missing preconditions or effects \cite{ding2023integrating, chen2024language}, and construct probabilistic world models \cite{curtis2025llm}. While promising, such methods largely rely on passive inference: once a model is generated, it is typically assumed to be correct or only updated indirectly through execution failures. However, ignoring uncertainty in expanded world knowledge can cause silent failures, as plans may still be executed even when object attributes or action effects are incorrectly predicted.

This limitation motivates our core insight: \textit{in open-world planning, model expansion itself must be treated as an uncertain and actively verifiable process}. From a decision-theoretic perspective, this corresponds to a Bayes-adaptive view of planning \cite{duff2002optimal}, in which the agent explicitly reasons over uncertainty in the underlying world model and selects actions that jointly achieve task goals and reduce model uncertainty.

\begin{figure}[ptb]
  \centering
  \includegraphics[width=\linewidth]{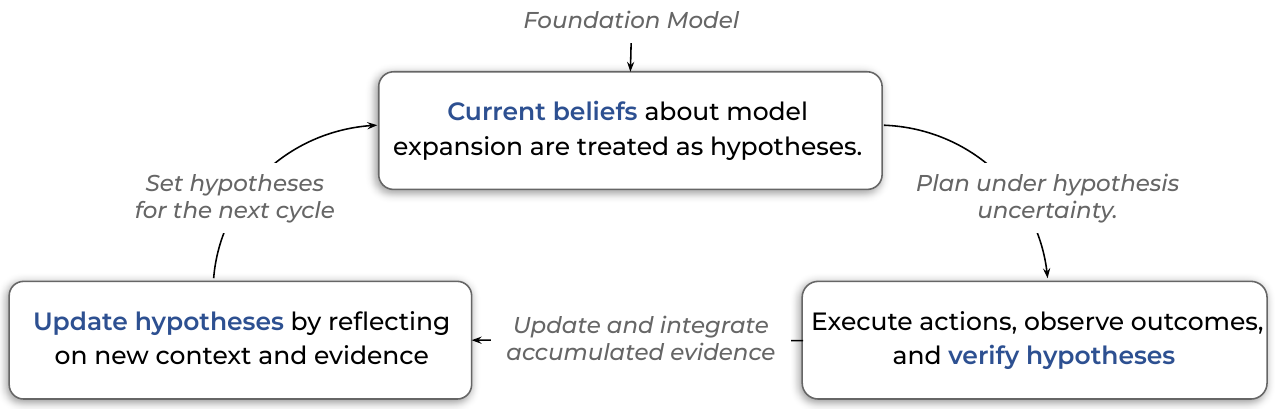}
  \vspace{-0.7cm}
  \caption{High-level idea. 
  It combines foundation-model priors as hypotheses with grounded interaction to iteratively expand its environment model.
  }
  \label{fig:intro_concept}
\vspace{-0.8cm}
\end{figure}

Building on this perspective, we propose \textbf{HUME} (\textbf{H}ypothesis-driven \textbf{U}ncertainty-aware \textbf{M}odel \textbf{E}xpansion), an open-world planning framework that integrates structured symbolic reasoning with the unstructured commonsense priors of language models through explicit, uncertainty-aware world model expansion (Fig. \ref{fig:intro_concept}).
The robot maintains a set of object-centric hypotheses representing unknown or uncertain aspects of the environment, such as object locations, attributes, or the effects of actions applied to them. These hypotheses are generated by foundation models and are treated as uncertain latent variables rather than as facts. Given this formulation, the planning problem is formulated over an augmented representation that includes both the known model and these uncertain hypotheses. 
Through appropriate determinization \cite{yoon2007ff}, we demonstrate that a classical planner can be applied to produce partially grounded plans that explicitly interleave task execution with verification. During the execution, the foundation models also help ground and update these hypotheses based on new sensory observations and historical information. These hypothesis instances thus serve as both imagination and memory, enabling continual active expansion of the world model.
We demonstrate the generality of HUME across diverse open-world tasks, including Block Processing World, mobile manipulation in unknown environments, and appliance operation. Experiments in both simulation and real-world show that our framework enables autonomous knowledge expansion and effective task execution in open-world settings for both classical formal planners and language model planners. Our results highlight the importance of integrating uncertainty-aware model expansion into planning, which we believe is critical for deploying service robots in real open-world environments.

\section{Related Work}
\label{sec:related}

\subsection{AI Planning with Foundation Models}
Recent advances in foundation models have spurred growing interest in using language models for zero-shot decision-making, predicting planning outcomes without explicit forward search.
These approaches leverage commonsense knowledge encoded in LLMs to directly generate structured plans, such as action sequences \citep{huang2022language, brohan2023can, huang2023inner, silver2022pddl, driess2023palm, liu2023reflect, rana2023sayplan}, or executable code \cite{singh2023progprompt, liang2023code, vemprala2023chatgpt, silver2023generalized}. Such plans are often refined through self-reflection using passive feedback \cite{wang2023describe, song2023llmplanner, shinn2023reflexion, liu2023reflect}, while retrieval-augmented methods further incorporate past experiences \cite{wang2024jarvis}.
Despite these advances, LLM-based planners still struggle with long-horizon coherence, constraint enforcement, and active information acquisition under uncertainty \cite{birr2024autogpt+, curtis2025trust,yang2025guiding,zhao2025seeing}.
To mitigate these limitations, recent work integrates LLMs with classical planning frameworks to combine symbolic structure. In such hybrid systems, LLMs could assist with goal interpretation \cite{xie2023translating, liu2023llm+}, planning tool selection \cite{birr2024autogpt+}, and heuristic guidance \cite{silver2022pddl, zhao2024large, yang2025guiding, kwon2026kinodynamic}. 
More recent approaches integrate the concept of model acquisition \cite{cresswell2011generalised, diehl2021automated} with foundation models to extend or revise planning domains, for example by generating domain descriptions from demonstrations \cite{wong2023learning, liu2024learning, yeunidomain, yang2025skillwrapper, liang2025visualpredicator, athalye2026pixels}, performing program induction \cite{curtis2025llm, zandonati2025rational}, inventing novel spatial predicates \cite{curtis2025trust, kumar2026open}, or inferring missing preconditions and effects \cite{ding2023integrating, chen2024language}.
However, such domain knowledge is often static after generation or updated only through passive external feedback \cite{chen2024language, zhu2025language, curtis2025llm}. 
Building on this line of work, we explore uncertainty-aware world model expansion, where robots plan actions to reach informative states for evaluating expanded knowledge.
From this perspective, our approach casts embodied question answering \cite{das2018embodied, ren2024explore} as a proactive process and integrates it into the full task-planning pipeline, by explicitly encoding verification queries within the planning problem itself.

\subsection{Planning in the Open World}
To relax closed-world assumptions, some approaches incorporate external knowledge bases as priors \cite{jiang2019open, chernova2020situated}, and rely on replanning when outcomes deviate from expectations. 
Without explicitly modeling uncertainty, such planners struggle to reason about silent failures and risks.
Belief-space planning addresses this challenge by maintaining probabilistic beliefs over world states \cite{kaelbling2013integrated, garrett2020online, curtis2024partially}, enabling principled reasoning under partial observability and environmental uncertainty. 
This formulation has been widely and successfully applied to open-world tasks, e.g., object search \cite{aydemir2013active, wandzel2019multi, wong2013manipulation}.
However, constructing such structured belief models typically requires extensive hand-engineering of concepts such as objects, attributes, and action effects, along with their associated priors and belief update mechanisms, which significantly limits scalability in open-world scenarios.
In contrast, LLMs encode rich unstructured priors over world knowledge, enabling implicit reasoning about objects, actions, and outcomes, which is leveraged by \cite{sun2024interactive} to generate information-gathering actions in short-horizon tasks.
To bridge unstructured priors with uncertainty-aware planning, \cite{curtis2025llm} uses LLMs to specify observation models and belief updates over predefined POMDP abstractions, but does not explicitly model uncertainty in the generated knowledge and relies on predefined passive feedback for updates. 
\cite{tangtru} demonstrates the ability of LLMs to generate high-quality particle beliefs over object locations and human goals for online POMDP planning. 
Recent work \cite{zhao2025seeing} combines visual predicate evaluation with uncertainty-aware planners for undefined object attributes, yet relies on manually specified information-gathering actions and does not address object locations or action effects.
Our method addresses the gaps by integrating LLM-derived priors into structured, uncertainty-aware knowledge expansion as hypotheses over factorized object locations, attributes, and action effects, conceptually aligning with Bayes-adaptive planning \cite{duff2002optimal}, in which uncertainty arises from unknown or evolving models of the world.

\section{Problem Formulation}
\begin{figure*}[hptb]
  \centering
  \includegraphics[width=0.98\linewidth]{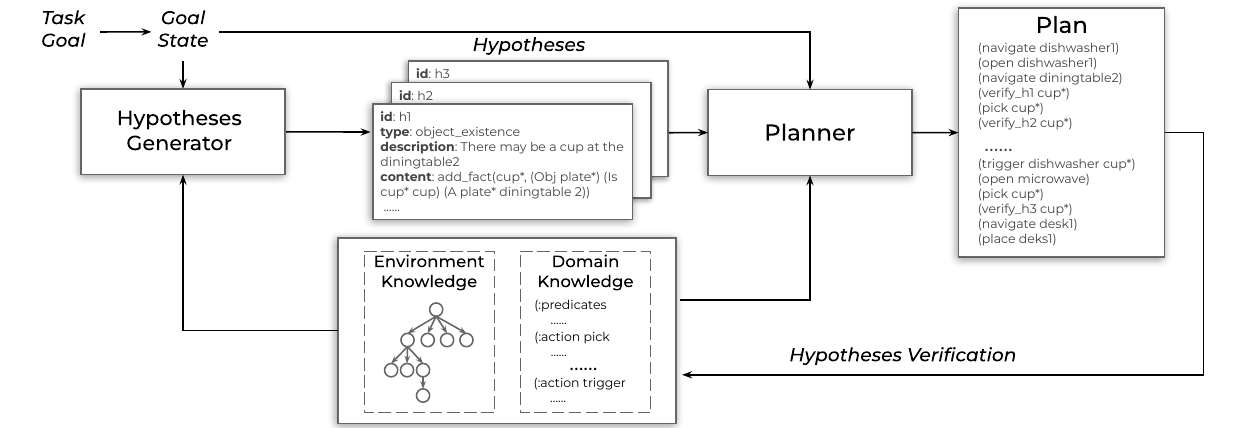}
  \vspace{-0.2cm}
  \caption{Overview of the HUME. The robot iteratively generates hypotheses to expand its model, plans with a model augmented by uncertain hypotheses, and executes actions to verify and update these hypotheses using newly acquired observations.}
  \label{fig:framework}
\vspace{-0.5cm}
\end{figure*}

\subsection{Problem Setting}
We consider planning under an incomplete world model, in which a robot operates in an environment with missing object-centric knowledge. The robot is equipped with an abstract environment model $\mathcal{M} = \langle \mathcal{S}, \mathcal{P}, \mathcal{A}, T \rangle$, where $\mathcal{S}$ denotes the symbolic state space defined by a predicate set $\mathcal{P}$, $\mathcal{A}$ is the action space, and $T : \mathcal{S} \times \mathcal{A} \rightarrow \mathcal{S}$ is a deterministic transition function over known predicates.
The true environment is governed by an unknown ground-truth model $\mathcal{M}_{\text{gt}} = \langle \mathcal{S}_{\text{gt}}, \mathcal{P}_{\text{gt}}, \mathcal{A}, T_{\text{gt}} \rangle$, which may include additional objects, predicates, or transition effects not captured in $\mathcal{M}$. 
As a result, $\mathcal{M}$ provides only a partial abstraction of the real world.
Tasks are specified via natural language instructions $L$, corresponding to an unknown goal set $\mathcal{G} \subseteq \mathcal{S}_{\text{gt}}$. The robot starts from an initial state $s_0 \in \mathcal{S}$ and aims to reach a state $s \in \mathcal{G}$. 
We therefore define a planning problem as $\Pi = \langle s_0, \mathcal{G}, \mathcal{M} \rangle$.
Since critical knowledge required to reach $\mathcal{G}$ may be missing from $\mathcal{M}$, planning directly in the initial model is generally infeasible. The central challenge is therefore to plan and act effectively under model incompleteness.

\subsection{Planning with Uncertain Model Expansion}
To address model incompleteness, we introduce a hypothesis-driven planning formulation that explicitly represents missing knowledge. We define a model expansion as a set of factorized hypotheses $\mathcal{H} = \{h_1, \ldots, h_n\}$, where each hypothesis $h_i$ encodes an object-centric fact, such as object existence, attribute membership, or previously unmodeled action effects. Collectively, these hypotheses parameterize an expanded world model.
Hypotheses are generated by a function $g : (\mathcal{M}, L) \rightarrow \mathcal{H}$, which produces task-relevant hypotheses conditioned on the current model and instruction, avoiding exhaustive reasoning over all unknowns. 
The truth of each hypothesis $h_i$ is defined by $v(h_i) \in \{0,1\}$.
The robot maintains a belief over the truth of hypotheses $b(\mathcal{H}) = p(v_1, \ldots, v_n \mid \tau)$, capturing uncertainty about the true environment model.
For each hypothesis $h_i$, a verification action is generated by a function $\phi : h_i \rightarrow A^i_{\text{verify}}$, where $A^i_{\text{verify}}$ denotes information-gathering actions. The overall action space is augmented as $\mathcal{A}' = \mathcal{A} \cup \{A^i_{\text{verify}}\}$.
Decision-making is formulated as a Bayes-adaptive MDP~\cite{duff2002optimal} with state $(s_0, b(\mathcal{H}))$.
Given an action $a \in \mathcal{A}'$, transitions follow the expected model $\mathcal{M}^* = \mathcal{M} \circ \mathcal{H}$ under the current hypotheses, and the hypotheses are updated based on observed outcomes. 

The planning objective is to compute a plan $\pi: \mathcal{S} \times b(\mathcal{H}) \rightarrow \mathcal{A}^*$, i.e., an action sequence, that maximizes expected cumulative reward by jointly reasoning over task execution and hypothesis verification.

\section{Open-World Planning with Hypothesis-Driven Uncertainty-Aware Model Expansion}

\begin{algorithm}[t]
\small
\caption{Hypothesis-Driven Open-World Planning}
\label{alg:overview}
\begin{algorithmic}[1]
\Require Task $L$, Model $\mathcal{M}_0$, State $s$
\State Initialize hypothesis set $\mathcal{H} \gets \emptyset$, $\mathcal{M} \gets \mathcal{M}_0$
\State History $C_{hist} \gets \emptyset$
\State $\mathcal{G} \gets \Call{Translator}{L}$
\While{not done and $i < I_{\max}$}
    \State $i \gets i + 1$
    \State $\mathcal{H} \gets \Call{HypothesesGenerator}{\mathcal{H}, \mathcal{M}, s, \mathcal{G}, C_{hist}}$
    \State $\pi \gets \Call{Planner}{s, \mathcal{G}, \mathcal{M}, b(\mathcal{H})}$
    \For{action $a \in \pi$}
        \If{$a = a_{verify}^h$}
            \State $v(h) \gets \Call{Verify}{h, C_{hist}}$
            \State $C_{hist} \gets C_{hist} \cup \{v(h)\}$
        \Else
            \State $s,obs \gets \Call{Execute}{a}$
            \State $C_{hist} \gets C_{hist} \cup \{(a, s, obs)\}$
        \EndIf
    \EndFor
\EndWhile
\end{algorithmic}
\end{algorithm}

We propose a hypothesis-driven framework for open-world planning in which a robot incrementally expands its world knowledge by generating, reasoning over, and verifying missing object-centric knowledge. As illustrated in Fig. \ref{fig:framework}, the framework operates in an iterative loop of model expansion, planning, and execution. It starts from an initial symbolic model that provides only a partial abstraction of the environment, consisting of environment knowledge represented as a scene graph and domain knowledge represented as a PDDL domain file. 
Given a task-specified goal state that cannot be achieved under the current model, a hypothesis generator identifies missing knowledge and formulates object-centric hypotheses, such as unknown object locations, novel attributes, or unmodeled action effects required for the task. 
The planner treats these hypotheses as uncertain variables and, given the current model, hypotheses, and goal, constructs action sequences that both make progress toward the task and verify the hypotheses on which the plan depends.
During execution, verification actions invoke a vision–language model to confirm or reject hypotheses, triggering model updates and replanning when necessary. 
Through this closed-loop process (Algorithm~\ref{alg:overview}), the robot progressively adapts its world model and improves performance in previously unseen environments.

\subsection{Model Initialization}
We assume that for each task class there exists an initial symbolic model $\mathcal{M}$ capturing the robot’s prior knowledge of the environment. This model encodes known objects and locations, a predicate vocabulary, and transition dynamics defined by action preconditions and effects.
In our framework, the world model is decomposed into environment knowledge and domain knowledge. Environment knowledge is represented as a scene graph of rooms, locations, and objects, along with their relations. Domain knowledge is specified in PDDL format \cite{fox2003pddl2}, defining the preconditions and effects of actions. An example action definition is shown below. 
\vspace{-0.1cm}
\begin{lstlisting}
((*@\textbf{:action}@*) pick
  (*@\textbf{:parameters}@*) (?r - Robot ?o - Obj ?l - Location)
  (*@\textbf{:precondition}@*) (and (HandEmpty ?r) (At ?r ?l) 
    (At ?o ?l))
  (*@\textbf{:effect}@*) (and ((*@$\neg$@*)HandEmpty ?r) (Holding ?r ?o) 
    ((*@$\neg$@*)At ?o ?l))
)
\end{lstlisting}
\vspace{-0.1cm}
Since PDDL provides a deterministic description of action dynamics, the initial model typically captures only limited transition knowledge. In practice, this model may be obtained through manual specification by domain experts or automatically extracted offline from large-scale datasets by prompting language models with task descriptions and demonstrations \cite{yeunidomain}. In open-world settings, however, critical information is inevitably missing, including object locations, object attributes, and object-specific action effects.

\subsection{Model Expansion as Hypotheses}
To reason beyond the initial symbolic abstraction, we explicitly represent model expansion through a set of hypotheses. Each hypothesis $h \in \mathcal{H}$ corresponds to an object-centric fact with an initially unknown truth value. Hypotheses act as tentative extensions to the world model and can be incrementally introduced, reasoned over, and verified. For example, consider the following object existence hypothesis:

\begin{lstlisting}[style=pydata]
{
    id: h1,
    type: object_existence,
    description: There may be a plate in fridge2 in kitchen2,
    content: add_fact(plate*, (Obj plate*) (Is plate* plate) (At plate* fridge2)),
    condition: None,
    verification_condition: (At ?r fridge2) (IsOpen fridge2)
}
\end{lstlisting}

This hypothesis introduces a hypothetical object instance (\texttt{plate*}) into the symbolic model and assumes its possible existence at a specific location. The \texttt{content} field defines an API call that modifies the PDDL problem file; in principle, this API can also be generated by a language model. The verification condition specifies how the hypothesis can be tested, for example by requiring the robot to be co-located with the object in an accessible open area.
More hypotheses can be constructed by conditioning on other ones. For instance, an attribute hypothesis may depend on a previous hypothesis:

\begin{lstlisting}[style=pydata]
{
    id: h2,
    type: object_attribute,
    description: The plate* is orange,
    content: add_fact(plate*, (Orange plate*)),
    condition: h1,
    verification_condition: (Holding ?r plate*)
}
\end{lstlisting}

Here, hypothesis \texttt{h2} assumes an attribute of the object introduced by \texttt{h1} and is meaningful only if the object exists. Its verification requires a more informative interaction, such as the robot holding the object.
Hypotheses can also encode action effects, such as the outcome of pressing a button:

\begin{lstlisting}[style=pydata]
{
    id: h3,
    type: action_effect,
    description: Pressing button1 may turn on the kitchen1 light,
    content: add_effect(press, (when (= ?o button1) (light_on kitchen1))),
    condition: None,
    verification_condition: (At ?r kitchen1)
}
\end{lstlisting}

Collectively, the hypothesis set factorizes uncertainty over a single, unknown expanded world model. The \texttt{content} field specifies the structural model expansion introduced by a hypothesis, while the \texttt{verification\_condition} defines the preconditions required to verify it. If the verification condition is \texttt{None}, the hypothesis is assumed to be true in our pipeline without requiring explicit verification.

\subsection{Task-Oriented Hypothesis Generation}

We formulate hypothesis generation as a structured generation problem in which a large language model (LLM) maps a natural language instruction and the robot’s current knowledge of the environment into a hypothesis space. Let $L$ denote the natural language instruction, and let $\mathcal{M}_0$ represent the initial symbolic world model. The objective is to identify a set of hypotheses $\mathcal{H}$ that bridges the gap between the goal state implied by $L$ and the current scene graph $s_0 \in \mathcal{S}$.

Given an instruction $L$, we first parse it into a formal goal specification $\mathcal{G}$, represented as a conjunction of logical predicates. For example, the instruction \textit{``Serve me a hot veggie burger at a desk''} is parsed as:
\texttt{(exists (?o ?l) (and (burger ?o) (veggie ?o) (hot ?o) (desk ?l) (at ?o ?l)))}.
By comparing the goal specification $g$ with the current scene graph and domain knowledge in $\mathcal{M}_0$, the robot identifies missing objects $\mathcal{O}_{\text{miss}}$ and missing predicates $\mathcal{P}_{\text{miss}}$. To facilitate more structured reasoning, the LLM is prompted to further categorize the missing predicates into attribute predicates (e.g., visual or semantic properties) and effect predicates (e.g., state changes resulting from interactions), yielding
$\mathcal{P}_{\text{miss}} \rightarrow \{\mathcal{P}_{\text{attr}}, \mathcal{P}_{\text{eff}}\}$.
Hypothesis generation is then performed by an LLM-based generator $\Phi_{\text{LLM}}$, which conditions on the current scene $s$, the goal condition $G$, the categorized missing predicates, and a historical context $C_{\text{hist}}$:
\begin{equation}
    H = \Phi_{\text{LLM}}
    (s_0, g, \{\mathcal{P}_{\text{attr}}, \mathcal{P}_{\text{eff}}\}, C_{\text{hist}}, \mathcal{M}),
\end{equation}
where $\mathcal{H}$ denotes a set of plausible hypotheses. To ensure consistency and avoid redundant exploration, the history context $C_{\text{hist}}$ maintains a trace of execution, previously generated hypotheses, and their verification outcomes, thereby preventing cyclic generation of hypotheses that have already been refuted.

\subsection{Planning Under Hypothesis Uncertainty}

We model the open-world planning problem as a tuple
$\Pi = \langle \mathcal{S}^+, \mathcal{P}^+, \mathcal{A}^+, \mathcal{T}^+, s_0, \mathcal{G} \rangle$,
where uncertainty is explicitly represented through a set of hypotheses
$\mathcal{H}$ generated by the hypothesis generator.
Each hypothesis $h \in \mathcal{H}$ corresponds to a proposition whose truth value is not known a priori.
For each hypothesis $h$, we introduce a belief variable
$b(v(h)) \in \{\texttt{True}, \texttt{False}, \texttt{Unknown}\}$,
indicating the estimated truth of the hypothesis.
The augmented state space is defined as
$\mathcal{S}^+ = \mathcal{S} \times \mathcal{B}$,
where $\mathcal{B}$ denotes the space of belief assignments over all hypotheses.
Initially, all hypotheses are unverified, i.e.,
$\forall h \in H,\; b(v(h)) = \texttt{Unknown}$.
To resolve hypothesis uncertainty, we introduce a set of \emph{verification actions}
$\mathcal{A}_{\text{verify}} \subset \mathcal{A}^+$.
For each hypothesis $h \in \mathcal{H}$, an abstract action
$a_{\text{verify}}^h$ is generated to observe its truth value through sensing or interaction.
Executing $a_{\text{verify}}^h$ yields a non-deterministic outcome,
turning $b(v(h))$ to either \texttt{True} or \texttt{False}.

To enable efficient planning with classical deterministic planners, we apply all-outcomes determinization \cite{yoon2007ff} to verification actions, which allows the planner to reason over individual outcomes of an otherwise non-deterministic action. Specifically, each verification action $a_{\text{verify}}^{h}$ is determinized into two actions, $a_{\text{verify}}^{h+}$ and $a_{\text{verify}}^{h-}$, corresponding to the hypothesis being verified or refuted, respectively.
In our setting, we further apply an explicit branch cut by excluding $a_{\text{verify}}^{h-}$ from the planning graph. This is because falsified hypotheses cannot satisfy the current goal specification and therefore do not contribute to successful plans under the current hypothesis expansion. As a result, the planner is encouraged to adopt an optimistic strategy in the face of hypothesis uncertainty, while negative outcomes are handled at execution time through hypothesis rejection and replanning.

The PDDL definition of a verification action $a_{\text{verify}}^h$ follows the template:
\begin{lstlisting}
((*@\textbf{:action}@*) verify_h
  (*@\textbf{:parameters}@*) (?r - Robot ?o - Obj ?l - Location)
  (*@\textbf{:precondition}@*) (and (At ?r ?l) ((*@$\neg$@*)True ?h) (Related ?h ?o) (*@$\phi_{cond}$@*) (*@$\phi_{verify}$@*))
  (*@\textbf{:effect}@*) (and (True ?h))
)
\end{lstlisting}
where $\phi_{cond}$ encodes dependencies between hypotheses and
$\phi_{verify}$ specifies the sensing or interaction constraints required to verify $h$.
To bias the planner toward resolving uncertainty early, we associate a high cost
with actions that rely on unverified hypotheses.
Specifically, for any action $a$ involving an object associated with hypothesis $h$,
an additional penalty cost $c \gg 0$ is incurred if $b(v(h)) = \texttt{Unknown}$.
This design encourages the planner to schedule verification actions
$a_{\text{verify}}^h$ before manipulation actions that depend on $h$,
effectively trading short-term verification cost for reduced execution risk due to incorrect hypotheses.
The resulting deterministic planning problem is solved using the Fast Downward planner \cite{helmert2006fast}.

\subsection{Hypothesis Verification and Model Update}

During execution, standard actions are carried out by the robot’s low-level skill library.
When the plan prescribes a verification action $a \in \mathcal{A}_{\text{verify}}$, the system invokes a dedicated verification routine. 
Since the planning process explicitly reasons about the preconditions $\texttt{verification\_condition}$ required for hypothesis verification, such actions are executed only when the robot is already in a state suitable for reliable verification, as predicted by the language model.
In practice, verification actions are grounded using either prebuilt perception modules or a VLM, depending on the type of hypothesis. 
In our implementation, hypotheses concerning object existence are verified by comparing the hypothesized objects against a newly observed scene graph, whose nodes are constructed by object detectors. 
For more fine-grained hypotheses, such as object attributes or action effects, a VLM is queried with the current image or the latest several images together with the hypothesis description to perform question answering.
Based on the resulting sensory feedback, each hypothesis is either confirmed or refuted. If a hypothesis is verified, the corresponding fact is incorporated into the world model, provided that it does not conflict with existing knowledge. If the hypothesis is refuted, it is marked as false ($v_h \leftarrow \text{False}$) and recorded in the history context $C_{\text{hist}}$. The hypothesis generator is then re-invoked to produce alternative hypotheses conditioned on the accumulated history and newly available observations, triggering a replanning cycle.

\subsection{Necessary Assumptions}
We summarize the key assumptions underlying our framework for extending automated planning with open-world knowledge from foundation models. These assumptions are consistent with prior work in automated planning and LLM-based planning, and are necessary for making the problem tractable and empirically feasible. Our framework relies on the following assumptions:
(1)\textit{ Task Scoping}: We focus on uncertainty from underspecified symbolic world modeling and assume reliable execution of a library of atomic skills when under full observability.
(2) \textit{Goal Requirements}: Task goals must be expressible in structured logic, and define the abstraction of missing knowledge and be verifiable using the robot’s sensing capabilities at some known configurations. In particular, missing object concepts and predicates are assumed to be encoded in the task instructions. 
(3) \textit{Language Model}: The language model is assumed to be able to generate correct hypotheses within a bounded number of attempts, possess the internal reasoning ability to propose verification conditions, and evaluate hypotheses from raw sensory observations.
\section{Experiments and Results}
To examine the usage and impact of uncertainty-aware model expansion in open-world planning, we evaluate HUME guided by the following research questions:

\RQI{}: \textit{Does explicitly representing model expansion enable effective planning under incomplete knowledge?}

\RQII{}: \textit{How does the planner's awareness of hypothesis uncertainty influence decision-making and exploration behavior?}

\begin{figure}[!tbp]
  \centering
    \vspace{-0.3cm}
  \includegraphics[width=0.99\linewidth]{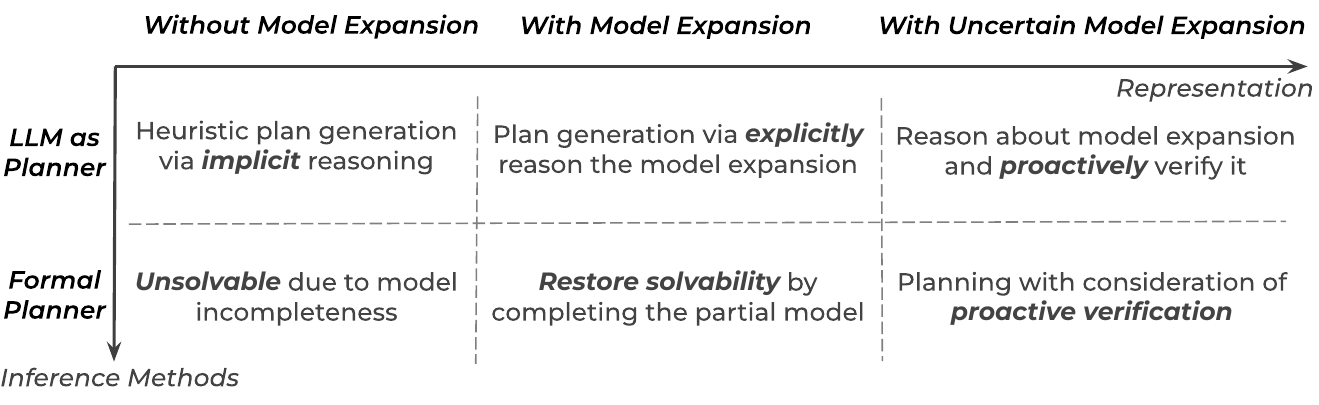}
  \vspace{-0.6cm}
  \caption{
  Overview of the analysis of six planning categories, organized by representation and inference methods.
  }
  \label{fig:approaches}
\vspace{-0.7cm}
\end{figure}

\subsection{Approaches}

We compare six approaches that differ along two principal axes: (1) \textit{\textbf{Representation}}, namely whether model expansion is explicit and whether uncertainty awareness is incorporated; and (2) \textbf{\textit{Inference Method}}, which distinguishes between formal symbolic planners and Large Language Models (LLMs). See Fig.~\ref{fig:approaches} for an overview. We use \texttt{gpt-4.1-2025-04-14} as the language model across all modules and approaches.

\begin{enumerate}[leftmargin=*,parsep=0pt, topsep=0pt]
\item \textit{Formal Planner w/o Model Expansion:} A standard PDDL planner operating on the initial partial model.
\item \textit{LLM as Planner w/o Model Expansion:} An LLM Planner that directly predicts action sequences from observations and instructions, without explicit world model expansion, 
representing a set of LLM-based methods \cite{huang2023inner, singh2023progprompt, liang2023code, song2023llmplanner}.
\item \textit{Formal Planner with Model Expansion:} A PDDL planner augmented with hypotheses generation, treating all hypothesized knowledge as definite facts. This represents a set of language-augmented symbolic planners \cite{ding2023integrating, chen2024language}.
\item \textit{LLM as Planner with Model Expansion:} An LLM Planner uses the same hypothesis generation mechanism, but relies on an LLM to plan over the expanded knowledge base. 
\item \textit{Formal Planner with Uncertain Model Expansion:} A PDDL planner considers hypotheses as uncertain object-centric facts, enabling plans that both achieve task goals and actively verify assumptions.
\item \textit{LLM as Planner with Uncertain Model Expansion:} An LLM Planner considers hypotheses as uncertain object-centric facts, enabling direct comparison with symbolic planning under same representation.
\end{enumerate}

\subsection{Simulation Scenarios}

\begin{figure}[t]
  \centering
  \begin{minipage}{0.46\linewidth}
    \centering
    \includegraphics[width=\linewidth]{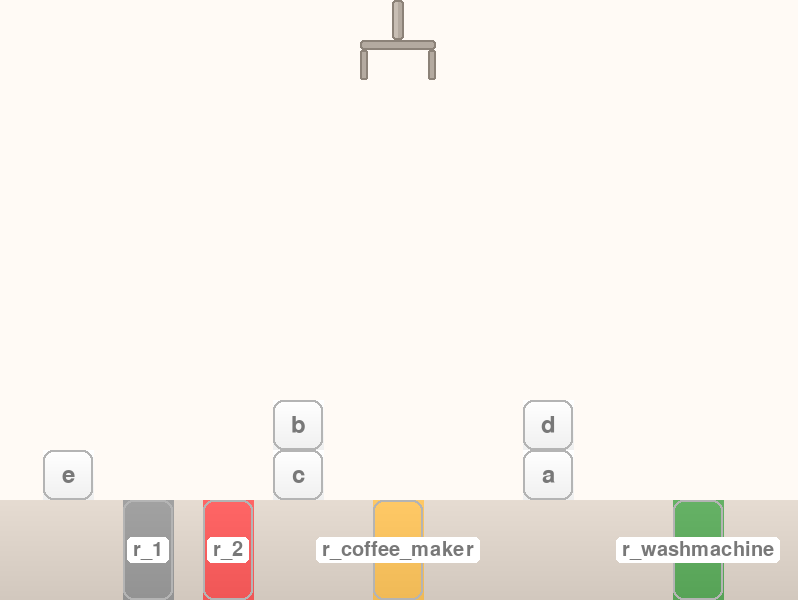}
  \end{minipage}
    \hspace{0.04\linewidth}
  \begin{minipage}{0.46\linewidth}
    \centering
    \includegraphics[width=\linewidth]{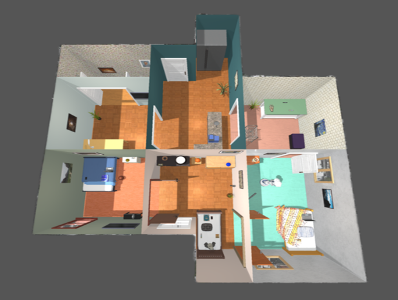}
  \end{minipage}
  \caption{Illustration of the Simulation Setup: Block Processing World (Left) and Mobile Manipulation in Unknown Environments (Right).}
  \label{fig:sim_setup}
  \vspace{-0.5cm}
\end{figure}

\subsubsection{Block Processing World}
As illustrated in Fig.~\ref{fig:sim_setup} left, Block Processing World extends the classic Blocks World domain with multiple block processors, each producing a specific effect, such as cleaned, wet, or hot. 
The robot does not know each processor’s effect a priori and needs to interact with processors to discover their effects. 
Similar to Blocks World, the goal is to build a target configuration while ensuring selected blocks exhibit designated effects.
See the Appendix for more details.

\subsubsection{Mobile Manipulation in Unknown Environments}
In this setting (Fig.~\ref{fig:sim_setup}, right), a mobile manipulator operates in a household-like environment with objects of unknown locations and attributes.
The robot must rearrange objects to achieve a target configuration, such as delivering an object with a specific attribute to a target location. Since object location, properties, and action effects are unknown, the robot must actively explore and interact with the environment to infer object attributes, uncover hidden objects, and predict action outcomes.
We implement the benchmark in AI2-THOR \cite{kolve2017ai2} and evaluate eight tasks across five ProcTHOR houses \cite{deitke2022procthor}, with three trials per house.

\begin{small}
\begin{enumerate}[noitemsep, label=\textbf{T\arabic*.}, align=left, labelwidth=0.5cm]

    \item \textit{Store the remote control in a drawer.}
    \item \textit{Take a fork to the dining table.}
    \item \textit{Take a transparent bowl to the countertop.}
    \item \textit{Move a damask-style pillow to a sofa.}
    \item \textit{Carry a metal kettle to a fridge.}
    \item \textit{Place a polished-gold vase on a desk.}
    \item \textit{Deliver coffee in a mug with a logo to the dining table.}
    \item \textit{Take a wet teal cloth to the countertop.}
\end{enumerate}
\end{small} 

\begin{figure}[h]
  \centering
  \vspace{-0.3cm}
  \begin{minipage}{0.493\linewidth}
    \centering
    \includegraphics[width=\linewidth]{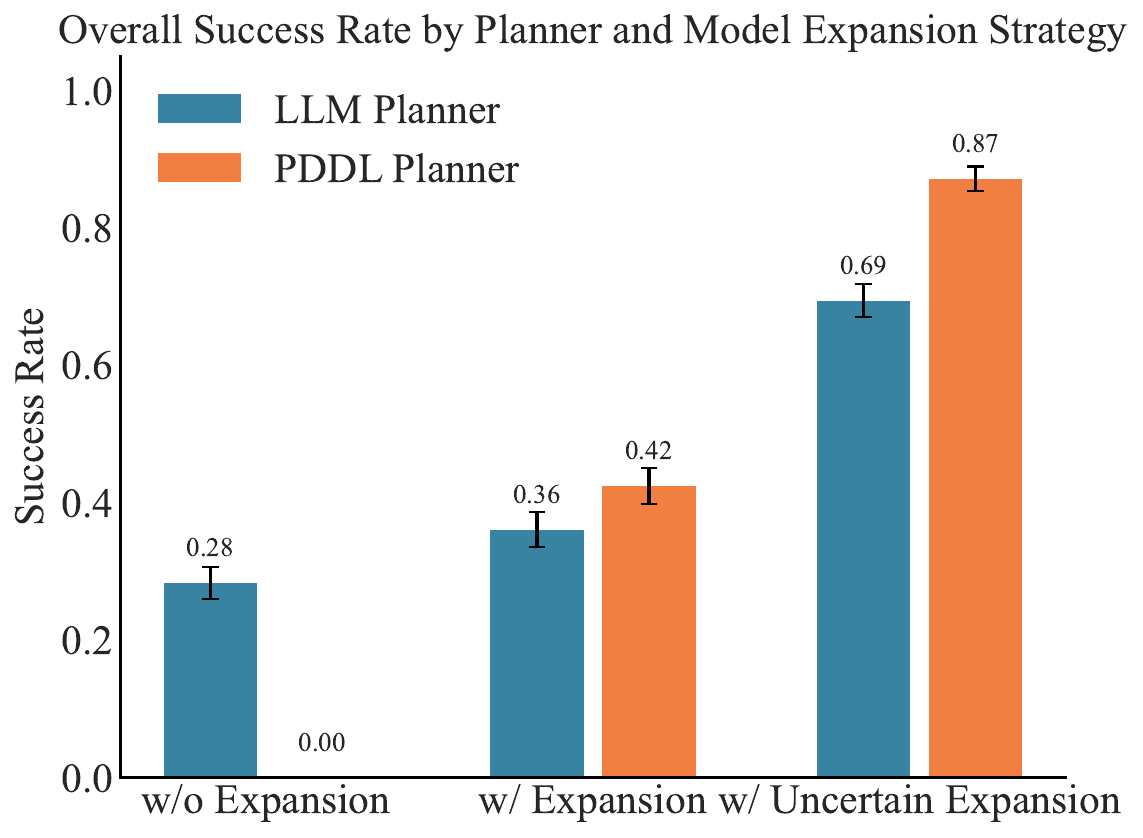}
  \end{minipage}
  \hfill
  \begin{minipage}{0.493\linewidth}
    \centering
    \includegraphics[width=\linewidth]{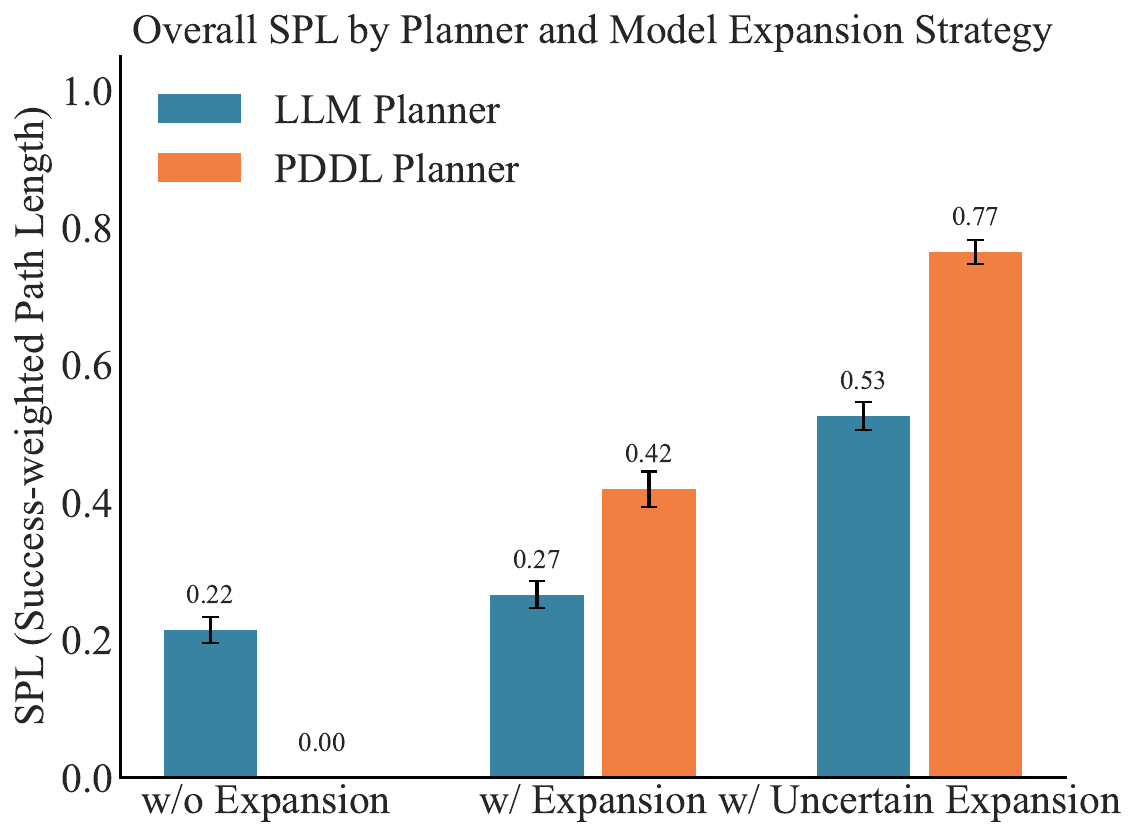}
  \end{minipage}
  \vspace{-0.3cm}
  \caption{Simulation Results on Block Processing World: Success Rate and Success weighted by Path Length (SPL).}
  \label{fig:block_results_1}
  \vspace{-0.5cm}
\end{figure}

\begin{figure}[t]
  \centering
  \includegraphics[width=1.0\linewidth]{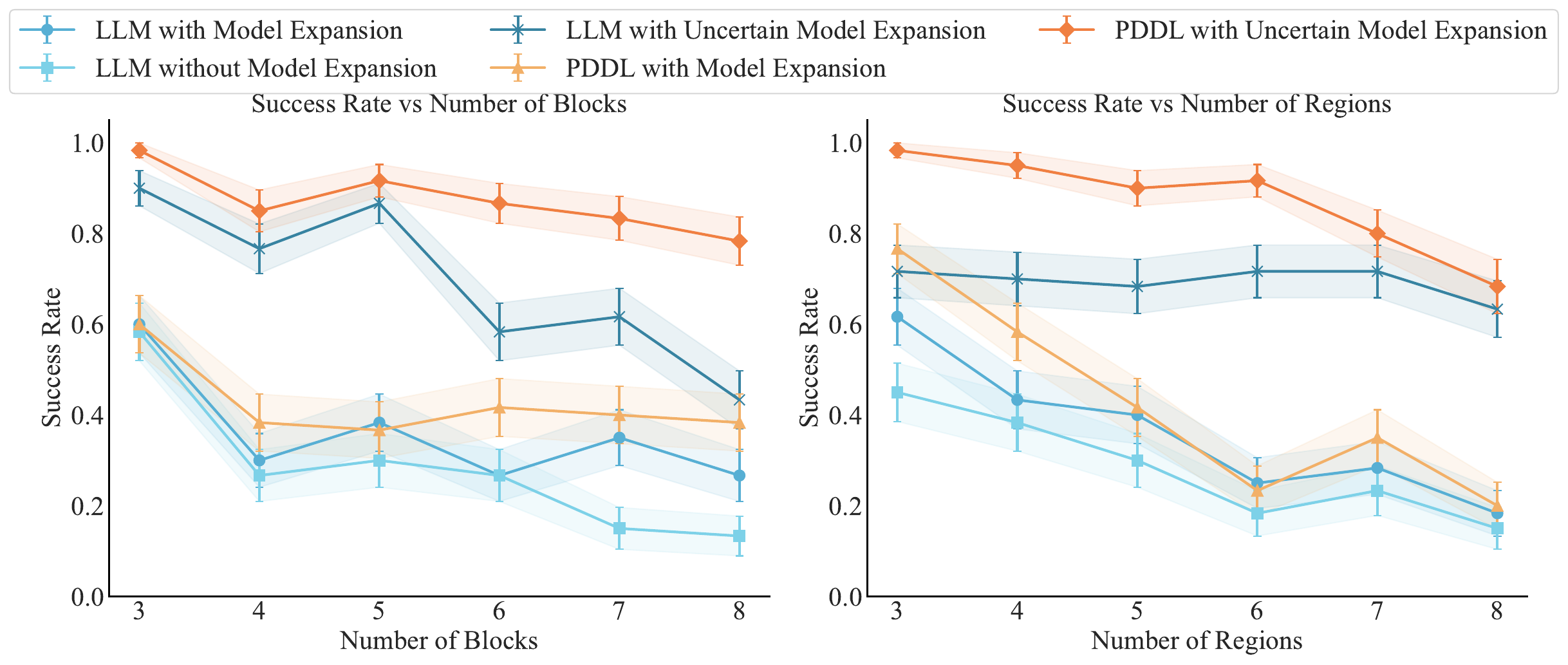}
    \vspace{-0.7cm}
  \caption{Simulation Results on the Block Processing World: Performance Trends with Varying Numbers of Blocks and Processors.}
  \label{fig:block_results_2}
  \vspace{-0.3cm}
\end{figure}

\begin{figure}[t]
  \centering
  \begin{minipage}{0.493\linewidth}
    \centering
    \includegraphics[width=\linewidth]{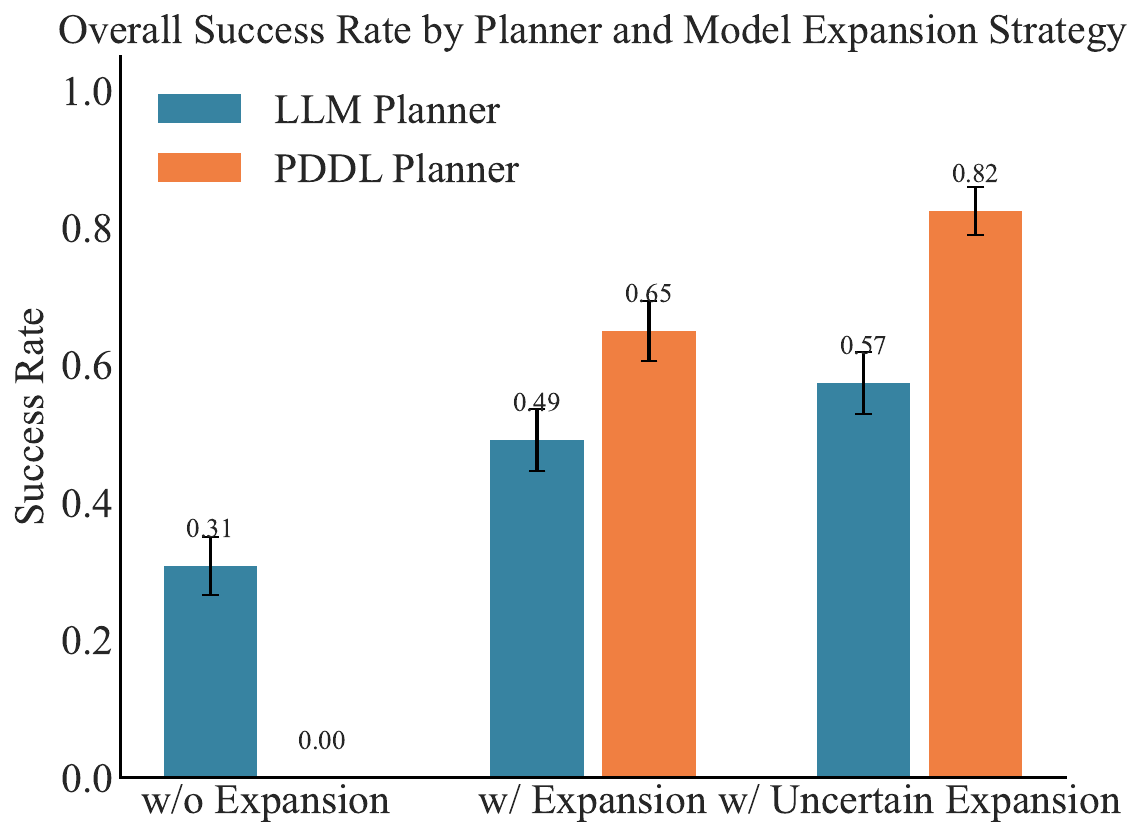}
  \end{minipage}
  \hfill
  \begin{minipage}{0.493\linewidth}
    \centering
    \includegraphics[width=\linewidth]{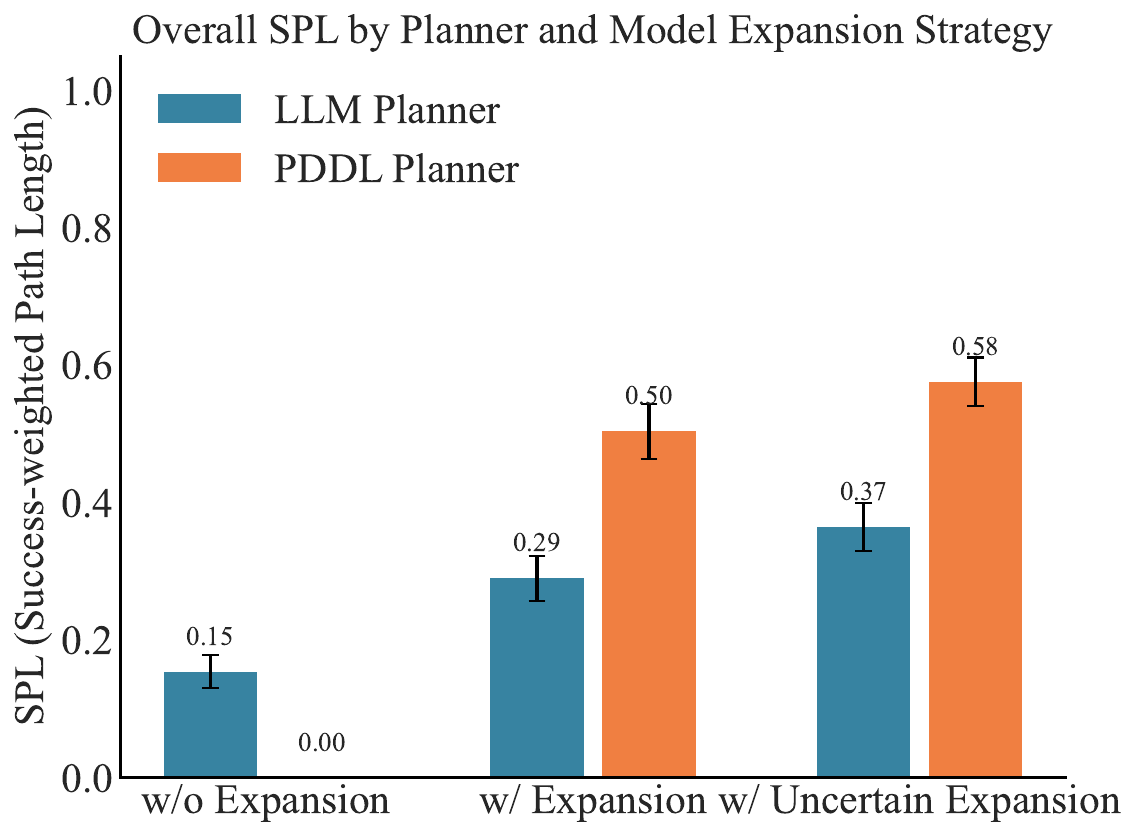}
  \end{minipage}
  \caption{Simulation Results on Mobile Manipulation in Unknown Environments: Success Rate and Success weighted by Path Length (SPL)}
  \label{fig:mm_result_1}
  \vspace{-0.6cm}
\end{figure}

\begin{figure}[t]
  \centering
  \includegraphics[width=0.99\linewidth]{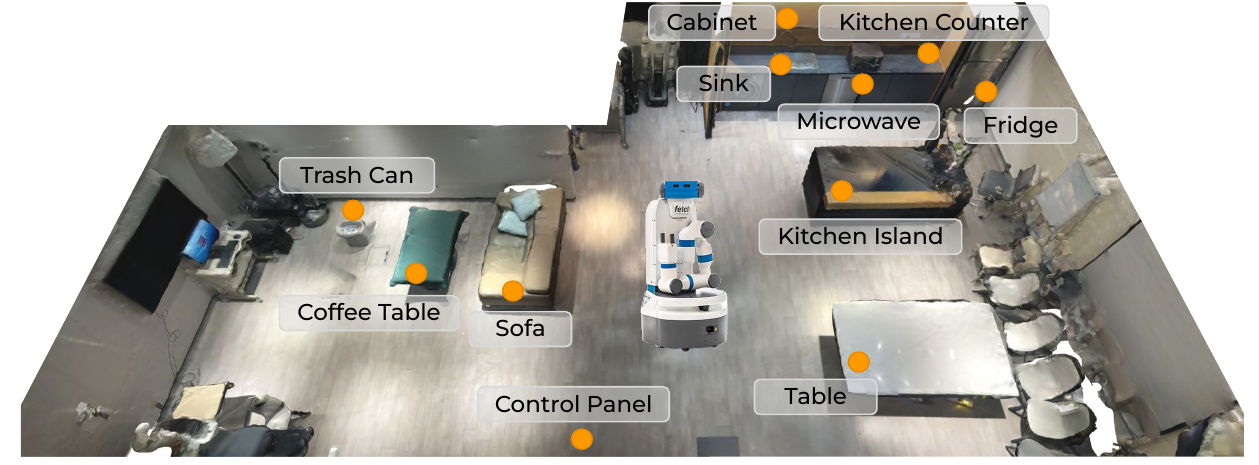}
  \vspace{-0.5cm}
  \caption{The real-world experimental setup comprising a kitchen, living room, and dining room, with a Fetch mobile manipulator.
  }
  \label{fig:real_setup}
\vspace{-0.5cm}
\end{figure}

\subsection{Results and Analysis in Simulation}

We evaluate the proposed approaches in both the Block Processing World and the AI2-THOR mobile manipulation tasks. As shown in Fig. \ref{fig:block_results_1}, \textit{explicitly representing and performing model expansion is critical for effective planning under incomplete world knowledge}. Both the LLM-based planner and the formal PDDL planner achieve substantial performance gains when hypotheses are introduced and maintained, while planners without model expansion consistently fail, demonstrating that static initial models are insufficient for open-world tasks. The formal planner benefits from explicit hypothesis representation which restores solvability, whereas the LLM planner relies on heuristic plan generation via implicit reasoning, which is less reliable without explicit hypothesis maintenance.
\textit{Incorporating uncertainty awareness into the expanded model further improves performance across planners.} Uncertainty-aware model expansion consistently outperforms deterministic expansion in both success rate and SPL, highlighting the importance of reasoning about hypothesis validity rather than only generating hypotheses. Similar trends are observed in Fig. \ref{fig:block_results_2} as the number of regions increases, where uncertainty-aware expansion yields more robust performance under increasing task complexity.
In Fig. \ref{fig:block_results_2}, \textit{as problem openness increases or task complexity decreases, explicit and uncertainty-aware model expansion becomes a stronger determinant of success than the specific choice of planner}. In these regimes, the performance gap between LLM-based and formal planners narrows, while proper handling of uncertain hypotheses remains the dominant factor affecting performance, suggesting a planner-agnostic advantage.

These trends extend to the AI2-THOR mobile manipulation benchmark (Fig.~\ref{fig:mm_result_1}), where both planners benefit from model expansion and uncertainty-aware reasoning. 
However, the relative improvement of the LLM-based planner is smaller than in BlockWorld, likely due to the increased complexity arising from the large number of objects and locations in mobile manipulation environments, which makes verification and planning constraints harder for LLMs to enforce.

\subsection{Real World Experiments}
We validate HUME on a physical Fetch mobile manipulator \cite{wise2016fetch} operating in a typical home environment comprising a kitchen, living room, and meeting room (Fig.~\ref{fig:real_setup}). For perception, the robot maintains a 3D scene graph based on an existing localization map, while new objects are instantiated through a modular perception pipeline. This pipeline employs OWLv2~\cite{minderer2024scaling} for object detection and the Segment Anything model~\cite{kirillov2023segment} for mask extraction, with RGB-D observations fused for 3D back-projection.
For low-level control, we implement a set of modular skills, including \texttt{navigate}, \texttt{pick}, \texttt{place}, and \texttt{trigger}, as well as learning-based skills like \texttt{open} and \texttt{close}. The learning-based skills are obtained via post-training of the $\pi_{0.5}$ model \cite{intelligence2504pi0} on collected demonstration data. Further details of the real-world mobile manipulation system are provided in the Appendix.

The robot is evaluated on five tasks exhibiting substantial openness ($T1$--$T5$):
\begin{small}
\begin{enumerate}[noitemsep, label=\textbf{T\arabic*.}, align=left, labelwidth=0.5cm]
    \item \textit{Deliver a zero-sugar drink to the table.}
    \item \textit{Place the remote with a red button into the cabinet.}
    \item \textit{Move the smiley-face mug to the fridge.}
    \item \textit{Serve a heated chicken burger on the coffee table.}
    \item \textit{Throw away the blue-floral bowl and turn off the living room light.}
\end{enumerate}
\end{small}

To isolate planning performance from execution noise, we mitigate random execution failures by retrying conceptually feasible low-level skills until success. When execution failures result in invalid states (e.g., object drops), we manually reset the state based on the expected outcome. We evaluate all five planners on these five tasks, with three trials per task.

\begin{figure}[h]
  \centering
  \begin{minipage}{0.493\linewidth}
    \centering
    \includegraphics[width=\linewidth]{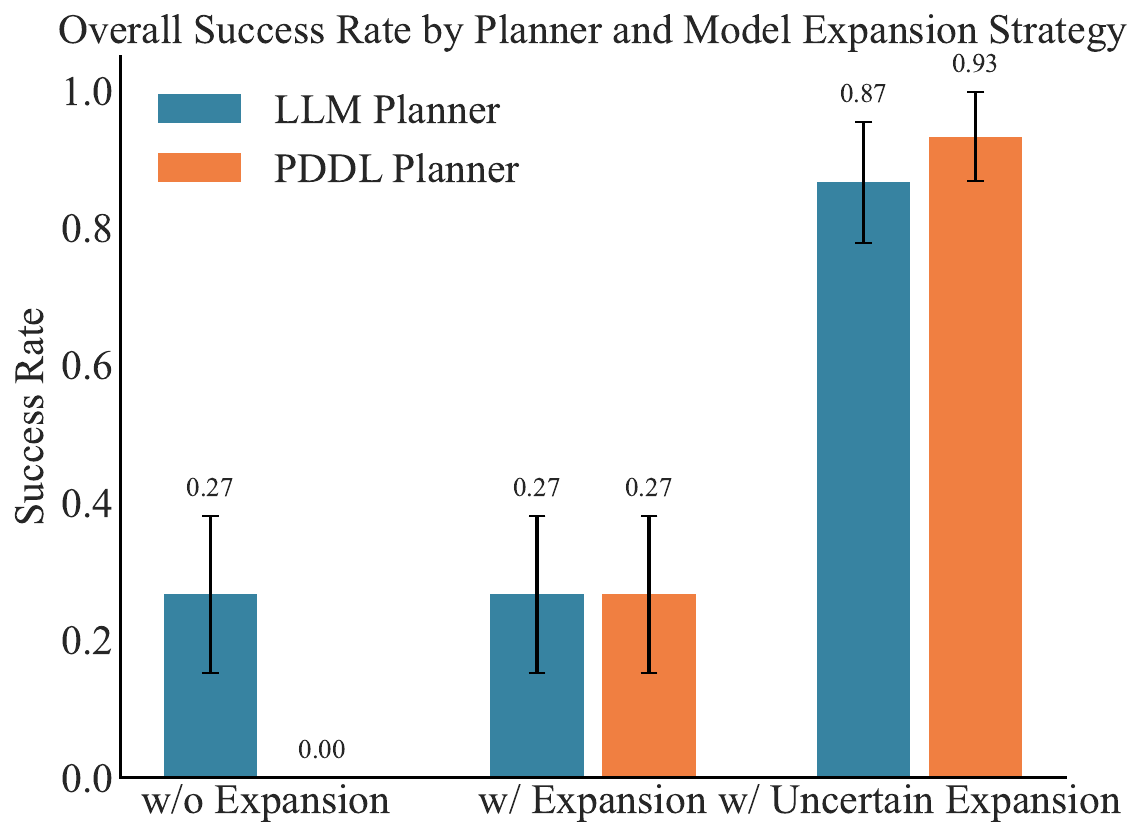}
  \end{minipage}
  \hfill
  \begin{minipage}{0.493\linewidth}
    \centering
    \includegraphics[width=\linewidth]{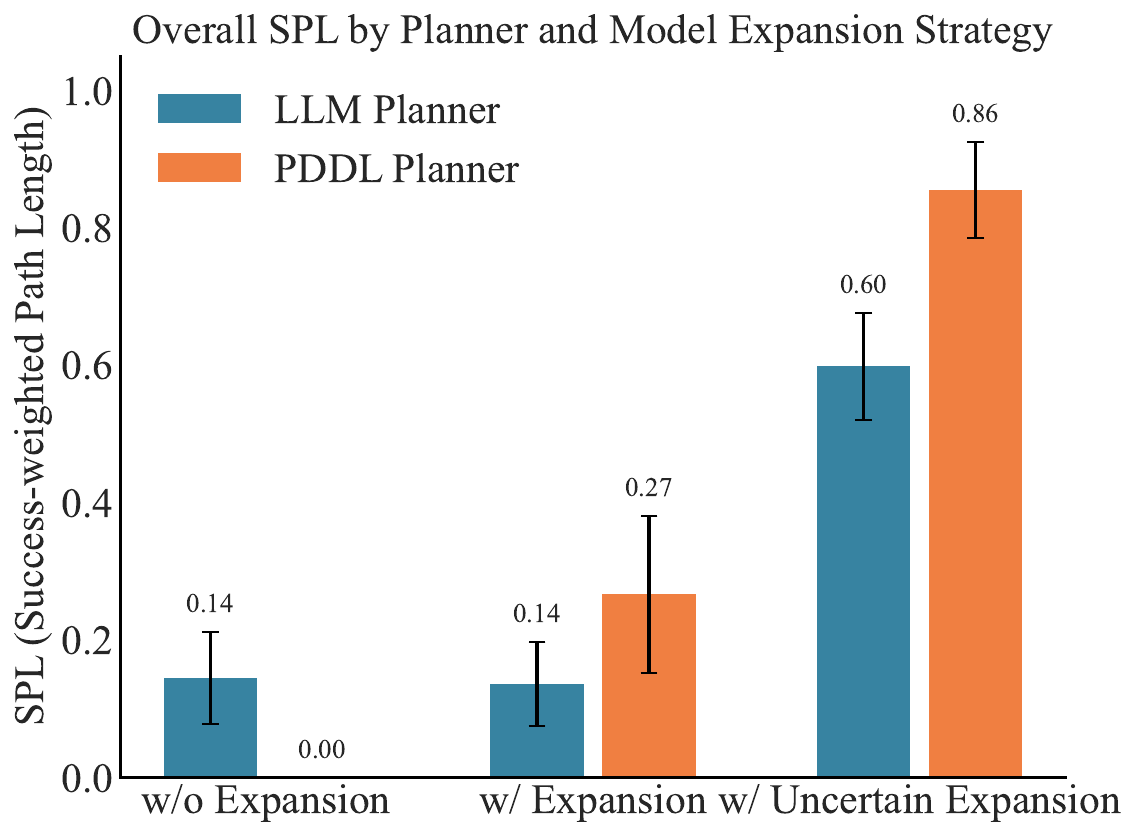}
  \end{minipage}
  \vspace{-0.3cm}
  \caption{Real-World Experiment Results: Success Rate and Success weighted by Path Length (SPL)}
  \label{fig:mm_result_2}
  \vspace{-0.3cm}
\end{figure}

\begin{figure*}[t]
  \centering
  \setlength{\tabcolsep}{1pt}
  \begin{tabular}{cccccc}
    \includegraphics[width=0.16\linewidth]{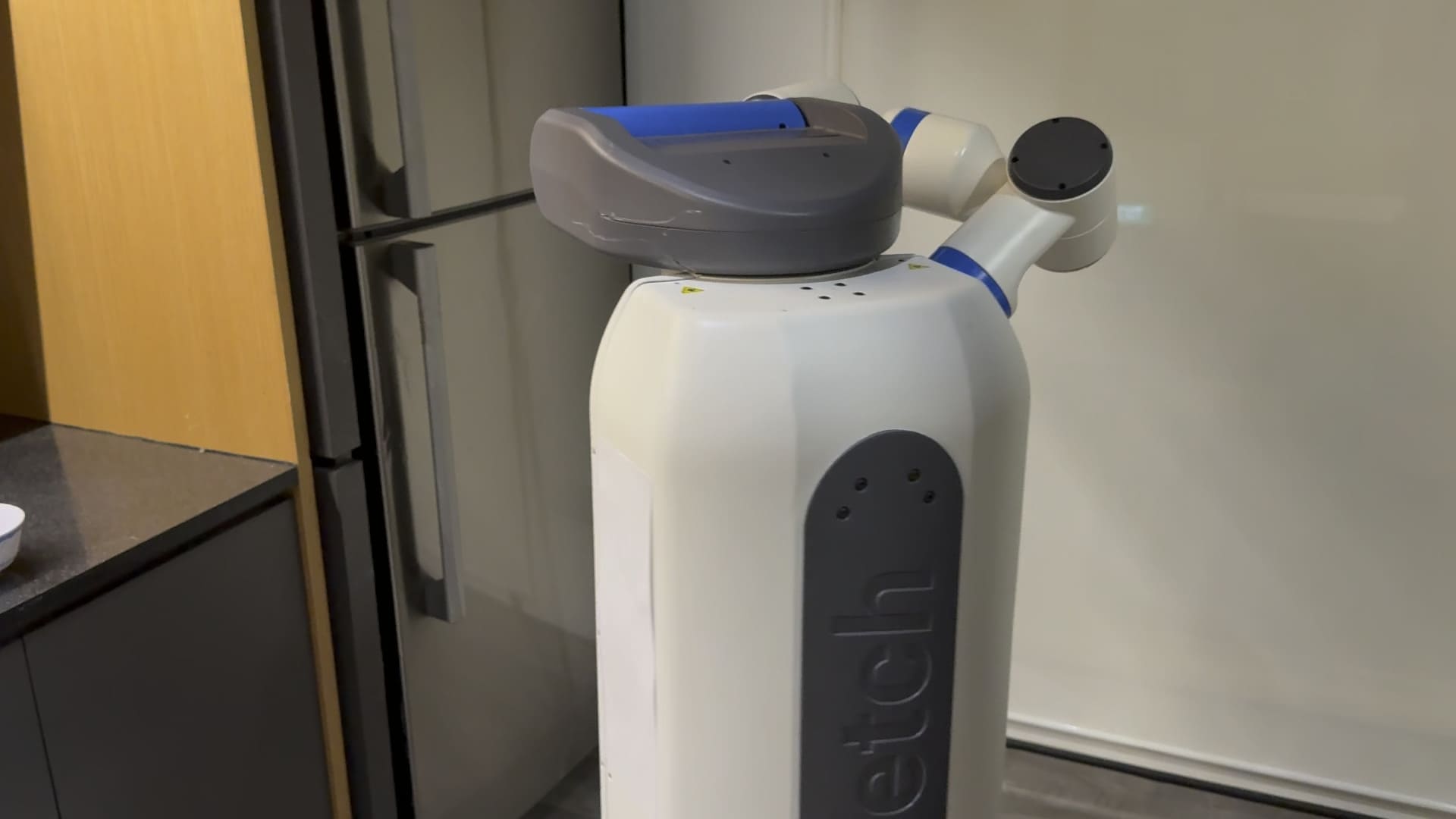} &
    \includegraphics[width=0.16\linewidth]{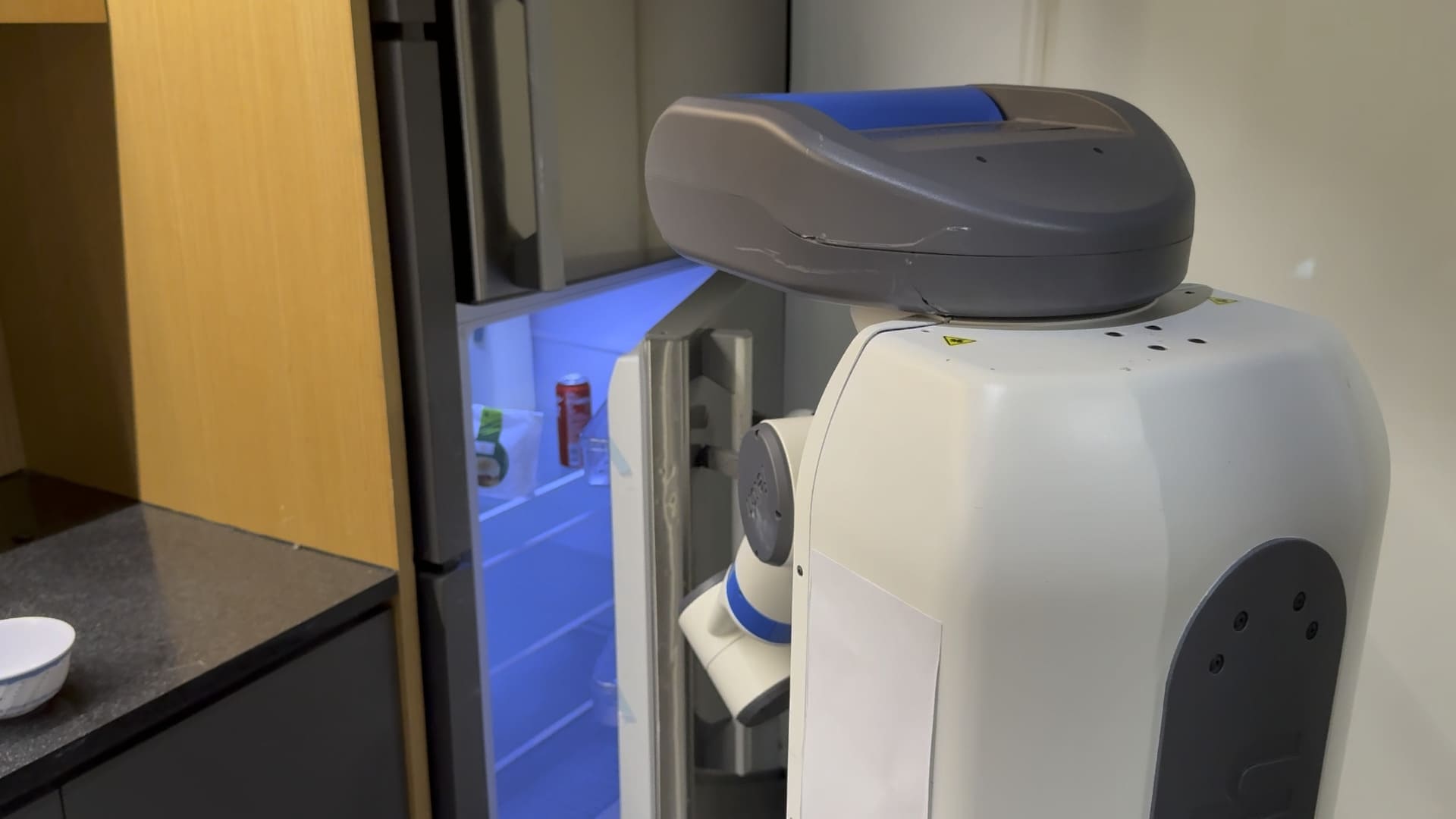} &
    \includegraphics[width=0.16\linewidth]{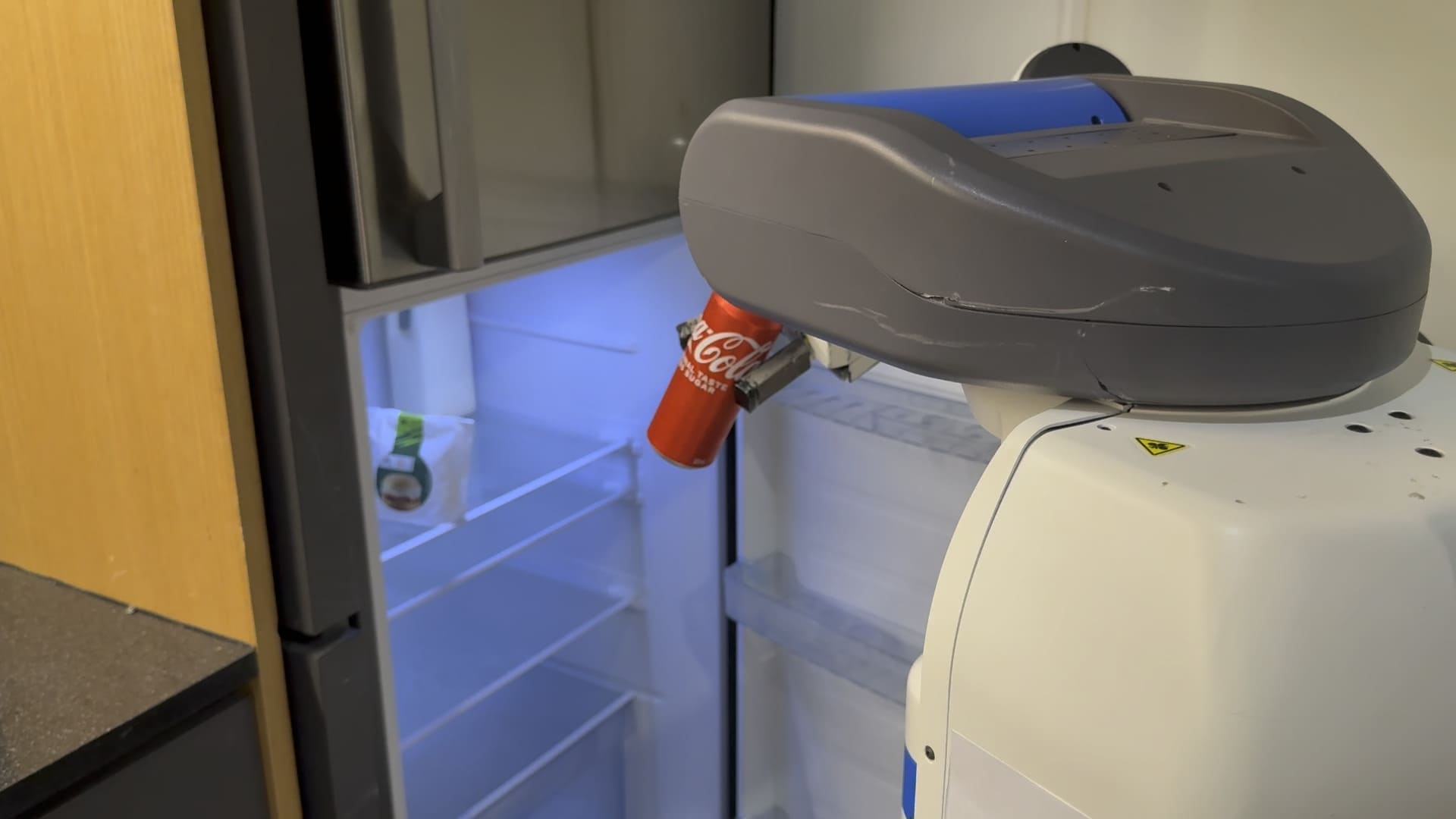} &
    \includegraphics[width=0.16\linewidth]{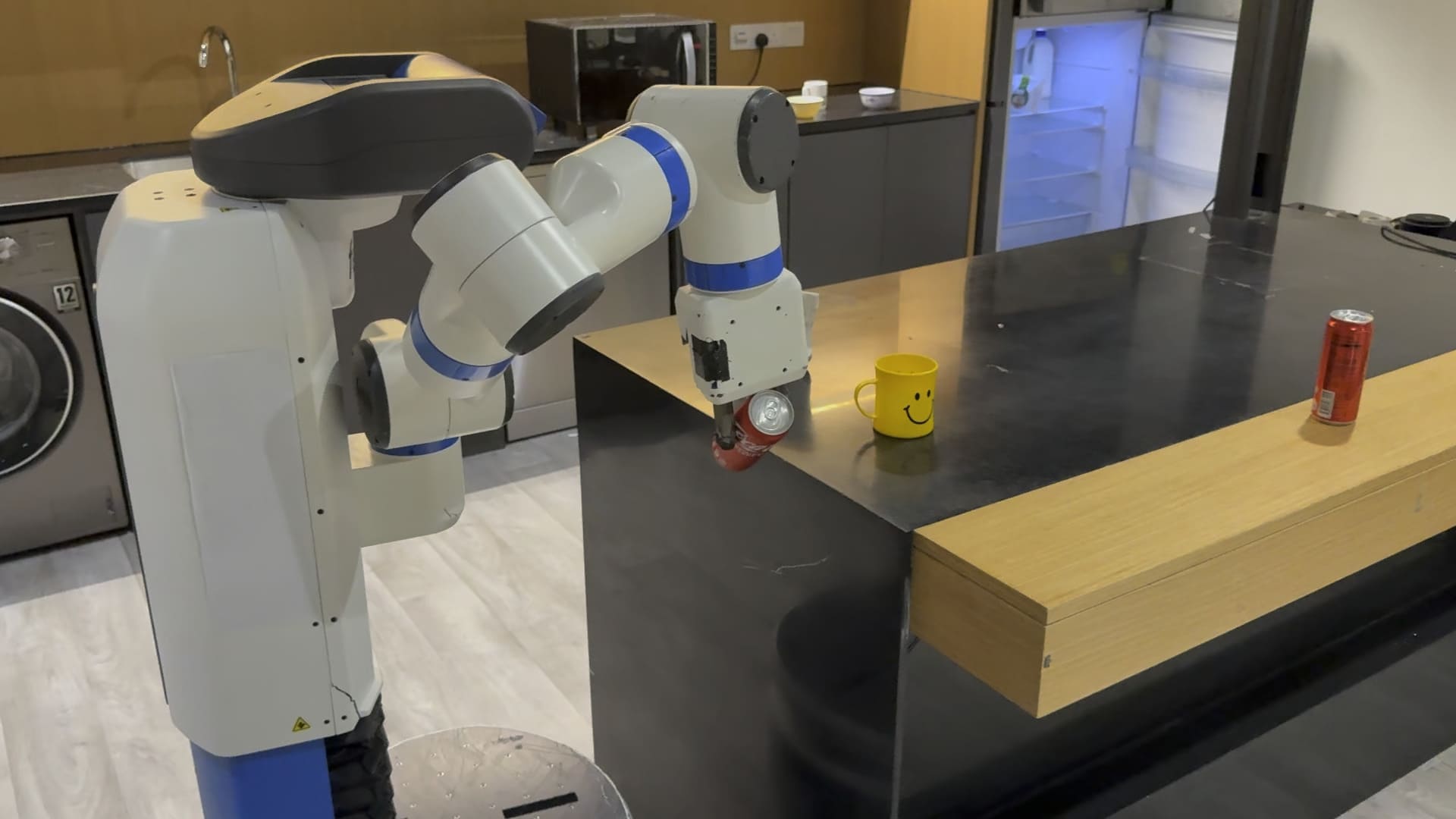} &
    \includegraphics[width=0.16\linewidth]{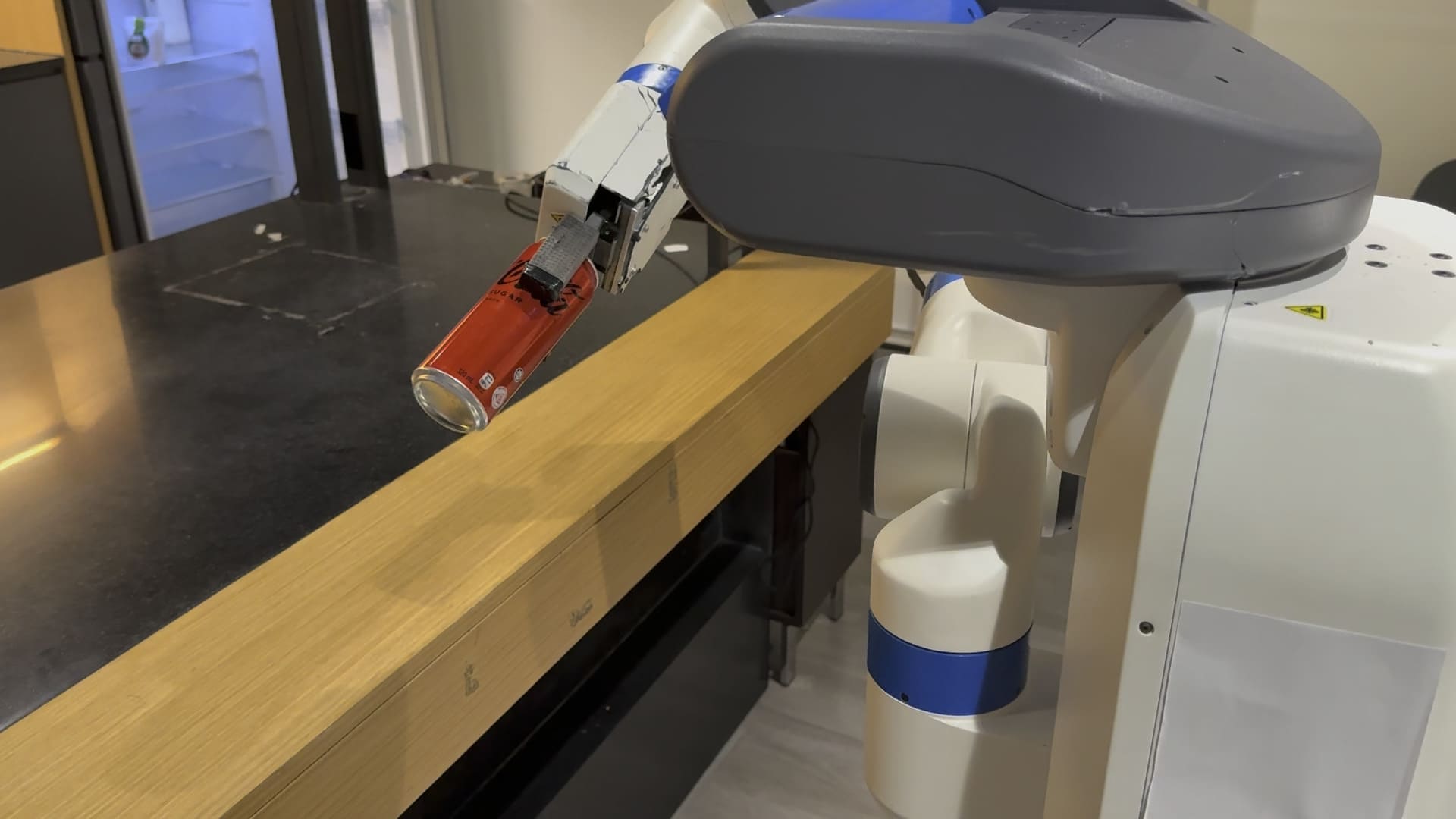} &
    \includegraphics[width=0.16\linewidth]{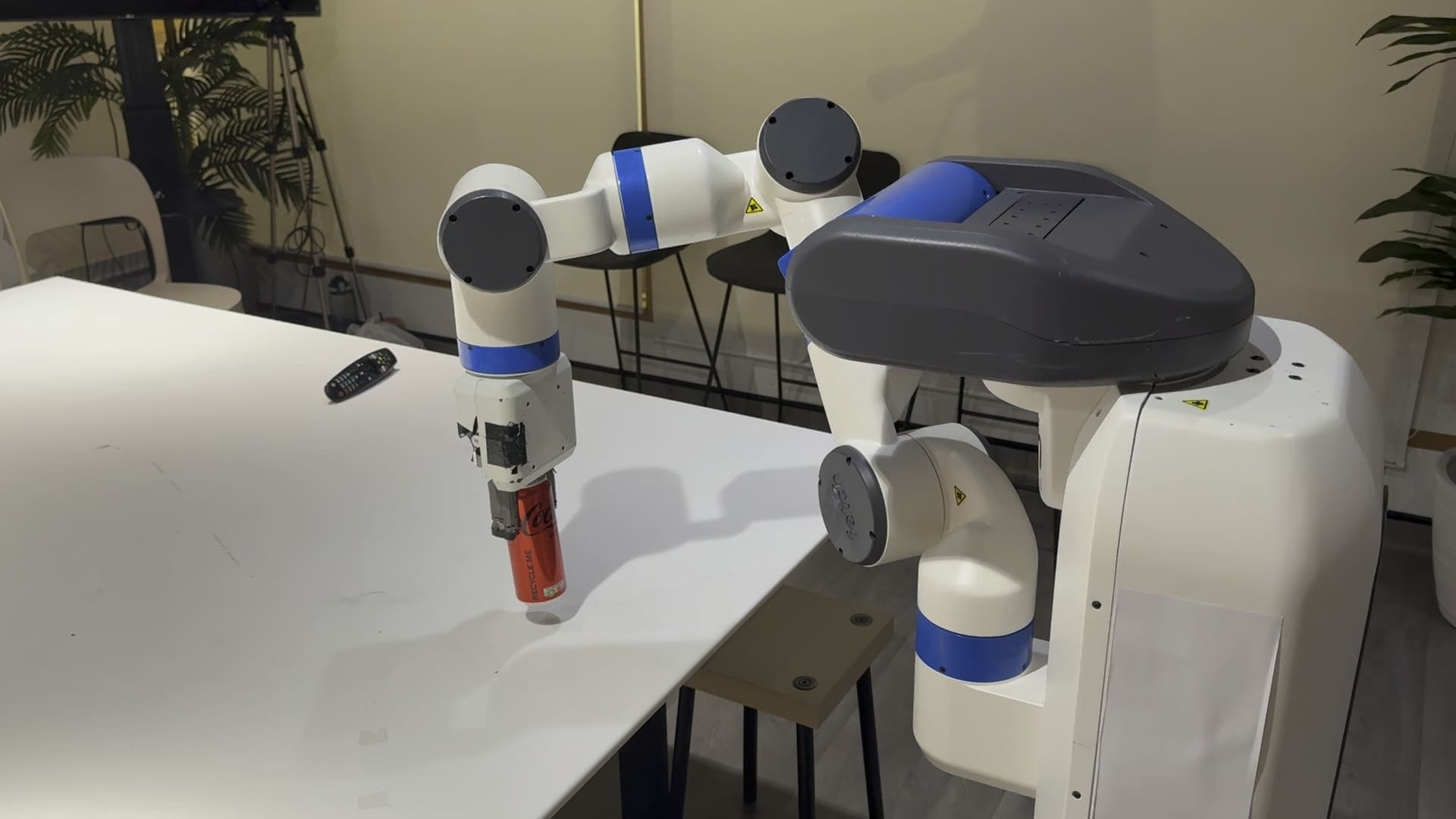} \\
  \end{tabular}
  \caption{Snapshot of completing the task \textbf{T1}: \textit{Deliver a zero-sugar drink to the table.} The robot is required to infer the drink’s location, identify the zero-sugar option, and deliver it to the table.}
  \label{fig:real_example_1}
  \vspace{-0.2cm}
\end{figure*}

\begin{figure*}[t]
  \centering
  \setlength{\tabcolsep}{1pt}
  \begin{tabular}{cccccc}
    \includegraphics[width=0.16\linewidth]{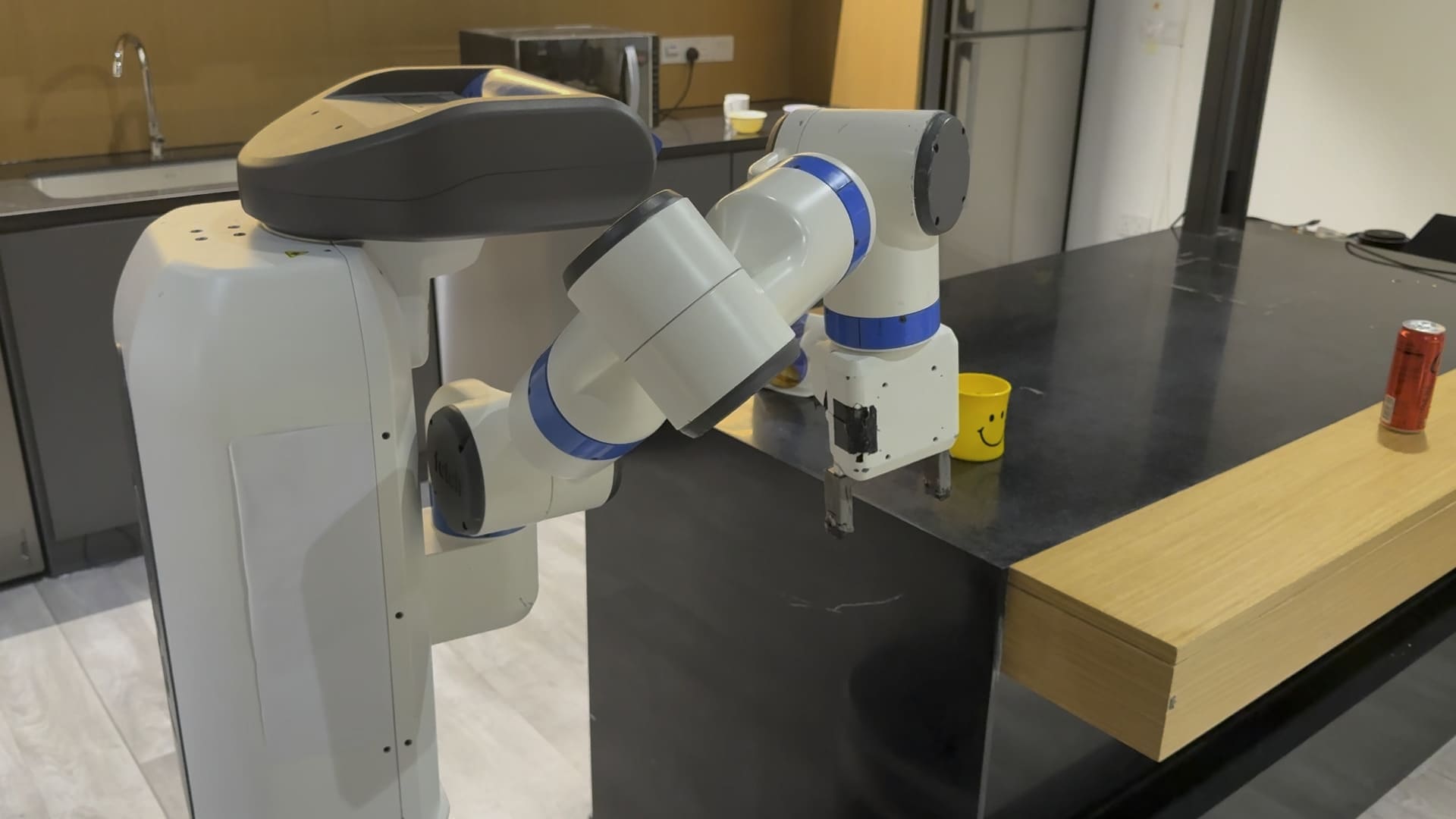} &
    \includegraphics[width=0.16\linewidth]{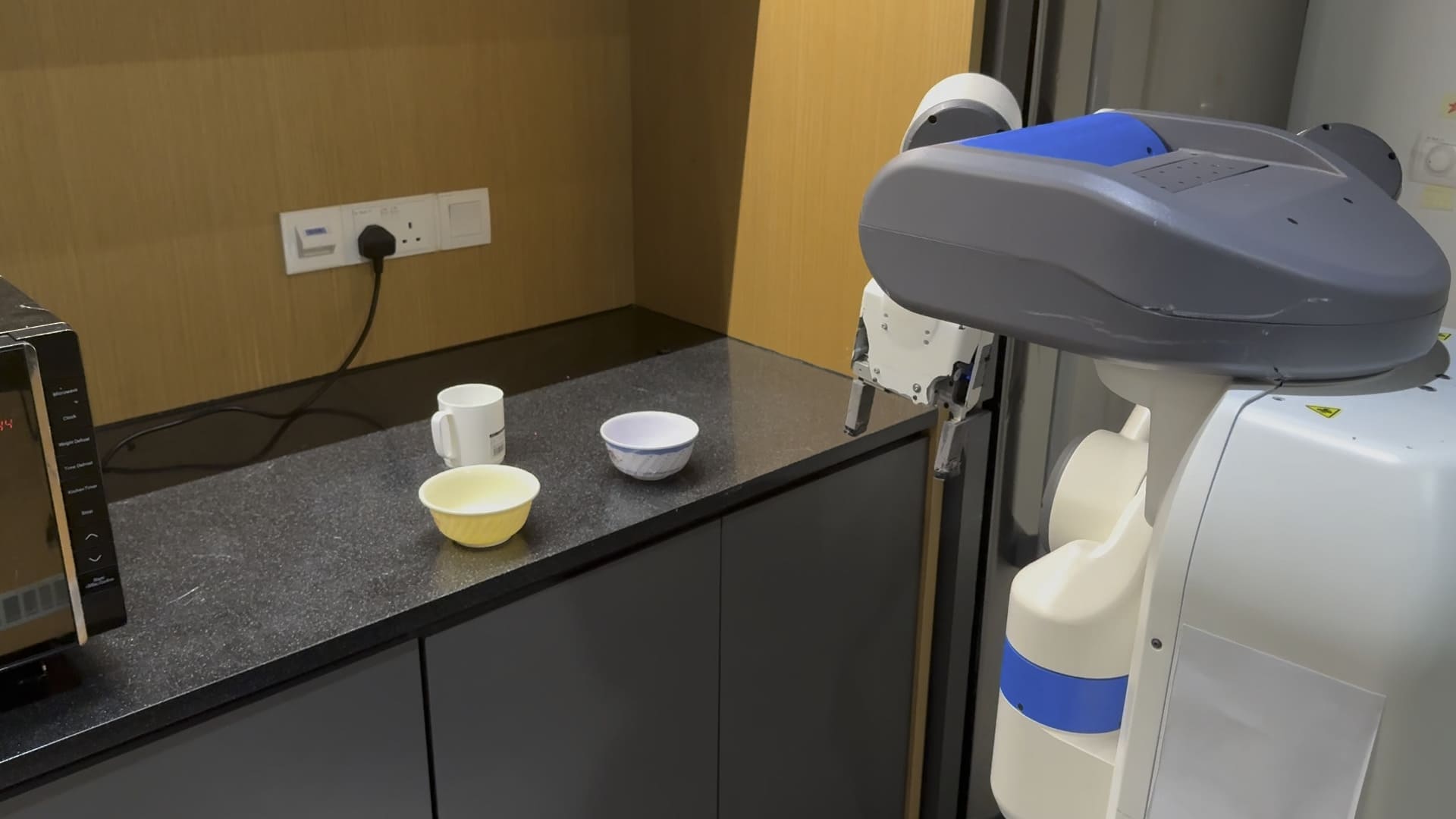} &
    \includegraphics[width=0.16\linewidth]{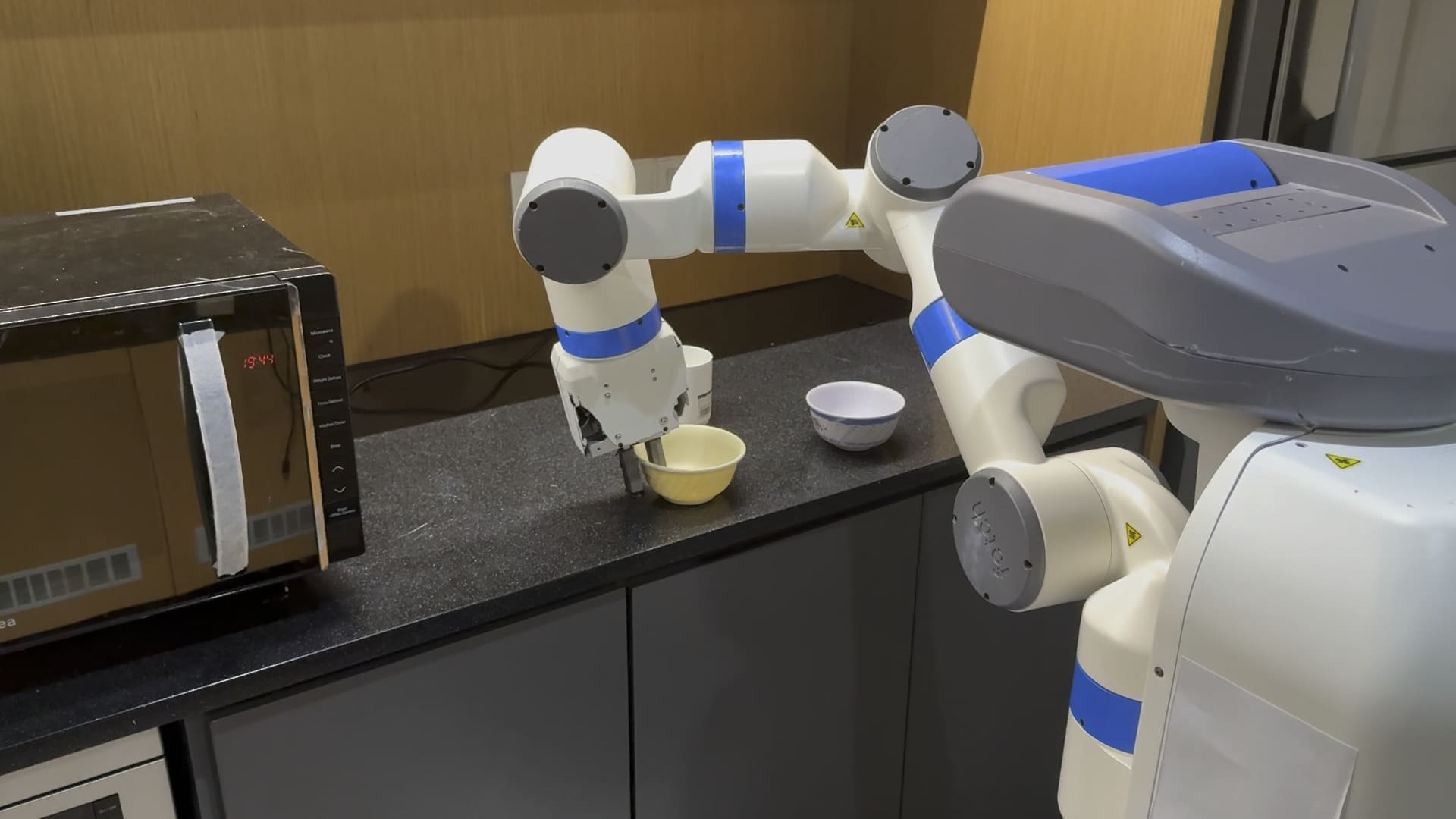} &
    \includegraphics[width=0.16\linewidth]{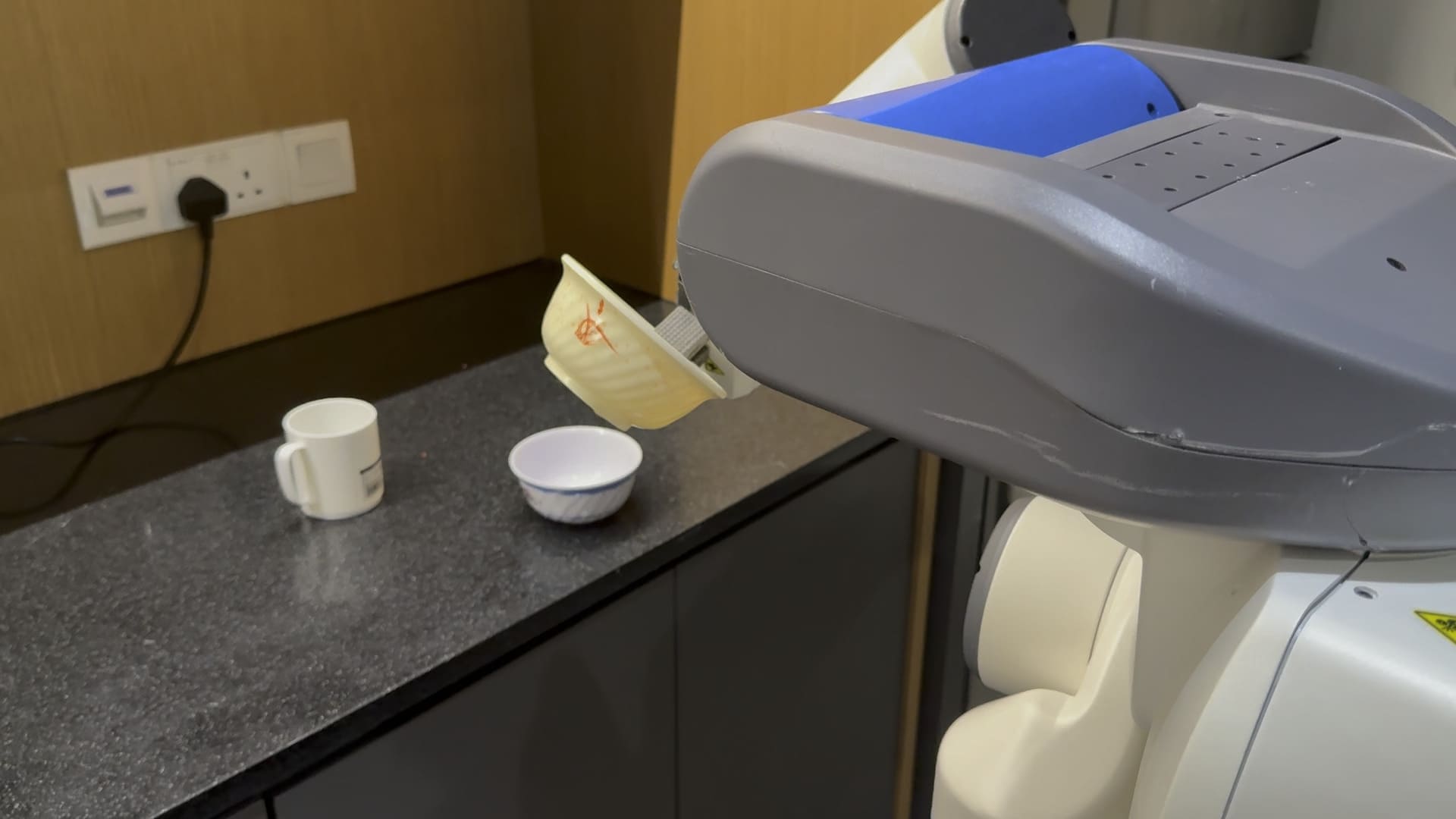} &
    \includegraphics[width=0.16\linewidth]{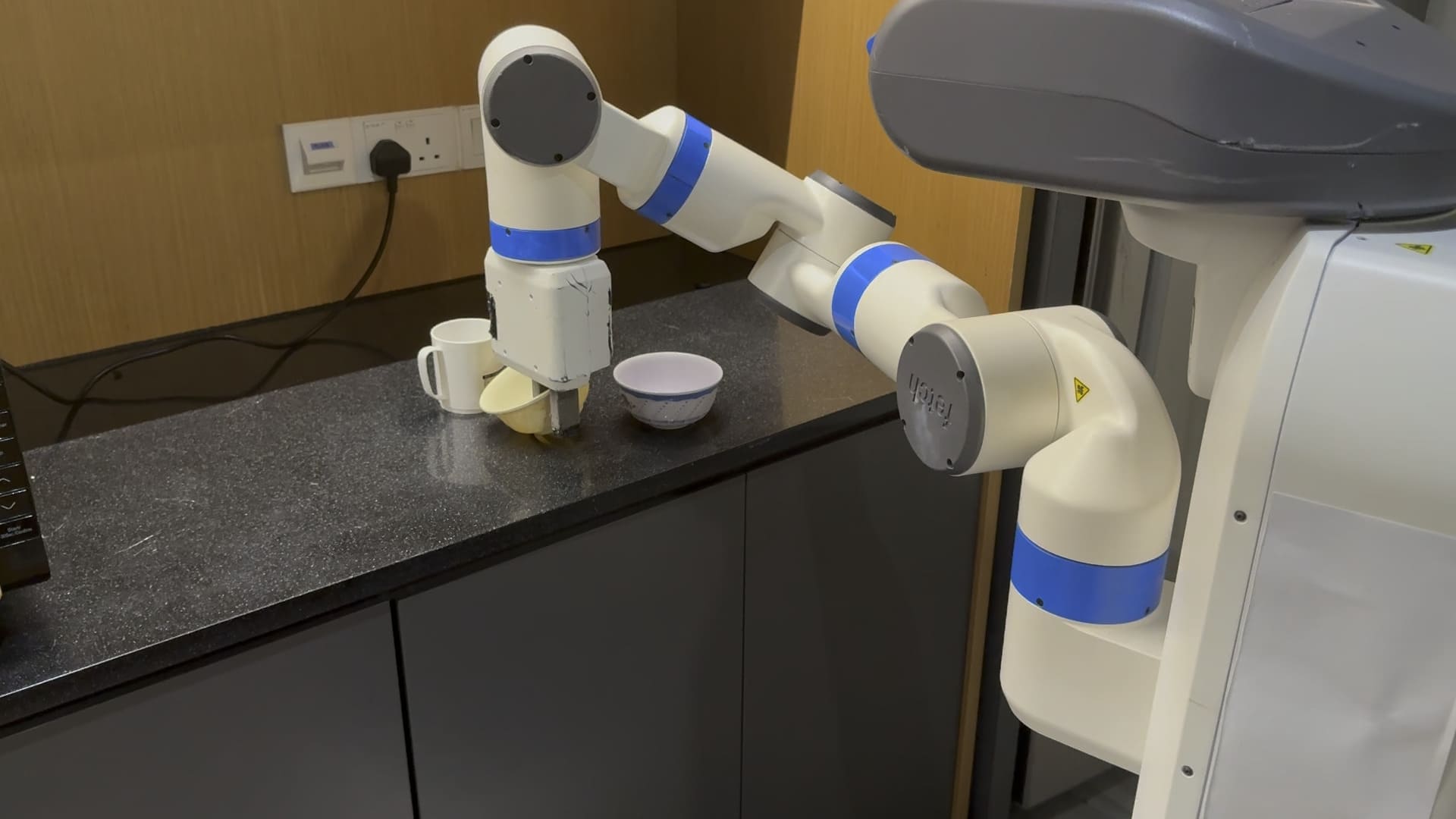} &
    \includegraphics[width=0.16\linewidth]{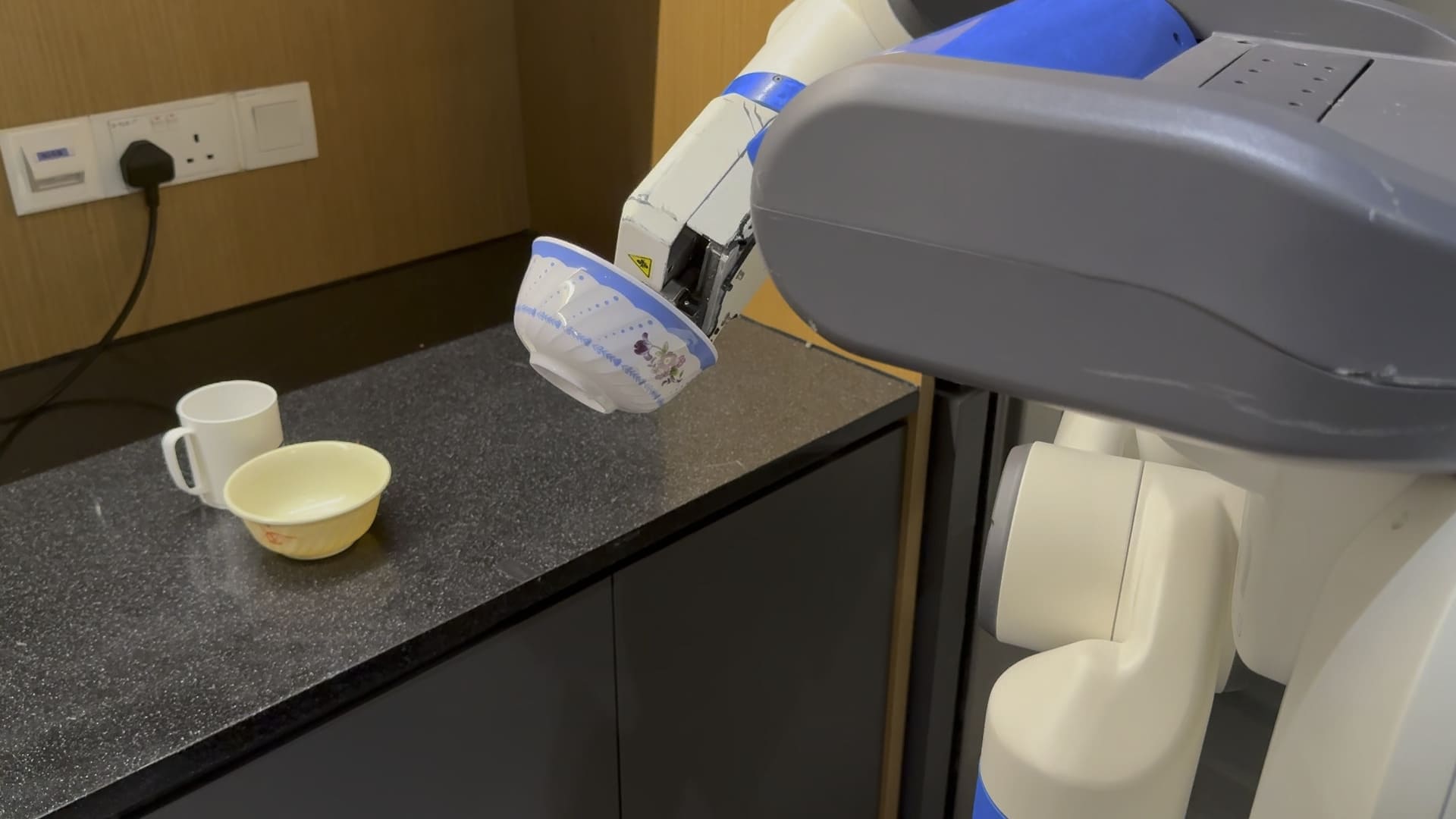} \\
    \includegraphics[width=0.16\linewidth]{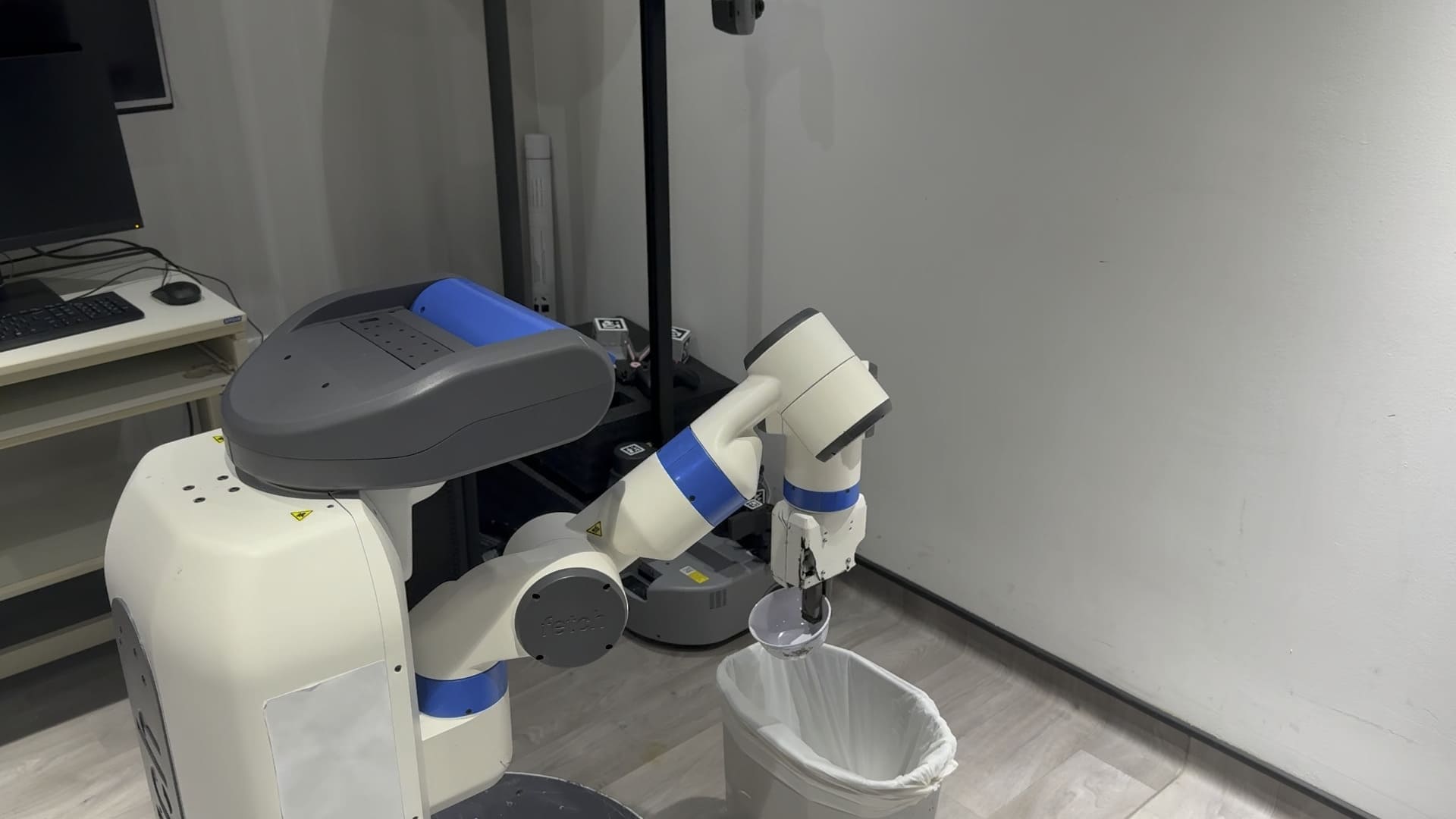} &
    \includegraphics[width=0.16\linewidth]{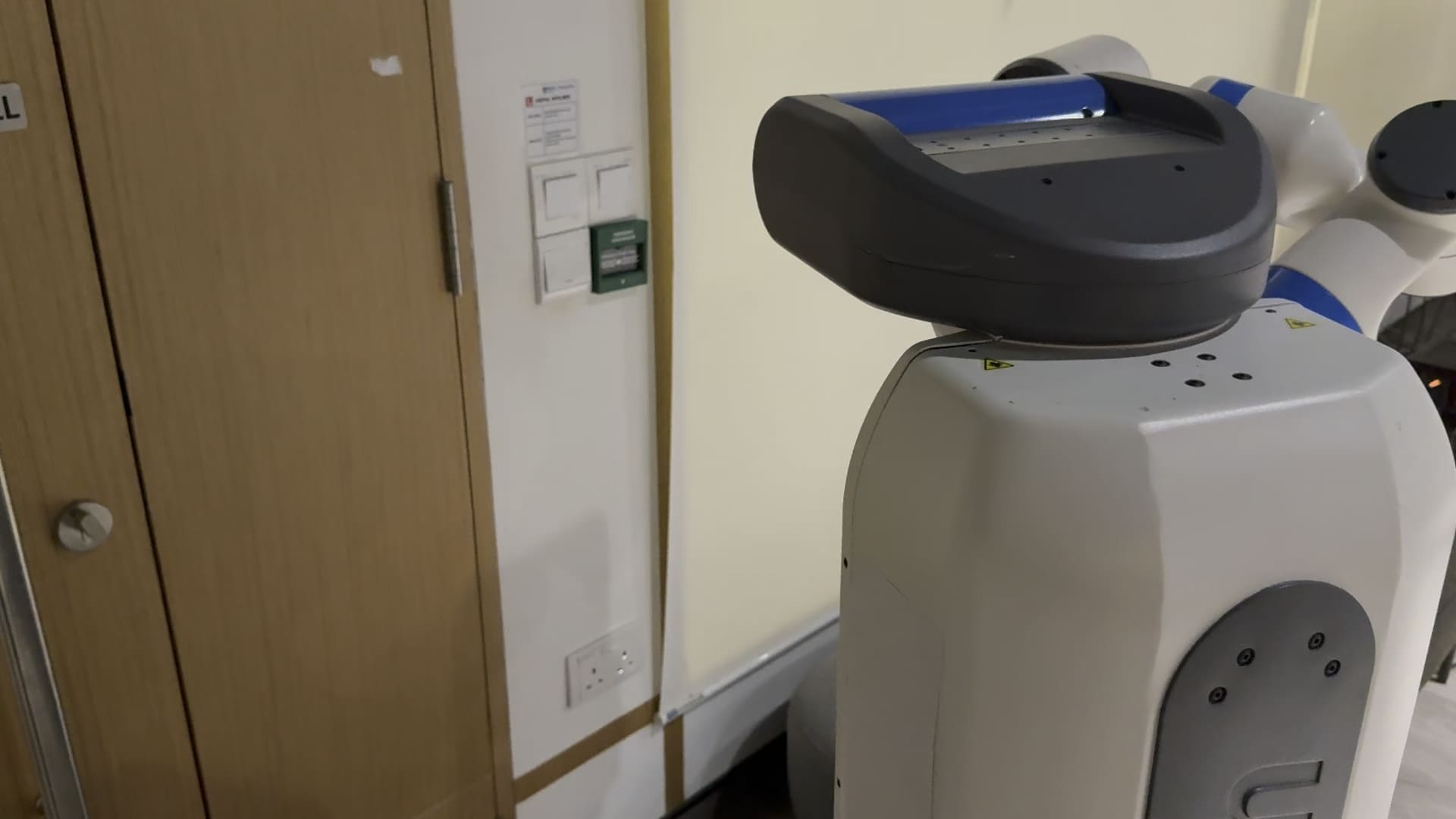} &
    \includegraphics[width=0.16\linewidth]{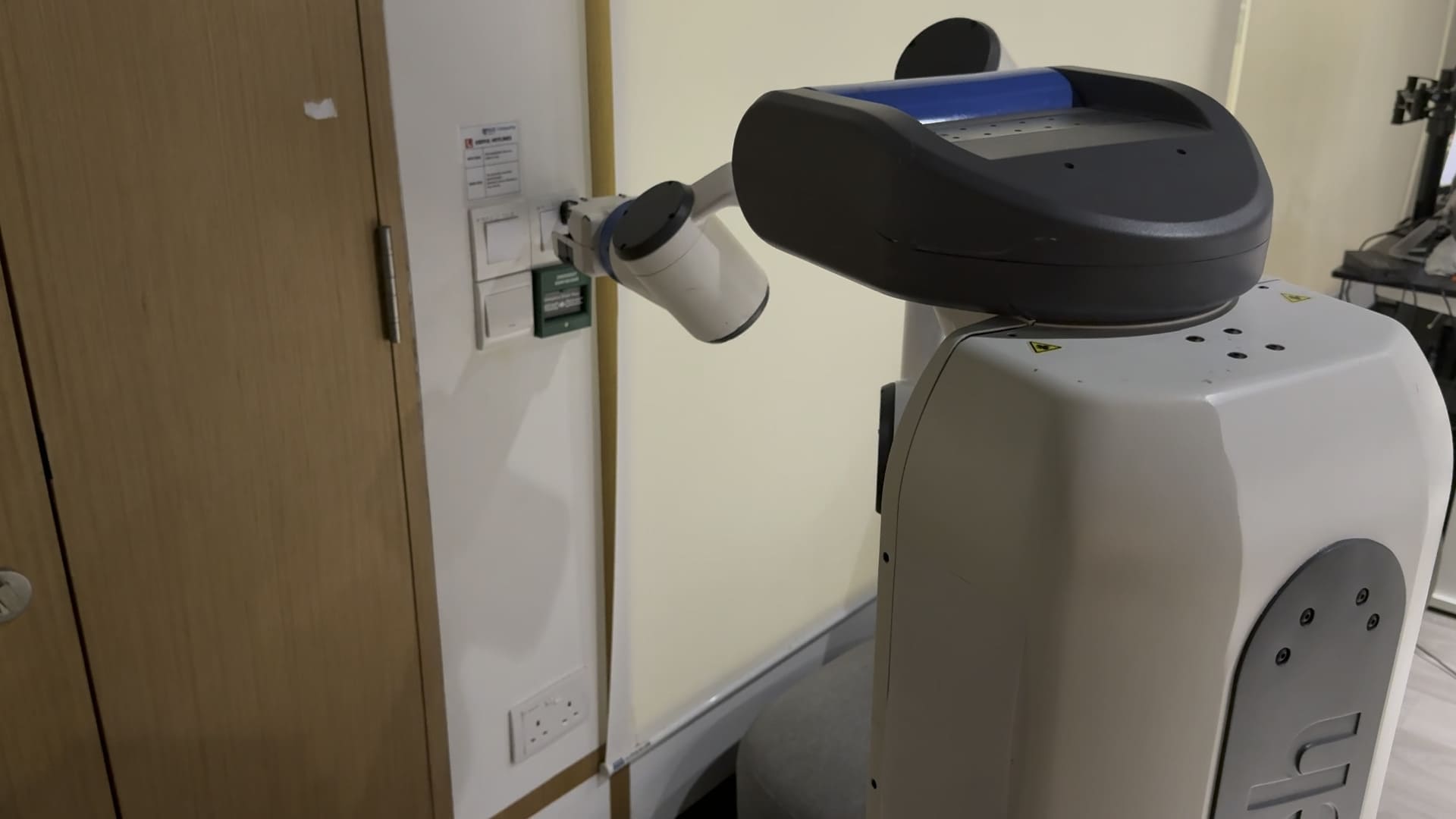} &
    \includegraphics[width=0.16\linewidth]{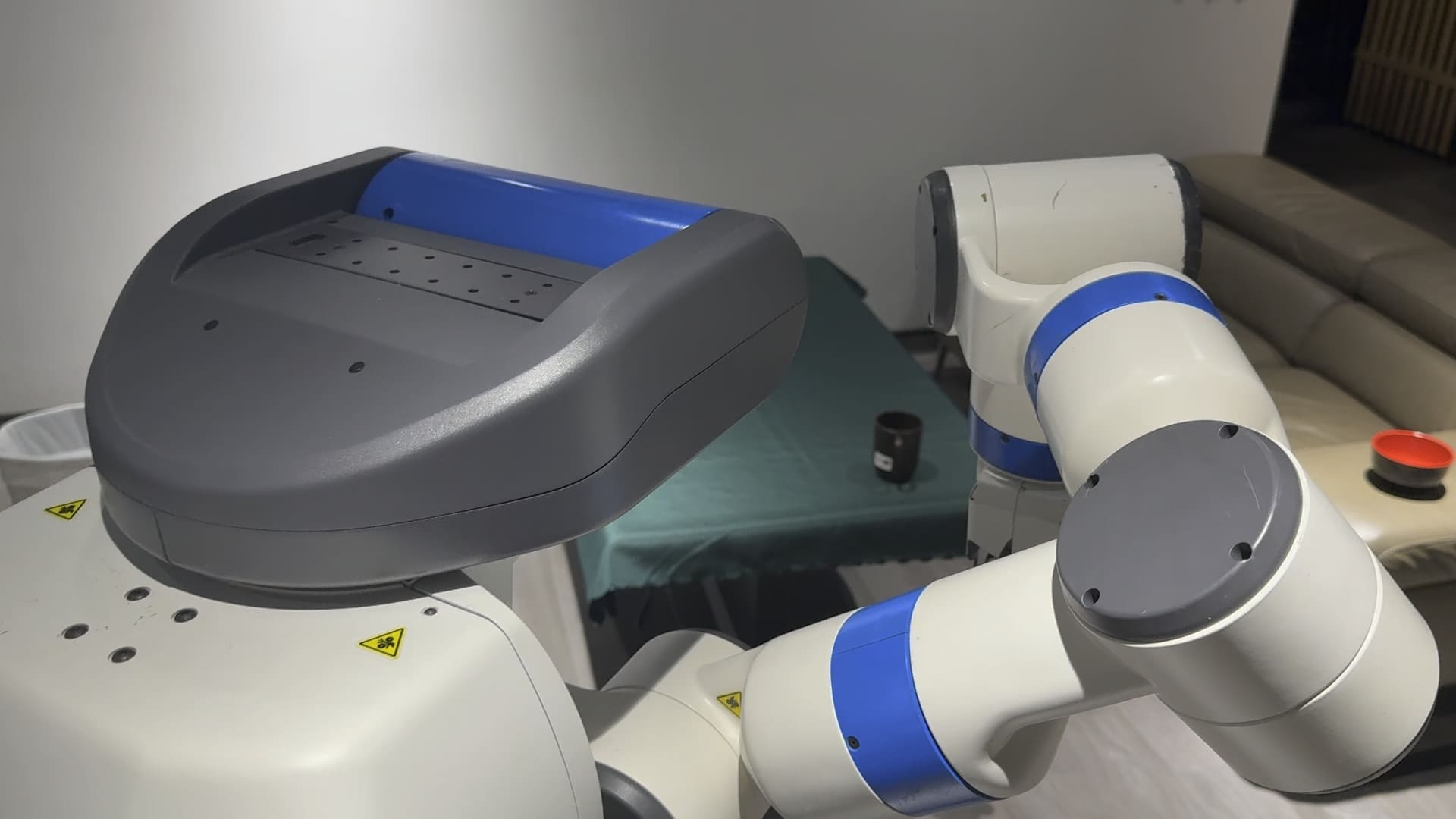} &
    \includegraphics[width=0.16\linewidth]{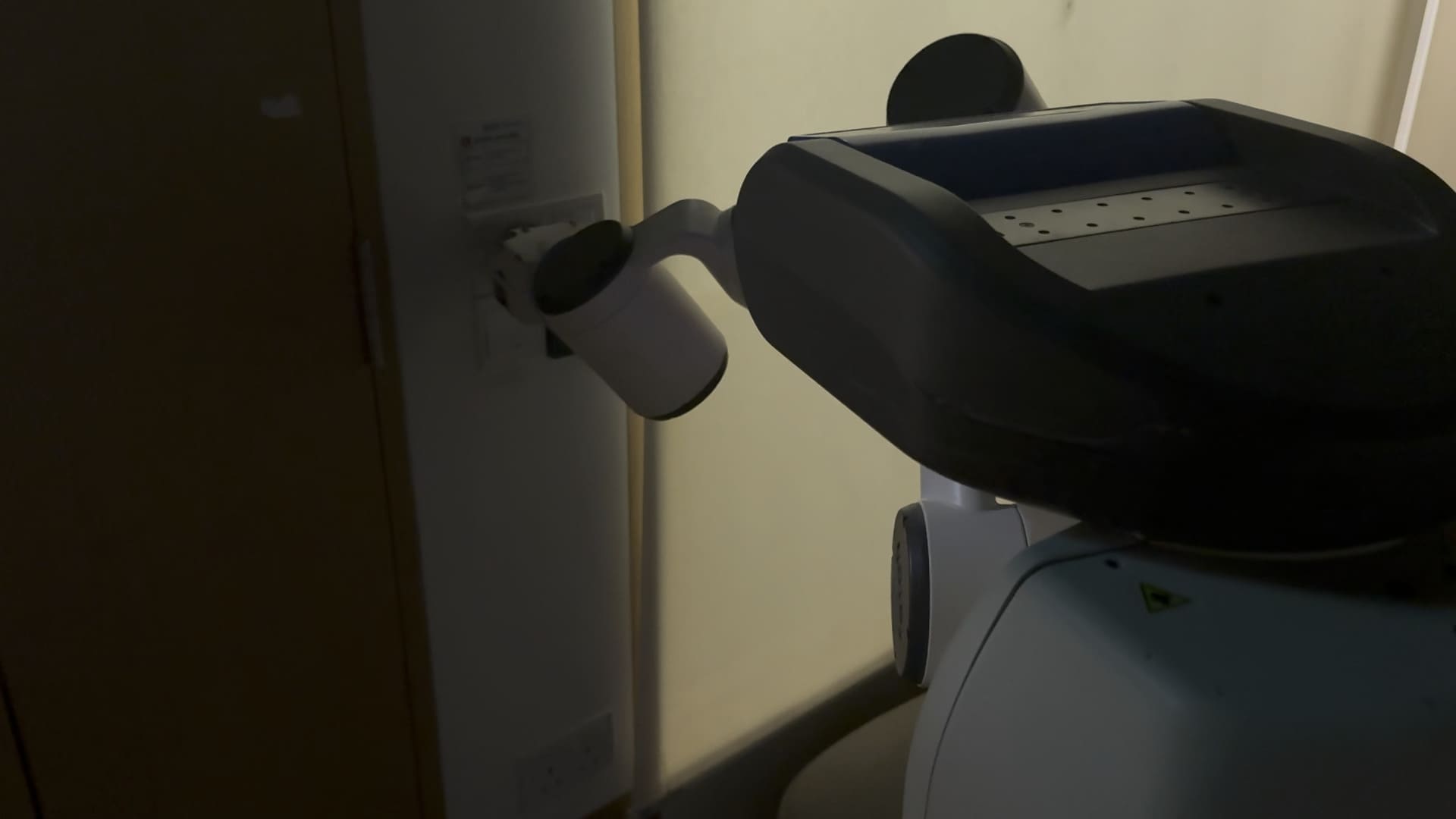} &
    \includegraphics[width=0.16\linewidth]{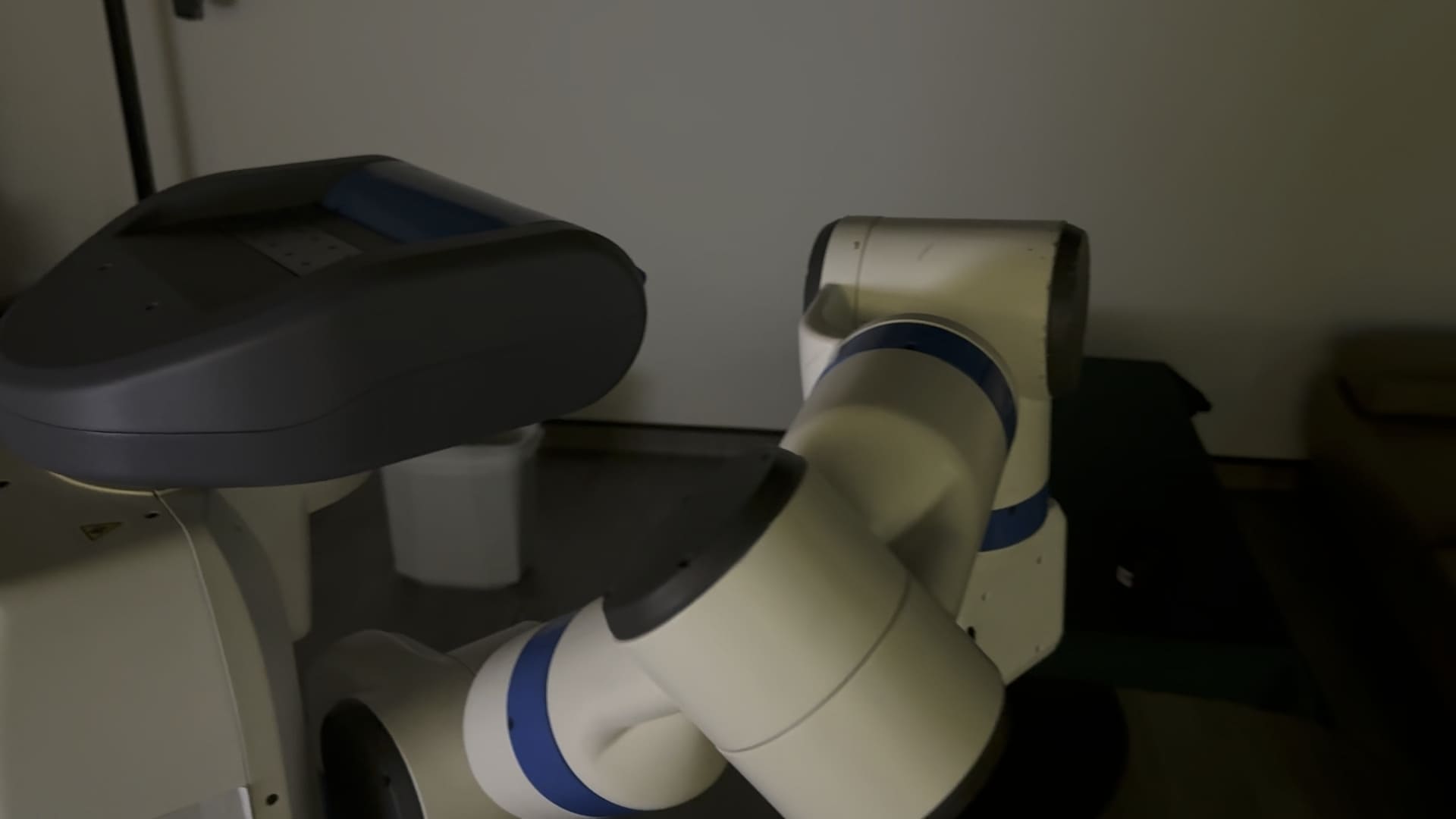} \\
  \end{tabular}
  \caption{Snapshot of completing the task \textbf{T5}: \textit{Throw away a blue-floral bowl and make the living room light-off.} The robot is required to locate the bowl, identify the blue-floral pattern, and determine how to turn off the living room light.}
  \label{fig:real_example_2}
  \vspace{-0.5cm}
\end{figure*}

\subsubsection{Result and Analysis}
As demonstrated in Fig. \ref{fig:mm_result_2}, the real-world experimental results show that uncertainty-aware model expansion significantly improves task performance for both LLM-based and formal planners. Compared to simulation, the real-world tasks more closely resemble the harder simulation settings (T3–T8), as they exhibit greater openness in object attributes due to the presence of distracting objects and longer task horizons. See Appendix for additional analysis.

Fig. \ref{fig:real_example_1} and Fig. \ref{fig:real_example_2} illustrate the execution of two real-world long-horizon household tasks using the proposed hypothesis-driven PDDL planner.
In Fig. \ref{fig:real_example_1} (\textit{T1: Deliver a zero-sugar drink to the table}), the robot first hypothesizes that a drink is inside the refrigerator and that it is zero-sugar. 
It navigates to the refrigerator, opens it, observes candidate drinks, and verifies object existence. After picking up a drink, attribute verification rejects the initial hypothesis, triggering hypothesis regeneration and replanning. The robot then explores the kitchen island, where both the existence and attribute hypotheses are verified. It takes the correct drink and delivers it to the table.
In Fig. \ref{fig:real_example_2} (\textit{T5: Throw away a flower-patterned bowl and turn off the living room light}), the robot first hypothesizes the bowl’s location and attribute, as well as a switch's location and its action effect. It searches the kitchen island and rejects the initial bowl-location hypothesis. After regenerating hypotheses and navigating to the kitchen counter, the robot observes two bowls, rejects the first through attribute verification, confirms the second, and discards it into the trash can. 
The robot moves to the panel area and successfully verifies the existence of switches. Based on its initial hypothesis, it presses the first switch, moves to the living room, and determines that the first switch does not control the living room light. It then updates its hypothesis, generates a new plan accordingly, and confirms that the second switch successfully turns off the living room light.
These examples highlight the importance of maintaining and verifying hypotheses in real-world tasks, and demonstrate the intelligent behavior enabled by the proposed mechanism. Further analysis is provided in the Appendix and video.

\subsection{Application in Appliance Operation}
\begin{figure}[t]
  \centering
  \includegraphics[width=0.95\linewidth]{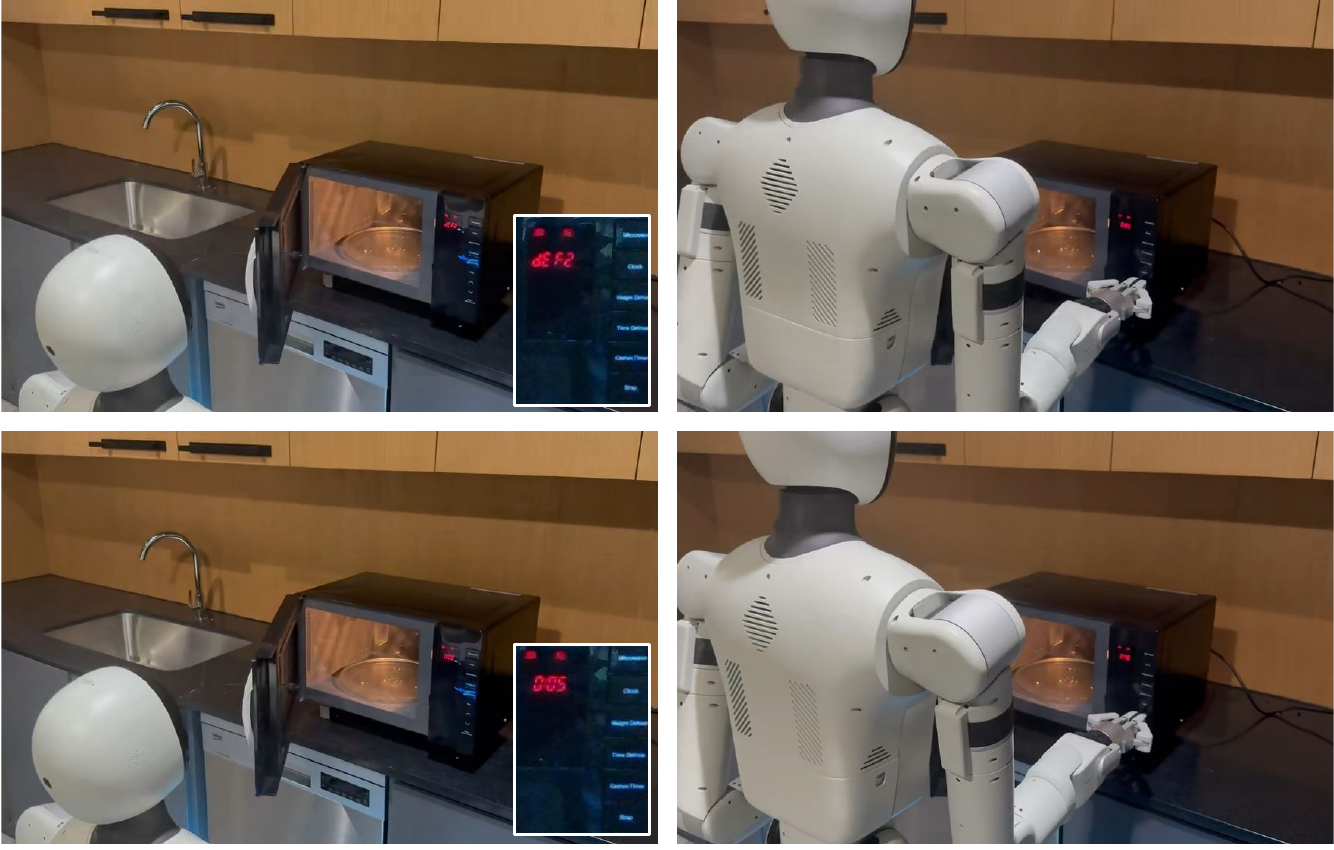}
  \vspace{-0.2cm}
  \caption{Demonstration of a robot discovering microwave button functions through interaction.
  }
  \label{fig:appliance_demo}
\vspace{-0.7cm}
\end{figure}

To demonstrate the generalizability of HUME to broader applications, we apply it to an appliance-operation task using the Autolife S2 humanoid robot. (see Appendix for the detailed setting).
In this setting, the robot is required to operate a microwave to \textit{defrost the food for 40 seconds}.
For the PDDL solver, since the FastDownard doesn't support numeric planning, we use the numeric planner called ENHSP \cite{scala2016heuristics} with Unified-Planning library \cite{unified_planning_softwarex2025}.
As shown in Fig. \ref{fig:appliance_demo}, the robot scans the buttons, matching known templates to corresponding PDDL operators, while the increase and decrease buttons lack predefined numerical effects.
The robot first hypothesizes a 10-second time increase from pressing the increase button in time-defrost mode and plans and acts accordingly.
Verification using the past two frames indicates that the time increases by 5 seconds, prompting the robot to revise its hypothesis. The updated hypothesis is then used for subsequent planning and execution without requiring further verification, successfully setting the time to 40s. Further details are provided in the Appendix and video.

\section{Discussion}
\label{sec:discussion}
Our experiments show that hypothesis-driven uncertainty-aware model expansion (HUME) offers a principled way to connect structured symbolic planning with the open-world knowledge of LLMs. Classical planning systems provide logical and geometric rigor but are brittle under incomplete models, while LLMs are flexible but unreliable as standalone planners. Our key insight is to use LLMs not to replace formal planning, but to propose uncertain model expansions whose validity is tested through planning and execution, allowing the planning system to manage uncertainty over preconditiaons, dependencies, and feasibility.

Despite its effectiveness, the proposed framework makes several simplifications that limit its generality and highlight important open challenges.

\textit{\textbf{a) Choice of uncertainty representation and planning under uncertainty.}}
The current system represents generated hypotheses as ternary beliefs: true, false, or unknown. Verification actions are treated optimistically during planning, with negative outcomes handled through replanning. This design supports tractable, real-time planning, but limits the system’s ability to reason about graded confidence or risk. We adopt it because foundation models do not yet provide reliably calibrated probabilities over generated knowledge, and richer uncertainty-aware planning remains computationally challenging in long-horizon settings. Consequently, the system may behave overly optimistically when risk-sensitive decisions are required. In future work, we aim to integrate uncertainty calibration tools \cite{liu2025uncertainty} into the model expansion process and explore more advanced online POMDP solvers \cite{hoerger2025vectorized,jin2026vec} for risk-aware planning under uncertainty.

\textit{\textbf{b) Hypothesis verification conditions.}}
We assume that language models can propose verification conditions sufficient to determine a hypothesis and that these conditions can be enforced by a formal planner. However, some real-world verification conditions depend on geometry, long-horizon interactions, or complex perception, which may be difficult to infer from language alone. In such complex cases, deciding how to verify a hypothesis itself becomes an active perception or planning problem. Future work will extend the current approach by integrating language-predicted verification conditions with geometry- and perception-aware reasoning, making hypothesis verification more suitable for integrated task and motion planning (TAMP) settings \cite{garrett2021integrated, curtis2024partially, shen2026tiptop}.

\textit{\textbf{c) Source of abstraction.}}
Our current implementation requires the task instruction to define the abstraction of missing knowledge. This means that missing object concepts and predicates are explicitly encoded in the task goals. We make this assumption to ensure methodological completeness and to keep hypothesis generation task-oriented. In future work, we aim to explore failure reasoning modules \cite{liu2023reflect} as an additional source of abstraction, using execution feedback to identify missing concepts and formulate new hypotheses that are not directly specified by the task goals.

\textit{\textbf{d) Open-ended exploration in open-world settings.}}
We avoid allowing the LLM to terminate hypothesis generation when it deems a task impossible, in order to prevent overly pessimistic early termination. Our task scope assumes that the task is achievable in the underlying environment and focuses on optimistically uncovering missing knowledge. In practice, termination is bounded by the maximum number of hypothesis-generation attempts and replanning steps. More broadly, proving task impossibility in open-world settings is generally intractable, a challenge shared across the open-world planning literature. Extending the framework to incorporate impossibility reasoning, such as belief-space pruning or contradiction detection, is an interesting direction for future work.

\section{Conclusion}

We presented a hypothesis-driven framework for open-world robot planning that treats model expansion as an explicit and uncertain process integrated into goal-directed planning. By actively generating and verifying hypotheses, the proposed approach enables robots to operate effectively under incomplete and evolving world models. Experiments in simulation and the real world show that uncertainty-aware model expansion significantly improves robustness and success for both symbolic and LLM-based planners. Our results highlight the importance of integrating uncertainty-aware model expansion into planning systems for deploying service robots in real open-world environments, while more fine-grained uncertainty estimation and tighter integration with advanced uncertainty-aware planning algorithms remain important directions for future work. Additional discussion of insights, limitations, and future work is provided in the appendix.

\section*{Acknowledgments}
This research is supported in part by Singapore Ministry of Digital Information under its AIVP program (grant no. A-8004133-00-00). Anxing Xiao is supported by the Research Scholarship from the National University of Singapore. We thank Chao Tang, Jian Zhang, Siwei Chen, and Panpan Cai for helpful discussions and suggestions on an earlier version of this work.


\bibliographystyle{plainnat}
\bibliography{references}

\clearpage

\onecolumn

\section*{Appendix Content}
\addcontentsline{toc}{section}{Appendix}

\setcounter{section}{0}
\setcounter{subsection}{0}
\setcounter{subsubsection}{0}

\renewcommand{\thesection}{A.\Roman{section}}
\renewcommand{\thesubsection}{\thesection-\Alph{subsection}}
\renewcommand{\thesubsubsection}{\thesubsection\arabic{subsubsection}}

\renewcommand{\theHsection}{appendix.A.\Roman{section}}
\renewcommand{\theHsubsection}{appendix.A.\Roman{section}.\Alph{subsection}}
\renewcommand{\theHsubsubsection}{appendix.A.\Roman{section}.\Alph{subsection}.\arabic{subsubsection}}

\startcontents[appendix]
\renewcommand{\contentsname}{Appendix Contents}
\printcontents[appendix]{}{1}{}
\clearpage

\lstset{
  basicstyle=\ttfamily\footnotesize,
  keywordstyle=\ttfamily,  
  columns=fixed, 
  keepspaces=true,
  showstringspaces=false,
  frame=none,        
  numbers=none,
  tabsize=2,
  breaklines=true,
  escapeinside={(*@}{@*)},  
}

\lstdefinestyle{pydata}{
  basicstyle=\ttfamily\footnotesize,
  columns=fixed,
  keepspaces=true,
  showstringspaces=false,
  frame=none,
  numbers=none,
  breaklines=true,
  backgroundcolor=\color{gray!10},
  xleftmargin=2pt,
  xrightmargin=2pt,
  aboveskip=4pt,
  belowskip=4pt,
  emph={id,type,description,content,condition,verification_condition},
  emphstyle=\bfseries,
  emph={[2]add_fact, add_effect},
  emphstyle={[2]\itshape},
  emph={[3]None},
  emphstyle={[3]\color{gray}}
}

\section{Experimental Setup Details}

\subsection{Real-world experimental setup}
As we explain in the main paper, we employed a Fetch mobile manipulator in a household environment. In this section, we introduce the details of the environments, the tasks, and the hardware and software infrastructure that enable our hypothesis-driven planning framework.

\begin{figure}[h]
    \centering
    \includegraphics[width=1\textwidth]{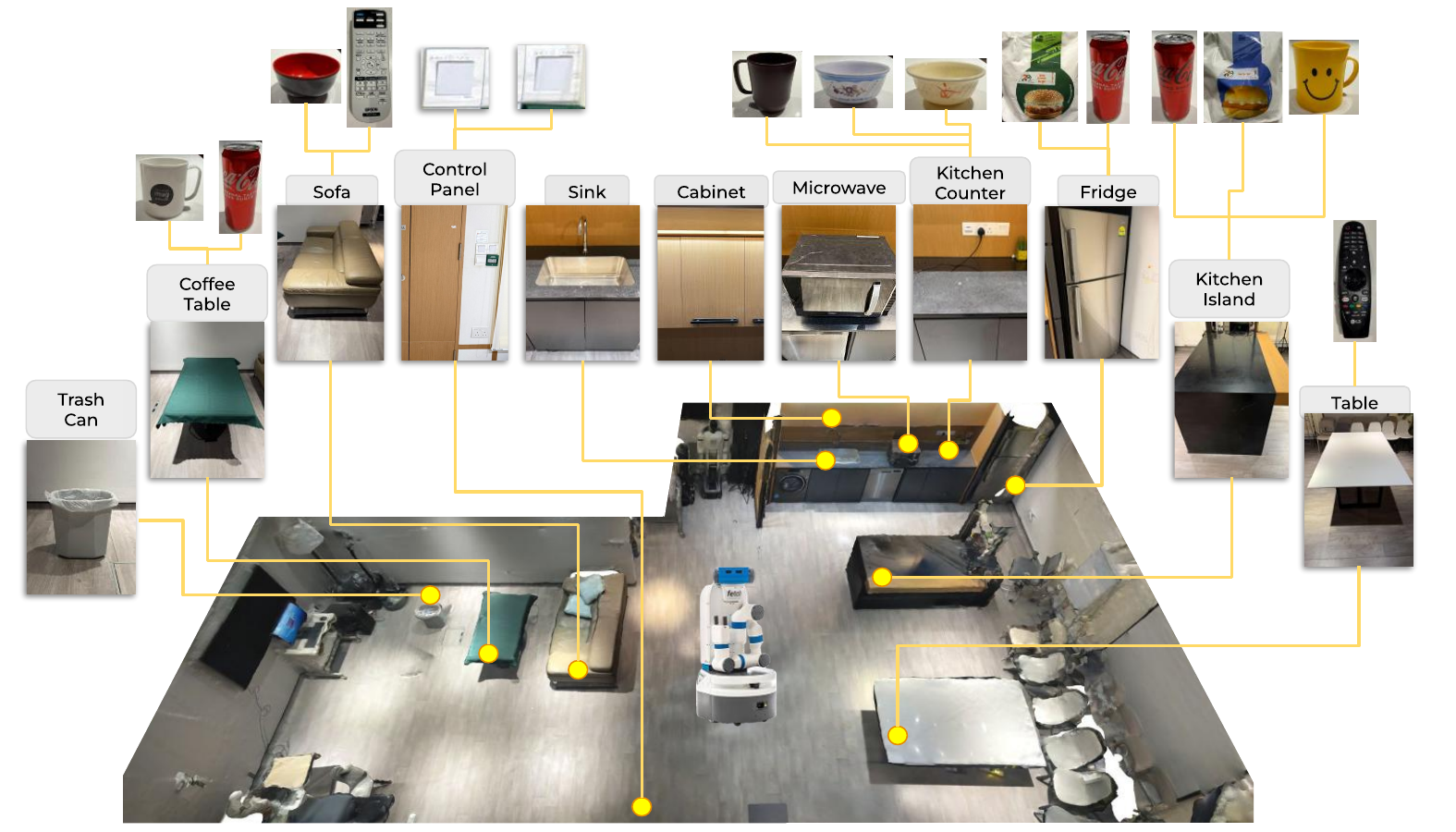}
    \caption{Real-world experimental setup}
    \label{fig:rw_env}
\end{figure}

\subsubsection{Environments and Tasks}
Our real-robot experiments were conducted in a real household environment comprising three rooms: a kitchen, a living room, and a meeting room.
Figure~\ref{fig:rw_env} illustrates a top-down view of the experimental setup. The kitchen, located at the top right and serving as the primary workspace, is equipped with several functional areas, including a sink, cabinet, microwave, refrigerator, kitchen counter, and kitchen island. The living room, located on the left, contains a trash can, coffee table, sofa, and control panel. The meeting room, located on the bottom right, includes a meeting table.

\paragraph{Detailed Task Definitions}
We evaluated the system on five distinct long-horizon tasks. Across all tasks, the robot is provided only with a map of static locations, corresponding to large furniture and appliances. Dynamic objects are not included in the initial scene graph. The robot can detect objects only when it is at the same location as them, and the location must be open. The robot is equipped with a set of primitive skills, including \texttt{navigate(loc)}, \texttt{pick(obj)}, \texttt{place(obj, loc)}, \texttt{open(loc)}, \texttt{close(loc)}, and \texttt{trigger(obj/loc)}.
In addition, the robot is only equipped with a single manipulator; therefore, when holding an object, it is limited to navigation and placement actions and cannot perform other manipulations.

\noindent \textbf{T1. Deliver a zero-sugar drink to the table.} \\
The robot is required to retrieve a specific beverage and place it on the table, where the target is defined by a fine-grained attribute ``zero-sugar'' that is typically indicated by small text or subtle logo variations. In our setup, three visually similar beverage cans are distributed across the environment: one in the refrigerator, one on the coffee table, and one on the kitchen island. Among them, only the can located on the kitchen island corresponds to the target zero-sugar variant, while the others are standard versions.

\noindent \textbf{T2. Place the remote with a red button into the cabinet.} \\
The robot is required to store a remote control (specified by a fine-grained visual attribute ``red button'') inside the cabinet. There are two remote controls in the environment, one with a red button located on the meeting table and one with no red button located on the sofa. This requires the robot to plan an action sequence to finish the long-horizon tasks while actively search and visually disambiguating the target based on subtle appearance cues.

\noindent \textbf{T3. Move the smiley-face mug to the fridge.} \\
The robot must move a mug with a specific visual pattern ``smiley face'' inside the fridge. Several mugs are placed at different locations in the environment: one at the kitchen counter, one on the kitchen island, and one on the coffee table. Among them, only the mug located on the kitchen island corresponds to the target smiley-face variant.

\noindent \textbf{T4. Serve a heated chicken burger on the coffee table.} \\
The robot is required to prepare and serve a heated chicken burger on the coffee table. Two burgers are present in the environment: one fish burger located in the refrigerator and one chicken burger located on the kitchen island. The target burger must additionally undergo a state change (``heated'') before delivery.
This task requires long-horizon planning as well as the integration of multiple reasoning components, including locating a burger, identifying its type, and determining how to heat it prior to placement.

\noindent \textbf{T5. Throw away the blue-floral bowl and turn off the living room light.} \\
The robot must throw away a bowl with a specific “blue-floral” pattern into the trash can. Additionally, the robot is required to make living room light off. In the environment, two bowls are located on the kitchen counter, only one of which has the blue-floral pattern. The control panel contains two light switches: one controlling the kitchen–meeting room area and the other controlling the living room. This task combines object-level attribute grounding with learning the state transitions associated with triggering different switches.

\subsubsection{Real-World Robot System}
\paragraph{Hardware Setup}
We adopt the commercially available mobile manipulator Fetch \cite{wise2016fetch}. Its base is a differential-drive mobile platform, and it is equipped with a 1-DOF torso, a 7-DOF arm with parallel-jaw gripper. The robot also includes an RGB-D camera mounted on its head and a LiDAR sensor on the base for localization.
Given the low resolution of the Fetch robot’s RGB-D camera, reliable identification of the target object’s attributes requires that different sides of the object be positioned directly in front of the camera due to hardware limitations, which guides the design of our software system. All non-GPU computations are performed on a local workstation connected to the robot via Wi-Fi, while computationally intensive GPU workloads are executed on an H100 server cluster.

\paragraph{Software Stack and Architecture}
The real-world robotic system adopts a modular, layered software architecture that integrates symbolic reasoning, reactive control, and learning-based components. At the core is a structured state representation that maintains a 3D scene graph including locations, objects, and robot status. Robot skills are represented as a library of reusable Behavior Trees (BTs) \cite{ogren2022behavior}, each corresponding to a PDDL-style domain operator \cite{fox2003pddl2} with explicitly defined general preconditions and state transition effects. 
These BTs support modular composition and reactive execution, enabling robust handling of changing environments.
Overall, the architecture bridges symbolic task reasoning with continuous perception and control, supporting scalable, robust, and adaptive behavior in open-world household settings.

\begin{itemize}
    
    \item \textbf{Perception}: The robot is initialized with a pre-scanned point cloud of the environment for collision avoidance. For the small objects, the perception pipeline involves a modular approach that includes object detection, segmentation, and point cloud registration. When the robot move to a new location or opens a closed location, it triggers the perception pipeline, the RGB image is send to OWLv2 \cite{minderer2024scaling} for detecting objects based on a set of finite object types modified from the vocabulary set \cite{lin2014microsoft}. The Segment Anything model \cite{kirillov2023segment} is used to generate the object mask for all detected objects, combined with Depth observations, to get a 3D back-projection point cloud of each object, and transfer to the global coordinate system. Object re-identification is performed using a simple nearest-neighbor strategy, where each object is matched to the closest existing point cloud of the same object category.

    \item \textbf{Navigation}: Our system integrates both predefined navigation waypoints and dynamically sampled navigation poses associated with movable objects. For localization and navigation planning, we construct an occupancy grid map using Gmapping \cite{grisetti2007improved}. Motion planning and execution are handled by the off-the-shelf MoveBase package from the ROS Navigation Stack \cite{guimaraes2016ros}.
    
    \begin{figure}[h]
    \centering
    \includegraphics[width=1\textwidth]{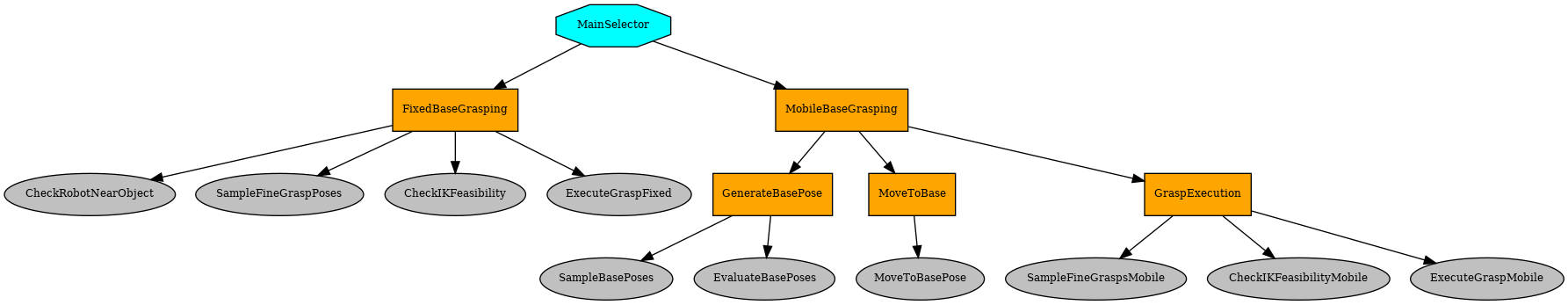}
    \vspace{-0.6cm}
    \caption{The behavior tree for picking up an object}
    \label{fig:pick}
    \vspace{-0.4cm}
\end{figure}

    \item \textbf{Model-based Primitive Skills}: 
The robot implements a set of model-based skills, including navigation, pick, place, and trigger actions. All skills are represented using Behavior Trees (BTs) \cite{ogren2022behavior}. Figure~\ref{fig:pick} illustrates the behavior tree for picking up a perceived object.
The pick skill first attempts fixed-base grasping at the current robot configuration. If this attempt fails, the system triggers a mobile grasping procedure. In this procedure, six candidate base poses are sampled using an inverse reachability map \cite{vahrenkamp2013robot} precomputed in an obstacle-free environment. For each candidate base pose, five pre-grasp poses are evaluated, corresponding to different approach directions toward the object (one top-down grasp and four side grasps). For each pre-grasp, up to 20 inverse kinematics (IK) solutions are sampled and checked for collision-free feasibility.
This process is used to evaluate the quality of each candidate base configuration, measured by the number of approach directions that admit at least one valid pregrasp IK solution. 
The base pose with the highest grasp feasibility score is selected as the navigation goal. After the robot navigates to the selected base pose, the target object is re-perceived, and the fixed-base grasping pipeline is executed. For collision checking and fast motion planning, we adopt the VAMP toolkit \cite{vamp_2024}. For the remaining behaviors, we adopt a design similar to the pick skill. For \texttt{pick} and \texttt{place} actions involving constrained containers, as well as \texttt{trigger} actions, the sampling of pre-poses is guided by simple heuristics that encourage alignment with the robot’s facing direction.

    \item \textbf{Imitation Learning–Based Primitive Skills}: 
For complex primitive skills such as \texttt{open} and \texttt{close}, we employ imitation learning to execute complex manipulation actions. We collect an average of 50 human-teleoperated demonstrations per action on a real robot using a VR controller, covering tasks such as opening and closing cabinets, fridges, and microwaves. These demonstrations are used to post-train the $\pi_{0.5}$ model \cite{intelligence2504pi0} to acquire the corresponding skills. The model takes RGB observations and the robot arm’s states as inputs and predicts joint angle movement sequences along with gripper actions. Before executing the skill, the robot will move to predefined base pose and arm configuration.

\begin{figure}[h]
    \centering
    \includegraphics[width=1\textwidth]{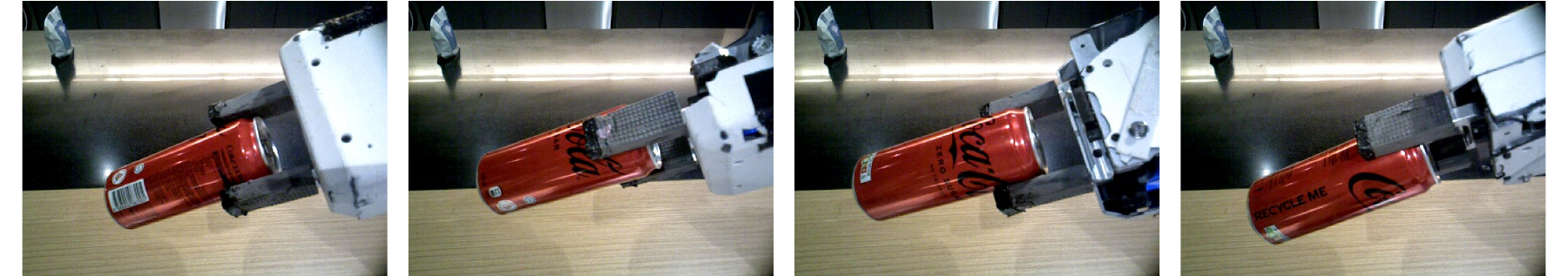}
    \vspace{-0.6cm}
    \caption{The four different views of the in-hand object}
    \label{fig:images}
    \vspace{-0.4cm}
\end{figure}

    \item \textbf{Image Capture for Holding Object}: 
As described in the hardware setup, the robot has limited sensory capabilities. In particular, it can only obtain clear visual observations of an object when the object is positioned very close to the head-mounted camera. To enable hypothesis verification of object attributes under this constraint, we adopt an in-hand observation pipeline that captures four images of the object from different viewpoints. As illustrated in Fig. \ref{fig:images}, the robot first holds the object directly in front of the head camera and captures an image. It then rotates its wrist by 90 degrees and acquires a second image. This process is repeated until four images of the in-hand object are obtained from distinct viewing angles.

    \item \textbf{Domain Knowledge for The System}: 
For each primitive skill, we have a PDDL operator define the precondition and postcondition for the skill. 
Table~\ref{tab:pddl_actions} details the preconditions and effects of the primitive actions. For more details, please check Section \ref{sec:pddl}.

\begin{table*}[h]
    \centering
    \caption{Action, Preconditions and Effects in Our System.}
    \begin{tabular}{>{\raggedright\arraybackslash}p{0.2\linewidth} >{\raggedright\arraybackslash}p{0.35\linewidth} >{\raggedright\arraybackslash}p{0.35\linewidth}}
        \toprule
        \textbf{Action} & \textbf{Preconditions} & \textbf{Effects} \\
        \midrule
        \texttt{navigate(?r, ?l1, ?l2)} & 
        \texttt{Robot(?r)}, \texttt{At(?r,?l1)} & 
        \texttt{At(?r,?l2)}, $\neg$\texttt{At(?r,?l1)} \\
        \midrule
        \texttt{pick(?r, ?o, ?l)} & 
        \texttt{HandEmpty(?r)}, \texttt{At(?r,?l)}, \texttt{At(?o,?l)}, \texttt{IsOpen(?l)} & 
        \texttt{Holding(?r,?o)}, $\neg$\texttt{HandEmpty(?r)}, $\neg$\texttt{At(?o,?l)} \\
        \midrule
        \texttt{place(?r, ?o, ?l)} & 
        \texttt{At(?r,?l)}, \texttt{Holding(?r,?o)}, \texttt{IsOpen(?l)}, $\neg$\texttt{Room(?l)} & 
        \texttt{HandEmpty(?r)}, \texttt{At(?o,?l)}, $\neg$\texttt{Holding(?r,?o)} \\
        \midrule
        \texttt{open(?r, ?l)} & 
        \texttt{At(?r,?l)}, \texttt{HandEmpty(?r)}, $\neg$\texttt{IsOpen(?l)}, $\neg$\texttt{Room(?l)} & 
        \texttt{IsOpen(?l)} \\
        \midrule
        \texttt{close(?r, ?l)} & 
        \texttt{At(?r,?l)}, \texttt{HandEmpty(?r)}, \texttt{IsOpen(?l)}, $\neg$\texttt{Room(?l)} & 
        $\neg$\texttt{IsOpen(?l)} \\
        \midrule
        \texttt{trigger(?r, ?l, ?o)} & 
        \texttt{At(?r,?l)}, \texttt{At(?o,?l)}, \texttt{HandEmpty(?r)}, $\neg$\texttt{Room(?l)} & 
        \texttt{Trigged(?l)} \\
        \bottomrule
    \end{tabular}
    \label{tab:pddl_actions}
\end{table*} 

\end{itemize}

\subsection{Simulation experimental setup}
In this section, we explain more details about the simulation experimental setup. 

\subsubsection{Block Processing World}
The task is to achieve a target configuration in which certain objects exhibit specified effects, corresponding to preparing ingredients in the correct states for cooking.
To simulate imperfect real-world recognition, labels are removed from a randomly selected half of the processors. The environment contains 3--8 blocks and 3--8 processors. The robot must plan while considering exploration of the environment and interact with the processors to discover their effects.

\paragraph{Setting and Task}
The Block Processing World extends the standard Block World with multiple block processors, each containing a region and a trigger button, analogous to different settings on kitchen appliances.
Activating a processor induces latent effects on the block placed within its region. A robot new to this “kitchen” is unaware of these effects and can only observe the block’s state in its hand. In our setting, we model ground-truth processor types and their corresponding state-change effects, encompassing common kitchen functions:
\noindent \textit{Stove} $\rightarrow$ \texttt{hot}, \textit{Freezer} $\rightarrow$ \texttt{frozen}, \textit{Faucet} $\rightarrow$ \texttt{wet}, \textit{Steamer} $\rightarrow$ \texttt{steamed}, \textit{Oven} $\rightarrow$ \texttt{baked}, \textit{Washing Machine} $\rightarrow$ \texttt{cleaned}, \textit{Toaster} $\rightarrow$ \texttt{toasted}, \textit{Pan} $\rightarrow$ \texttt{fried}, \textit{Kettle} $\rightarrow$ \texttt{boiled}, and \textit{Coffee Maker} $\rightarrow$ \texttt{brewed}.
The task is to achieve a target configuration in which certain objects exhibit specified effects, corresponding to preparing ingredients in the correct states for cooking.
An example goal is defined as a conjunction of spatial and state constraints, e.g.,
\texttt{(and (on e c) (on a r\_coffee\_maker) (on d r\_1) (ontable d) (hot c) (toasted e) (cleaned b))}.

\begin{figure}[h]
    \centering
    \includegraphics[width=1\textwidth]{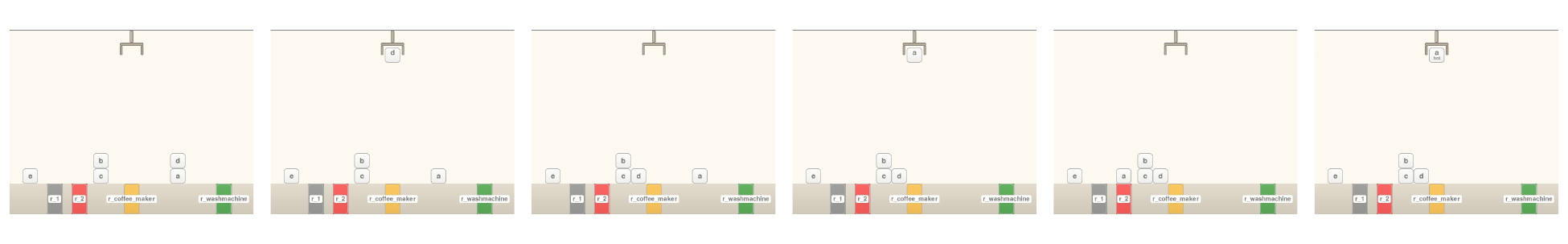}
    \vspace{-0.8cm}
    \caption{The visualization of the Block Processing World, triggering a processor will change the state of the block on the processor, and the robot can only observe the state of the block in its hand. In this example, triggering r\_2 (unrecognized type) make the block a ``\texttt{hot}''}
    \label{fig:blocks}
    \vspace{-0.2cm}
\end{figure}

\paragraph{Experimental Setup}
We generated problem instances with varying levels of complexity, involving 3–8 blocks and 3–8 locations. For each episode, both the initial configuration and the goal specification are randomly sampled 10 times. To simulate real-world appliance/button OCR recognition failures, the names of half of the processors are removed and replaced with numerical indices. All planners operate in a closed-loop manner within the environment, enabling them to react to execution failures and proactively replan when necessary. Details of the PDDL domain file and the prompts used across different methods are provided in Section~\ref{sec:appendix_prompts}.

\paragraph{Real-World Significance}
This setup abstracts the causal structure of real-world action–object interactions, such as using different appliances or triggering different buttons. It enables rigorous evaluation of logical reasoning and hypothesis generation in a controlled setting, without confounding factors from other sources of noise.

\subsubsection{Mobile Manipulation in Unknown Environment (AI2THOR)}
In this setting, we consider the senario of a mobile manipulator operating in a household-like environment with objects of unknown locations and attributes.

\begin{figure}[h]
    \centering
    \includegraphics[width=0.9\textwidth]{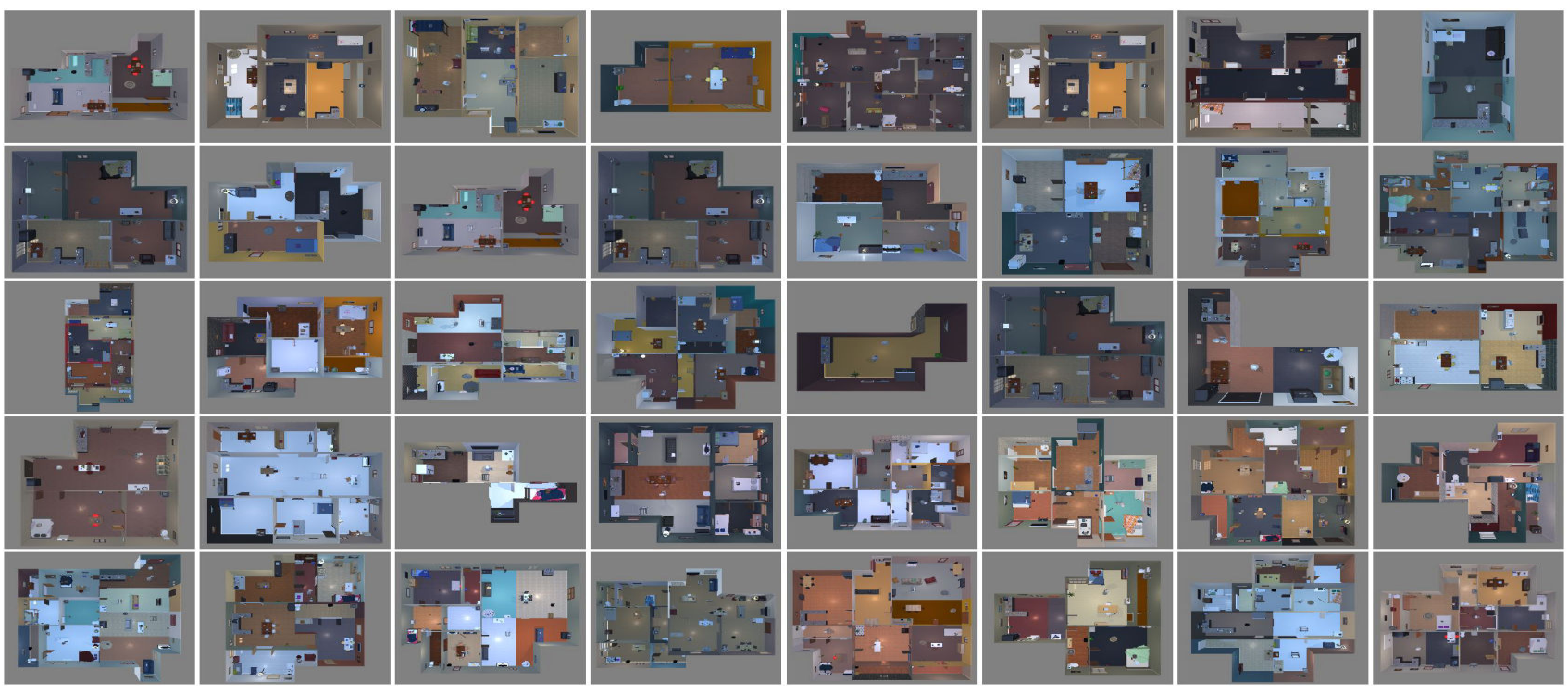}
    \vspace{-0.2cm}
    \caption{Visualizations of all ProcTHOR house environments used in our experiments.}
    \label{fig:visai2thor}
    \vspace{-0.2cm}
\end{figure}

\paragraph{Setting and Task}
The robot is tasked with rearranging objects to achieve a target configuration, such as delivering an object with a specific attribute to a designated location. Because object locations, properties, and action effects are initially unknown, the robot must actively explore and interact with the environment to infer object attributes, discover hidden objects, and predict action outcomes. We consider eight tasks spanning different levels of openness and complexity, as summarized below:

\begin{enumerate}[noitemsep, label=\textbf{T\arabic*.}, align=left, labelwidth=0.5cm]
    \item \textit{Store the remote control in a drawer.}
    \item \textit{Take a fork to the dining table.}
    \item \textit{Take a transparent bowl to the countertop.}
    \item \textit{Move a damask-style pillow to a sofa.}
    \item \textit{Carry a metal kettle to a fridge.}
    \item \textit{Place a polished-gold vase on a desk.}
    \item \textit{Deliver coffee in a mug with a logo to the dining table.}
    \item \textit{Take a wet teal cloth to the countertop.}
\end{enumerate}

\begin{figure}[h]
    \centering
    \includegraphics[width=0.90\textwidth]{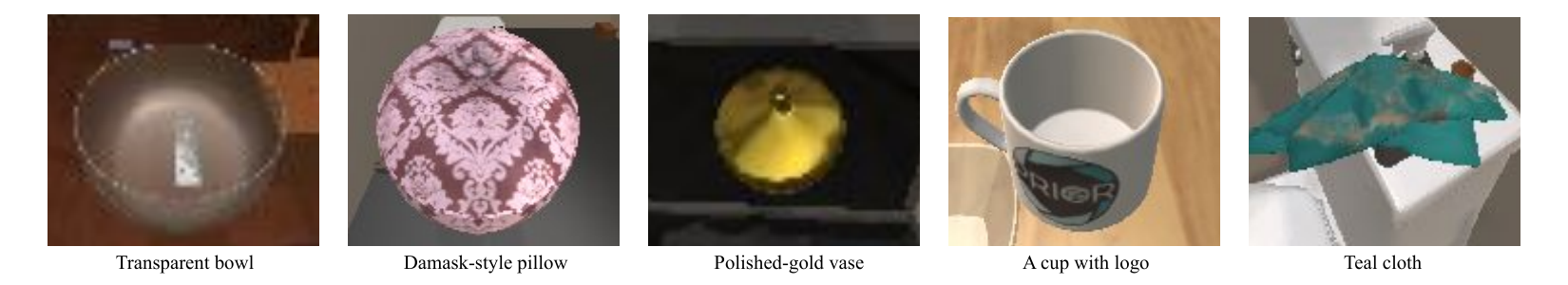}
    \vspace{-0.4cm}
    \caption{The visualization of the objects with the specified attributes in our experiments.}
    \label{fig:important_objects}
    \vspace{-0.2cm}
\end{figure}

\paragraph{Experimental Setup}
For each task, we evaluate the system over three trials across five distinct ProcTHOR houses to account for environmental variability. The houses are selected from the ProcTHOR dataset \cite{deitke2022} to ensure diversity in layout and object distribution. On average, each house contains 52 objects and 34 locations. In the simulation, when an object is picked up, the robot can directly observe the object’s most informative side.
The planner runs in a closed loop, triggering replanning upon execution failure or when a proactive model update is required. Details of the PDDL domain file and the prompts used across different methods are provided in Section~\ref{sec:appendix_prompts}.

\subsection{Appliance Operation Demo setup}

\begin{figure}[h]
    \centering
    \includegraphics[width=0.9\textwidth]{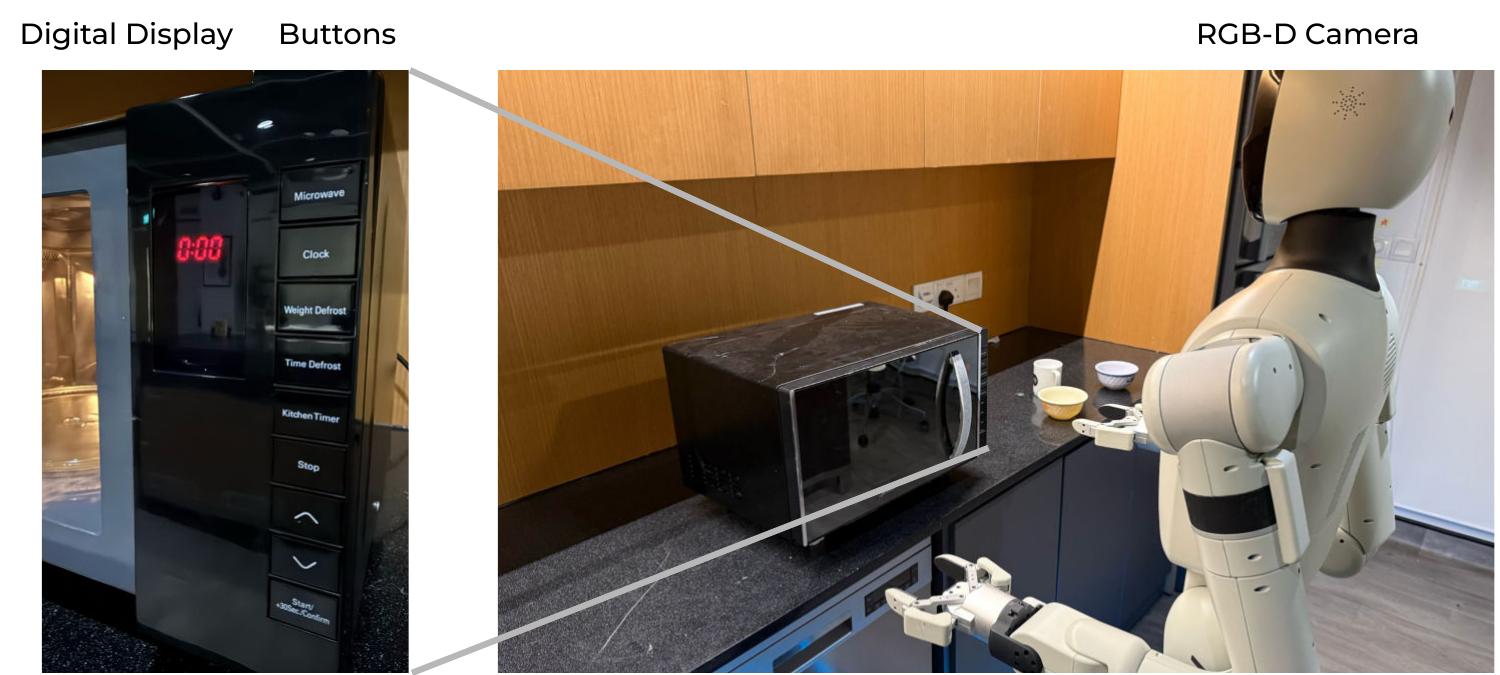}
    \vspace{-0.2cm}
    \caption{Visualization of the Appliance Operation Demo setup.}
    \label{fig:appliance}
    \vspace{-0.2cm}
\end{figure}

\subsubsection{Robot System Integration}
We consider a scenario in which the robot is required to operate an unknown appliance. 
In this demo, we use the Autolife S2 humanoid robot, which features an omnidirectional base, a 4-DOF torso, and two 7-DOF arms. The robot is equipped with one RGB-D camera and two fisheye cameras mounted on the head.
The robot is provided only with prior knowledge of button templates and their corresponding PDDL operators, as detailed in Table~\ref{tab:appliance_buttons}. The robot scans the appliance and detects the bounding box of each button using \texttt{qwen3-vl-235b-a22b-instruct} \cite{bai2025qwen3vltechnicalreport}. Given the detected bounding boxes, the robot computes a 3D point cloud for each button using the depth camera and transforms it into the robot's base frame. The press pose is defined as the fingertip making contact with the center of the button's point cloud. The pre-press pose is generated by offsetting the fingertip 8cm away from the button along the surface normal. Once the action sequence is planned, the robot executes all the button pressing EE pose using task space control. For the PDDL solver, since the FastDownard doesn't support numeric planning, we use the numeric planner called ENHSP \cite{scala2016heuristics} with Unified-Planning library \cite{unified_planning_softwarex2025}.

\begin{table*}[h]
    \centering
    \caption{Button Actions, Preconditions, and Effects for the Microwave Domain.}
    \begin{tabular}{>{\raggedright\arraybackslash}p{0.3\linewidth} >{\raggedright\arraybackslash}p{0.25\linewidth} >{\raggedright\arraybackslash}p{0.35\linewidth}}
        \toprule
        \textbf{Button} & \textbf{Preconditions} & \textbf{Effects} \\
        \midrule
        \texttt{mode-button(?m)} \newline \textit{(MicroWave, Clock, Defrost, Timer)} & 
        $\neg$\texttt{Running(?m)} & 
        \texttt{SelectedMode(?m)}, $\neg$\texttt{OtherModes(?m)} \\
        \midrule
        \texttt{increase(?m)} & 
        $\neg$\texttt{Running(?m)} & 
        \textit{Unknown Numeric Effect} \\
        \midrule
        \texttt{decrease(?m)} & 
        $\neg$\texttt{Running(?m)} & 
        \textit{Unknown Numeric Effect} \\
        \midrule
        \texttt{start(?m)} & 
        $\neg$\texttt{Running(?m)} & 
        \texttt{Running(?m)} \\
        \midrule
        \texttt{stop(?m)} & 
        \texttt{Running(?m)} & 
        $\neg$\texttt{Running(?m)}, \texttt{IdleMode(?m)}, $\neg$\texttt{AllModes(?m)} \\
        \bottomrule
    \end{tabular}
    \label{tab:appliance_buttons}
\end{table*}

\subsubsection{Tasks Setup}
The demonstration involves a multi-stage task: \textit{``Defrost the food for 40 seconds.''} Although the instruction appears simple, it entails substantial complexity for a robot. In our setting, the robot can only reliably perceive the microwave display and buttons when the RGB-D camera is horizontally aligned with the appliance, requiring the robot to “sit down” to observe the interface. Throughout the task, the robot must construct a planning problem based on its perceptual inputs, generate and evaluate hypotheses when necessary, and account for the plausibility of these hypotheses during planning and execution. Details of the PDDL domain file and the prompts are provided in Section~\ref{sec:appendix_prompts}.

\clearpage
\section{Extended Analysis of Experimental Results}
In this section, we provide an in-depth analysis of the experimental results presented in the main paper. We examine behavioral patterns during task execution, analyze common failure cases, and report detailed outcomes across different task settings.

\subsection{Behavioral Analysis in Real-World Tasks}
We present a detailed analysis of the robot's behavior across five distinct long-horizon tasks, highlighting its ability to hypothesize, verify, and replan in the face of uncertainty. The accompanying video provides a visual walkthrough of these executions.

\begin{figure}[h]
    \centering
    \includegraphics[width=1\textwidth]{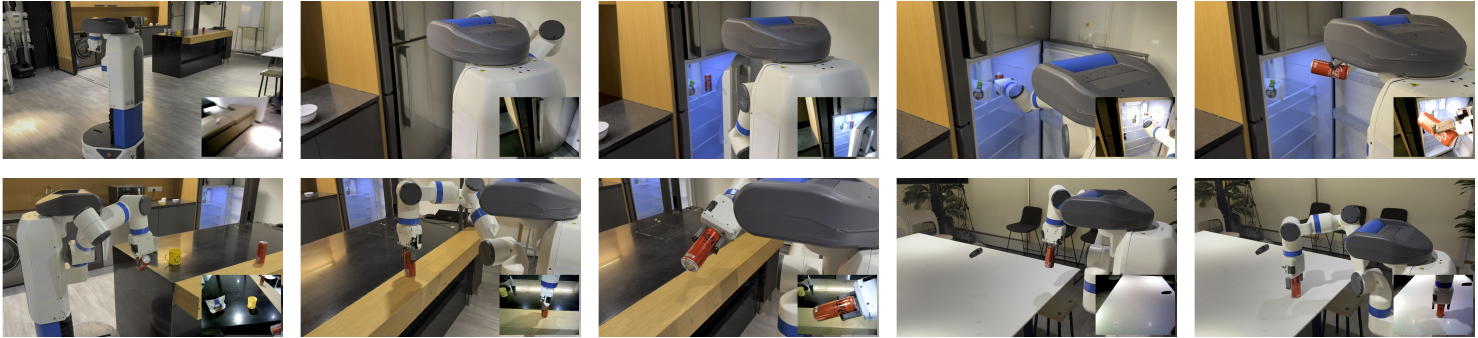}
    \vspace{-0.3cm}
    \caption{Behavioral Analysis of Task 1: Deliver a zero-sugar drink to the table.}
    \vspace{-0.3cm}
    \label{fig:task1}
\end{figure}
\paragraph{Task 1: Deliver a zero-sugar drink to the table (Fig.~\ref{fig:task1})}
In this task, the robot is searching for a zero-sugar drink. It initiates by hypothesizing that the target drink is located in the refrigerator. The robot navigates to the fridge, opens it, and observes a beverage can. After grasping the object, the robot performs an attribute verification check using its head camera. The verification module rejects the hypothesis that the held object is ``zero-sugar,'' identifying it as a standard version. This rejection triggers the hypothesis generation module to propose a new potential location. The robot then plans to explore the kitchen island. Upon arrival, it observes another drink. Since its gripper is currently occupied with the rejected drink, the robot places the current object down before grasping the new one. The robot picks up the second drink from the kitchen island, verifies it, and this time the attribute verification accepts the hypothesis. Finally, the robot successfully delivers the confirmed zero-sugar drink to the table.

\begin{figure}[h]
    \centering
    \includegraphics[width=1\textwidth]{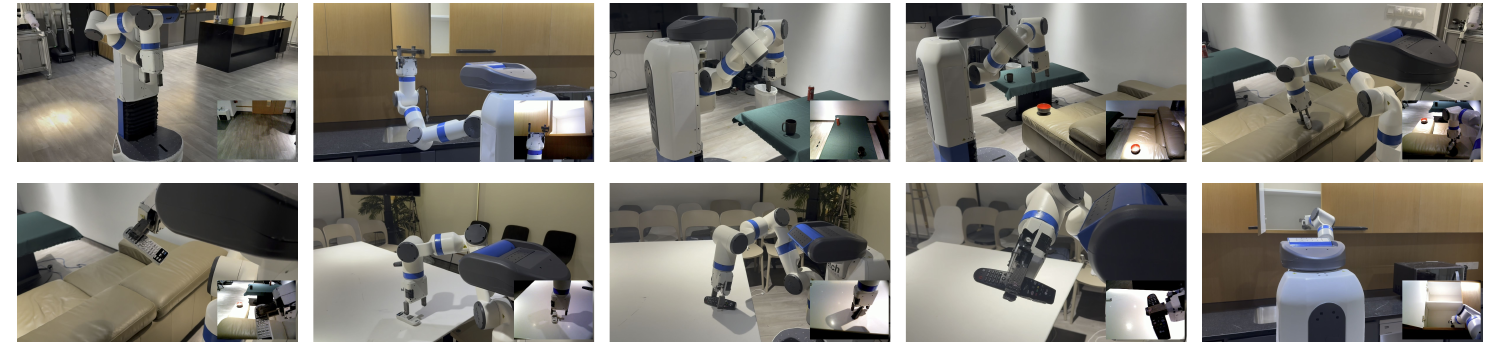}
    \vspace{-0.3cm}
    \caption{Behavioral Analysis of Task 2: Place the remote with red button into the cabinet}
    \label{fig:task2}
    \vspace{-0.3cm}
\end{figure}
\paragraph{Task 2: Place the remote with a red button into the cabinet (Fig.~\ref{fig:task2})}
The robot's goal is to find a remote with a red button and store it in the cabinet. Given the robot's single-arm constraint, the planner prioritizes opening the destination container (the cabinet) first to avoid holding an object while trying to open a door later. After opening the cabinet, the robot searches for the remote. It first navigates to the coffee table but finds no remote. It then proceeds to the sofa, where it detects a remote control. The robot grasps it and performs verification, but the system detects that this remote lacks the required ``red button'' attribute. The hypothesis is rejected, triggering a replan. The robot then navigates to the meeting room table, places the incorrect remote down, and grasps a second remote found there. Verification confirms the presence of the red button. The robot then carries the correct remote back to the cabinet and places it inside, completing the task.

\begin{figure}[h]
    \centering
    \includegraphics[width=1\textwidth]{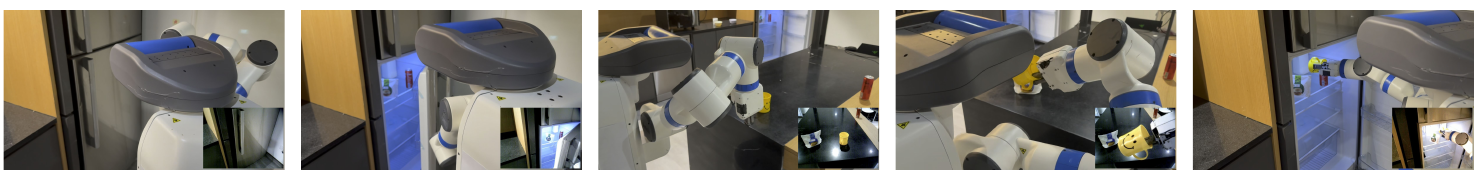}
    \vspace{-0.3cm}
    \caption{Behavioral Analysis of Task 3: Move the smiley-face mug to the fridge}
    \label{fig:task3}
    \vspace{-0.3cm}
\end{figure}
\paragraph{Task 3: Move the smiley-face mug to the fridge (Fig.~\ref{fig:task3})}
For this task, the robot must locate a smiley-face mug and place it in the fridge. The robot begins by navigating to the fridge and opening it to prepare the destination. It then proceeds to the kitchen island, where it observes a mug. The robot grasps the mug and verifies its pattern. The verification module confirms it is the ``smiley-face'' mug. With the correct object in hand, the robot simply navigates back to the already-open fridge and places the mug inside.

\begin{figure}[h]
    \centering
    \includegraphics[width=1\textwidth]{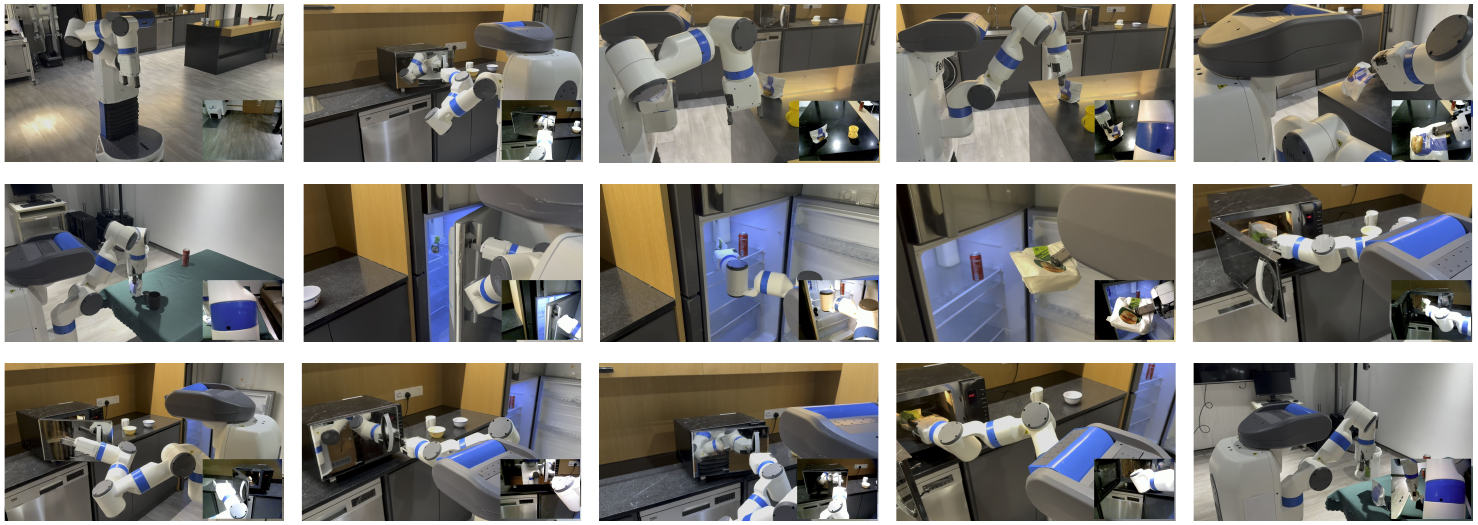}
    \vspace{-0.3cm}
    \caption{Behavioral Analysis of Task 4: Serve a heated chicken burger on the coffee table.}
    \label{fig:task4}
    \vspace{-0.3cm}
\end{figure}
\paragraph{Task 4: Serve a heated chicken burger on the coffee table (Fig.~\ref{fig:task4})}
The robot needs to deliver a heated chicken burger. The planner infers that the burger must be heated, implicating the microwave. The robot first navigates to the microwave and opens it. It then moves to the kitchen island, where it detects a burger. After grasping and verifying that it is indeed the ``chicken'' burger (distinguishing it from the fish burger in the fridge), the robot carries it to the microwave. It places the burger inside, closes the door, and triggers the heating action. Once the process is complete, the robot opens the microwave, retrieves the heated burger, and delivers it to the coffee table.

\begin{figure}[h]
    \centering
    \includegraphics[width=1\textwidth]{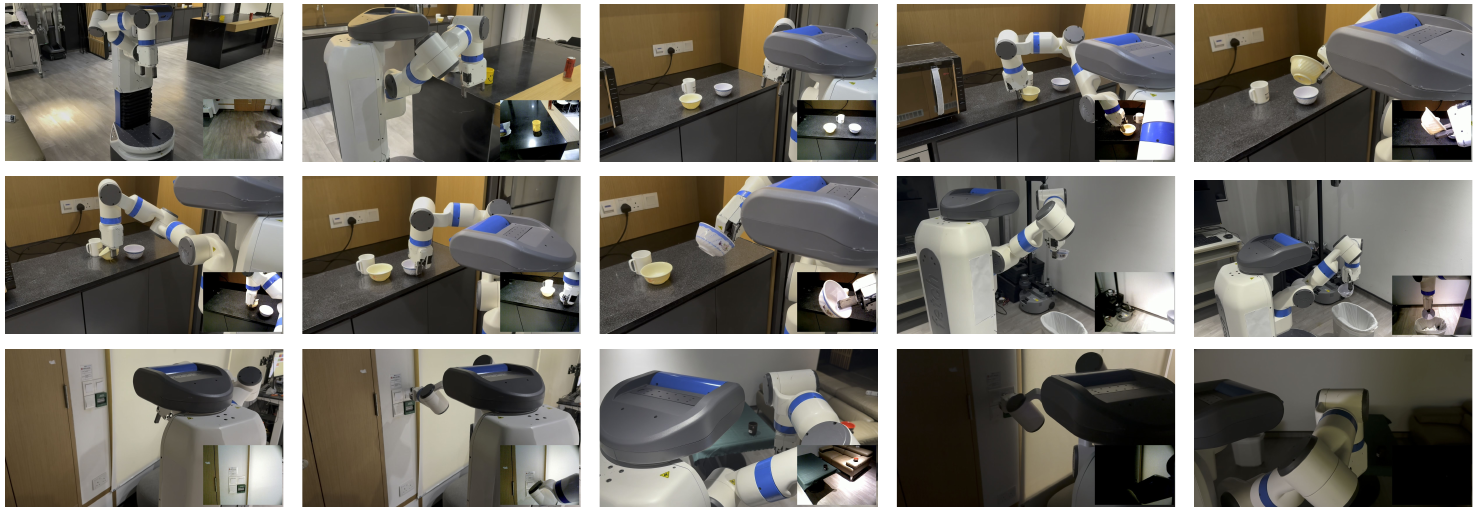}
    \vspace{-0.3cm}
    \caption{Behavioral Analysis of Task 5: Throw away the blue-floral bowl and turn off the living room light.}
    \label{fig:task5}
    \vspace{-0.3cm}
\end{figure}
\paragraph{Task 5: Throw away the blue-floral bowl and turn off the living room light (Fig.~\ref{fig:task5})}
This complex task involves both object manipulation and environmental interaction. The robot first hypothesizes the locations of the bowl and the light switch. It searches the kitchen island first but finds no bowl. A replan sends it to the kitchen counter, where it observes two bowls. It picks up the first one, but the verification module rejects it as not having the ``blue-floral'' pattern. The robot places it back and grasps the second bowl. This time, verification accepts the hypothesis. The robot proceeds to the trash can and discards the bowl. Next, it addresses the light task. It navigates to the control panel, where two switches are visible. It triggers the first switch, but subsequent verification (checking the light intensity in the living room) reveals the light is still on. Validating that the action did not yield the desired effect, the robot triggers the second switch. Verification confirms the living room light is now off, and the task is marked as complete.

\subsection{Failure Mode Analysis in Real-World Experiments}
\begin{figure}[h]
    \centering
    \includegraphics[width=1\textwidth]{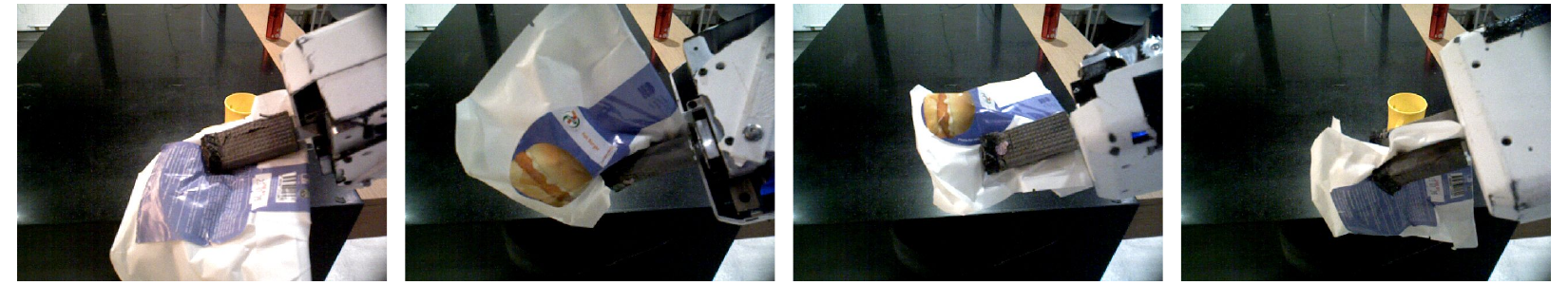}
    \caption{Failure mode: Uncertain Model Expansion failure due to attribute misclassification.}
    \label{fig:fail_attr}
\end{figure}

We categorize the primary failure modes observed during our experiments for all the approches into three types:

\paragraph{Failure with Uncertain Model Expansion}
Failures in this category typically stem from errors in the verification module. For example, in Task 4, the robot might grasp the fish burger and incorrectly classify it as a chicken burger due to perceptual noise or occlusion (Fig.~\ref{fig:fail_attr}). This leads to the robot proceeding with the wrong object, illustrating the dependency of the framework on reliable information gathering assumptions. We further analyze this failure mode in Section \ref{subsec:obj_attr_reasoning}.

\begin{figure}[h]
    \centering
    \includegraphics[width=1\textwidth]{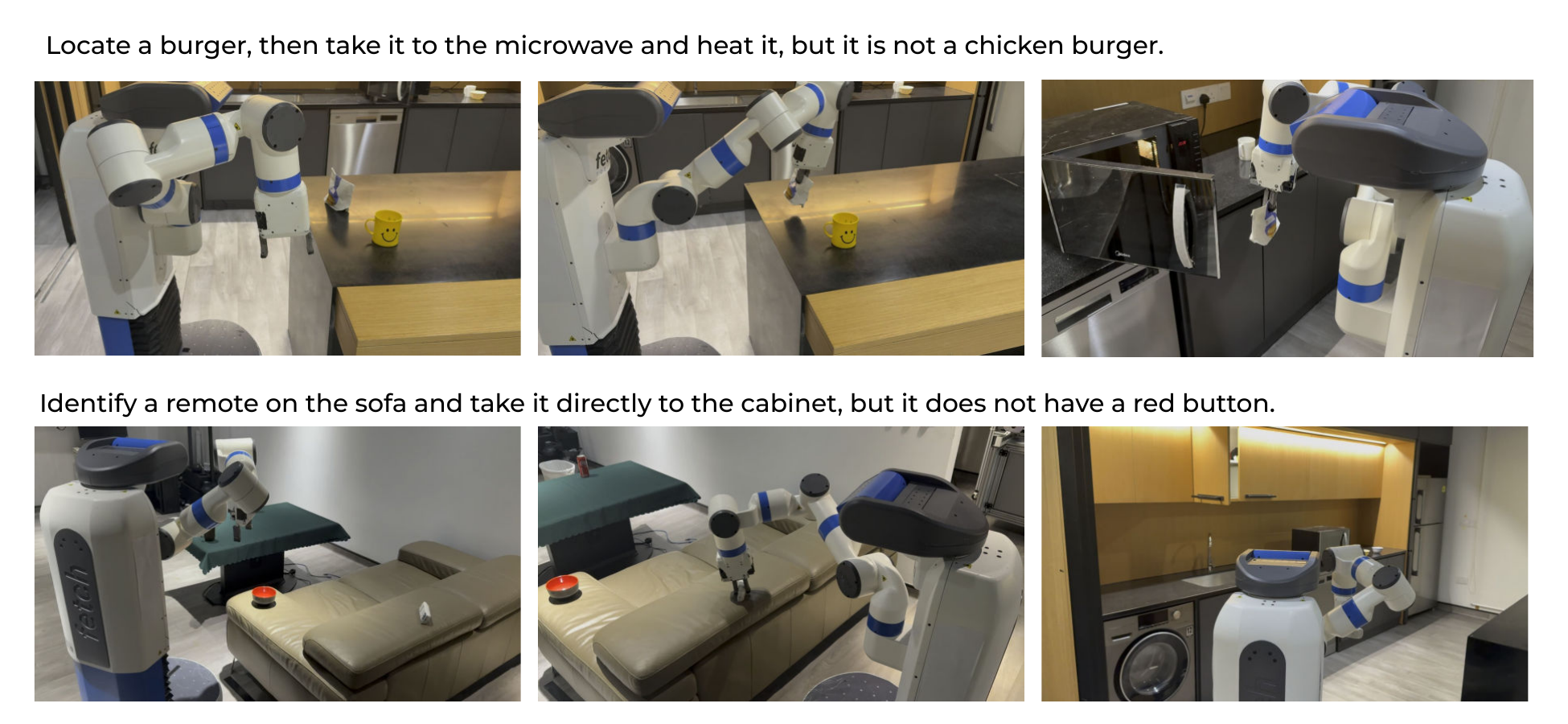}
    \caption{Failure mode: Missing proactive search/verification (Baseline).}
    \label{fig:fail_llm}
\end{figure}

\paragraph{Failure without Uncertain Model Expansion}
Without considering the proactive hypothesis verification in the planning process, the robot lacks the ability to actively search and disambiguate. In scenarios like Task 2, a baseline method might locate the first available remote (the one on the sofa) and immediately attempt to store it, failing to make sure the specific ``red button'' attribute. Similarly, for the burger task, it might heat the incorrect burger type. (Fig.~\ref{fig:fail_llm}).

\paragraph{Failure without Formal Planning}
Approaches that rely purely on reactive policies or unconstrained LLM-based planning frequently lead to deadlocks. Typical failure cases include grasping an object before confirming that the target container is open, attempting to place an object at an unreachable location without prior navigation, or trying to pick up an object when the gripper is already occupied.

While these issues can be mitigated through replanning by regenerating the plan with an LLM, doing so increases the context length, which negatively impacts memory efficiency and limits the effectiveness of enforcing verification constraints.

\subsection{Analysis of Object Attribute Reasoning}
\label{subsec:obj_attr_reasoning}
\begin{figure}[h]
    \centering
    \includegraphics[width=1\textwidth]{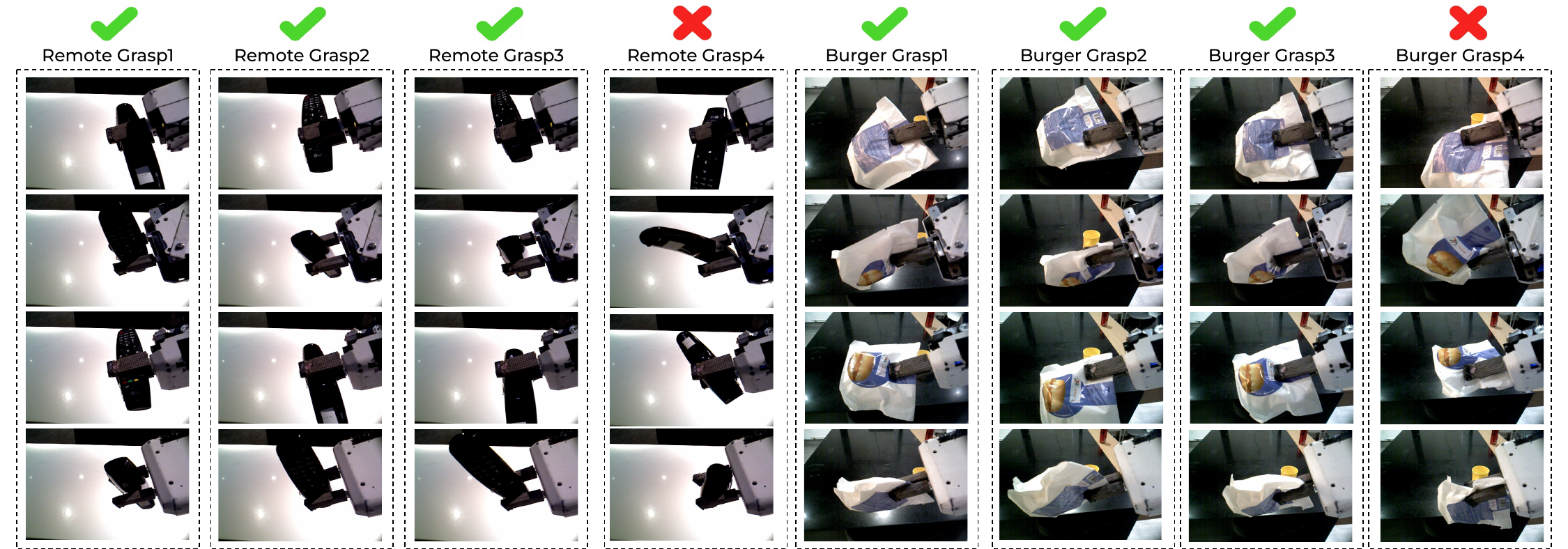}
    \caption{Analysis of grasping strategy impact on attribute verification.}
    \label{fig:grasp_analysis}
\end{figure}

We further analyze how different grasping strategies affect object attribute reasoning. As illustrated in Fig.~\ref{fig:grasp_analysis}, the grasp pose has a significant impact on the visibility of discriminative features during in-hand verification. For example, grasping a remote from the bottom may move the red button out of the camera’s field of view, while grasping a burger from the side can occlude the frontal view of the burger package.

These results indicate that grasp selection is not merely a manipulation primitive, but also an informational action. Although our current system relies on fixed grasp heuristics, future work could investigate active grasp selection strategies that explicitly maximize information gain for attribute verification.

\subsection{Detailed Quantitative Results in Simulation}
\begin{figure}[h]
    \centering
    \includegraphics[width=1\textwidth]{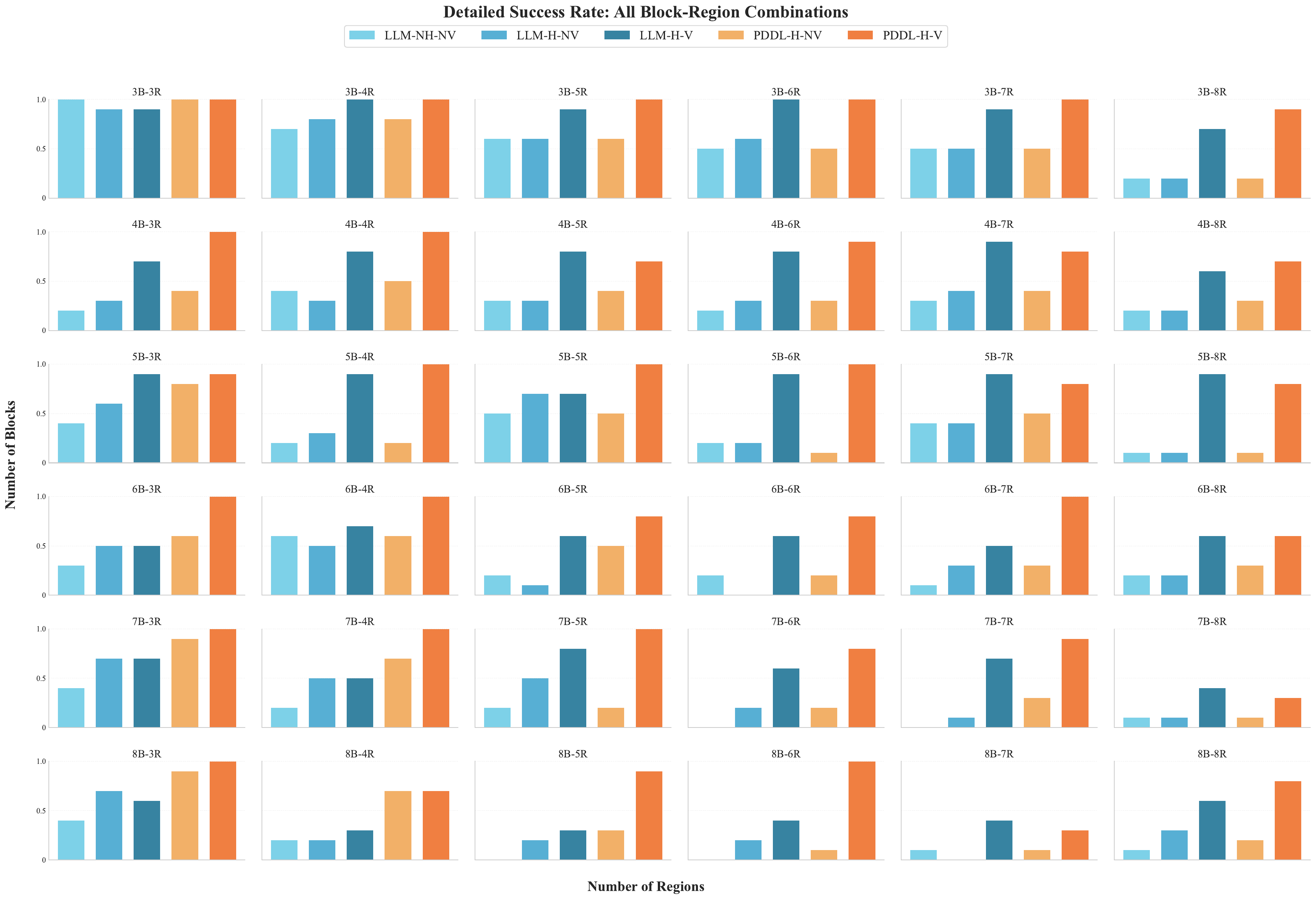}
    \caption{Detailed quantitative results for Block Processing World.}
    \label{fig:block_res}
\end{figure}

We provide complementary quantitative results from our simulation experiments.
\paragraph{Block Processing World}
Figure~\ref{fig:block_res} details the performance breakdown in the Block World domain. The results demonstrate that our method significantly outperforms baselines in tasks requiring the discovery of hidden causal effects (processor functions) and managing state constraints. 

\begin{figure}[h]
    \centering
    \includegraphics[width=1\textwidth]{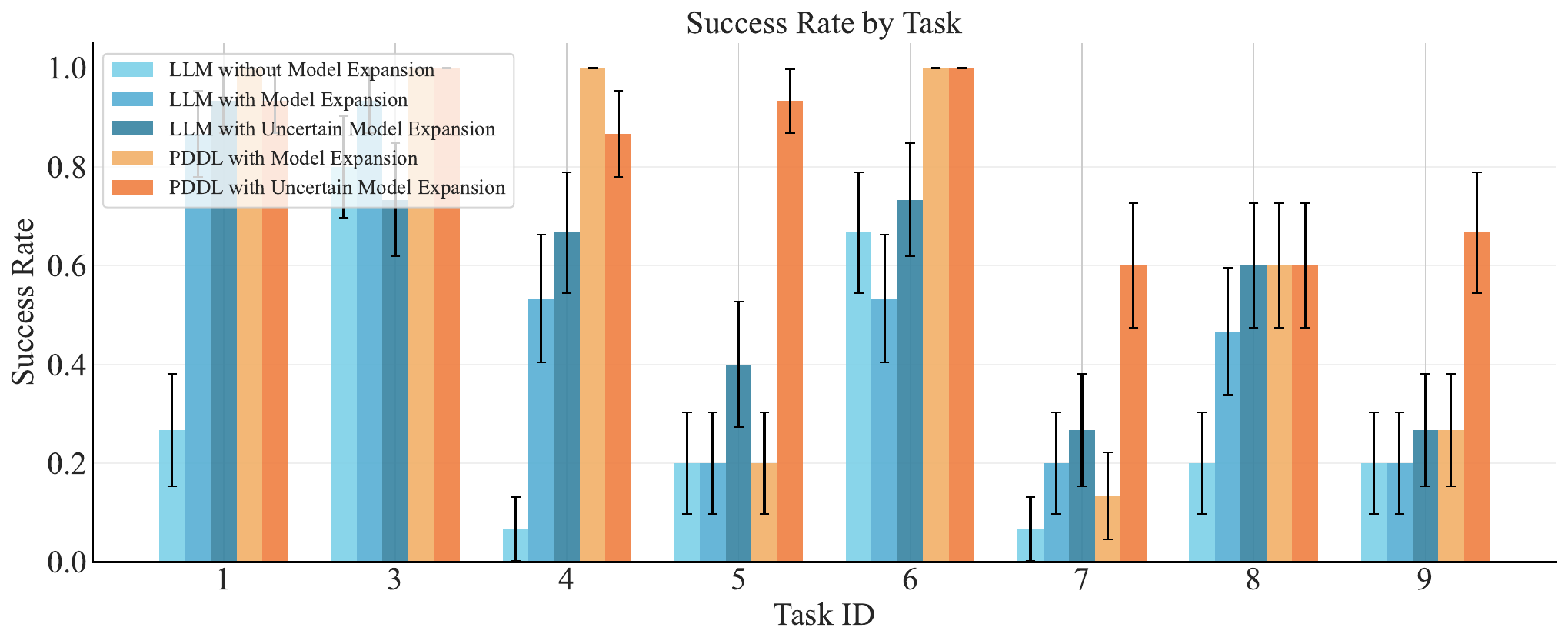}
    \caption{Detailed quantitative results for AI2THOR experiments.}
    \label{fig:ai2_res}
\end{figure}

\paragraph{Mobile Manipulation in AI2THOR}
Figure~\ref{fig:ai2_res} presents the detailed outcomes for the AI2THOR tasks. The breakdown reveals that implicit search and hypothesis management are crucial for success in massive, unknown environments. 

\clearpage
\section{Additional Implementation Details}
This section provides comprehensive implementation details for the proposed framework, including the PDDL domain definitions used across all experimental scenarios, the mechanism for planning under hypothesis uncertainty, and the complete set of prompts used for LLM-based hypothesis generation and verification.

\subsection{Initial PDDL Domains Across All Scenarios}
\label{sec:pddl}
We employ three distinct PDDL domain formulations corresponding to the three experimental scenarios: Block World manipulation, mobile manipulation in household environments, and microwave panel control. Each domain defines the state space, available actions, preconditions, and effects that form the foundation for symbolic planning.

\subsubsection{Block World Domain}
The Block World domain models a tabletop manipulation scenario where blocks can be stacked on each other or placed on designated regions that may trigger state changes. The domain includes the following predicates and actions:

\begin{lstlisting}[language=Lisp, caption={Block World PDDL Domain}, label=lst:blocks_domain]
(define (domain blocks)
    (:requirements :adl :strips :derived-predicates 
                   :existential-preconditions :conditional-effects)
    (:types)

    (:predicates
        (region ?x)         ; x is a triggerable region
        (block ?x)          ; x is a block
        (on ?x ?y)          ; x is on y
        (ontable ?x)        ; x is on the table
        (clear ?x)          ; nothing is on top of x
        (handempty)         ; robot hand is empty
        (handfull)          ; robot hand holds something
        (holding ?x)        ; robot is holding x
        (triggered ?x)      ; region x has been triggered
    )

    (:action pick-up
        :parameters (?x)
        :precondition (and (clear ?x) (ontable ?x) 
                           (handempty) (block ?x))
        :effect (and (not (ontable ?x)) (not (clear ?x))
                     (not (handempty)) (handfull) (holding ?x)))

    (:action put-down
        :parameters (?x)
        :precondition (and (holding ?x) (handfull) (block ?x))
        :effect (and (not (holding ?x)) (clear ?x)
                     (handempty) (not (handfull)) (ontable ?x)))

    (:action stack
        :parameters (?x ?y)
        :precondition (and (holding ?x) (clear ?y) 
                           (handfull) (block ?x))
        :effect (and (not (holding ?x)) (not (clear ?y))
                     (clear ?x) (handempty) (not (handfull)) (on ?x ?y)))

    (:action unstack
        :parameters (?x ?y)
        :precondition (and (on ?x ?y) (clear ?x) 
                           (handempty) (block ?x))
        :effect (and (holding ?x) (clear ?y) (not (clear ?x))
                     (not (handempty)) (handfull) (not (on ?x ?y))))

    (:action trigger
        :parameters (?x ?y)
        :precondition (and (block ?y) (on ?y ?x) 
                           (region ?x) (handempty))
        :effect (triggered ?x))
)
\end{lstlisting}

\subsubsection{Mobile Manipulation Domain}
The mobile manipulation domain extends the planning capabilities to household environments where a mobile robot navigates between locations, manipulates objects, and interacts with appliances. This domain is used for both AI2THOR simulations and real-world experiments.

\begin{lstlisting}[language=Lisp, caption={Mobile Manipulation PDDL Domain}, label=lst:mobile_domain]
(define (domain mobile_manipulation)
  (:requirements :strips :equality)
  (:types)

  (:predicates
    (Robot ?r)            ; r is a robot
    (Obj ?o)              ; o is an object
    (Room ?r)             ; r is a room
    (Location ?l)         ; l is a location
    (Type ?t)             ; t is an object type
    (Is ?o ?t)            ; object o is of type t
    (HandEmpty ?r)        ; robot's hand is empty
    (CanMove ?r)          ; robot can move
    (Holding ?r ?o)       ; robot is holding object o
    (At ?o ?l)            ; object o is at location l
    (AtRoom ?l ?r)        ; location l is in room r
    (IsOpen ?l)           ; location l is open/accessible
    (Trigged ?l)          ; location l has been triggered
  )

  (:action navigate
    :parameters (?r ?l1 ?l2)
    :precondition (and (Robot ?r) (CanMove ?r)
                       (Location ?l1) (Location ?l2) (At ?r ?l1))
    :effect (and (not (CanMove ?r)) (At ?r ?l2) (not (At ?r ?l1))))

  (:action pick
    :parameters (?r ?o ?l)
    :precondition (and (Robot ?r) (Obj ?o) (Location ?l)
                       (HandEmpty ?r) (At ?r ?l) (IsOpen ?l) (At ?o ?l))
    :effect (and (Holding ?r ?o) (CanMove ?r)
                 (not (HandEmpty ?r)) (not (At ?o ?l))))

  (:action place
    :parameters (?r ?o ?l)
    :precondition (and (Robot ?r) (Obj ?o) (Location ?l) (not (Room ?l))
                       (At ?r ?l) (IsOpen ?l) (not (HandEmpty ?r)) 
                       (Holding ?r ?o))
    :effect (and (HandEmpty ?r) (CanMove ?r) (At ?o ?l) 
                 (not (Holding ?r ?o))))

  (:action open
    :parameters (?r ?l)
    :precondition (and (Robot ?r) (Location ?l) (not (Room ?l))
                       (At ?r ?l) (HandEmpty ?r) (not (IsOpen ?l)))
    :effect (and (IsOpen ?l) (CanMove ?r)))

  (:action close
    :parameters (?r ?l)
    :precondition (and (Robot ?r) (Location ?l) (not (Room ?l))
                       (At ?r ?l) (HandEmpty ?r) (IsOpen ?l))
    :effect (and (not (IsOpen ?l)) (CanMove ?r)))

  (:action trigger
    :parameters (?r ?l ?o)
    :precondition (and (Robot ?r) (Location ?l) (not (Room ?l))
                       (At ?r ?l) (At ?o ?l) (HandEmpty ?r))
    :effect (and (Trigged ?l) (CanMove ?r)))
)
\end{lstlisting}

\subsubsection{Microwave Panel Domain}
The microwave panel domain models appliance control through button presses. Different buttons switch between operating modes, and numeric adjustments are mode-dependent.

\begin{lstlisting}[language=Lisp, caption={Microwave Panel PDDL Domain}, label=lst:microwave_domain]
(define (domain microwave_panel)
  (:requirements :strips :typing :equality 
                 :numeric-fluents :negative-preconditions 
                 :conditional-effects)
  (:types microwave)

  (:predicates
    (microwave-mode ?m - microwave)
    (clock-mode ?m - microwave)
    (weight-defrost-mode ?m - microwave)
    (time-defrost-mode ?m - microwave)
    (kitchen-timer-mode ?m - microwave)
    (idle-mode ?m - microwave)
    (running ?m - microwave)
    (homepose)
  )

  (:functions)

  (:action press-microwave-button
    :parameters (?m - microwave)
    :precondition (not (running ?m))
    :effect (and (not (idle-mode ?m)) (not (clock-mode ?m))
                 (not (weight-defrost-mode ?m)) 
                 (not (time-defrost-mode ?m))
                 (not (kitchen-timer-mode ?m))
                 (microwave-mode ?m) (not (homepose))))

  (:action press-time-defrost-button
    :parameters (?m - microwave)
    :precondition (not (running ?m))
    :effect (and (not (idle-mode ?m)) (not (microwave-mode ?m))
                 (not (clock-mode ?m)) (not (weight-defrost-mode ?m))
                 (not (kitchen-timer-mode ?m))
                 (time-defrost-mode ?m) (not (homepose))))

  (:action press-increase-button
    :parameters (?m - microwave)
    :precondition (not (running ?m))
    :effect (not (homepose)))

  (:action press-decrease-button  
    :parameters (?m - microwave)
    :precondition (not (running ?m))
    :effect (not (homepose)))

  (:action press-start-button
    :parameters (?m - microwave)
    :precondition (not (running ?m))
    :effect (running ?m))

  (:action press-stop-button
    :parameters (?m - microwave)
    :precondition (running ?m)
    :effect (and (not (running ?m)) (not (microwave-mode ?m))
                 (not (clock-mode ?m)) (not (weight-defrost-mode ?m))
                 (not (time-defrost-mode ?m)) 
                 (not (kitchen-timer-mode ?m))
                 (idle-mode ?m)))

  (:action go-home
    :parameters (?m - microwave)
    :precondition (not (homepose))
    :effect (homepose))
)
\end{lstlisting}

\subsection{Implementation Details for Planning Under Hypothesis Uncertainty}
A key technical challenge in our framework is incorporating hypothesis uncertainty into the symbolic planning process. Since hypotheses represent uncertain knowledge about object existence, attributes, or action effects, the planner must reason about both verified and unverified information while prioritizing verification of uncertain elements.

\subsubsection{Hypothesis-Related Predicates}
We extend the PDDL domain with additional predicates to track hypothesis status:

\begin{lstlisting}[language=Lisp, caption={Hypothesis Tracking Predicates}, label=lst:hypothesis_predicates]
(related ?h ?x)           ; hypothesis h is related to object/location x
(true ?h)                 ; hypothesis h has been verified as true
(hypothesis_related ?x)   ; derived: x has unverified hypotheses
(all_hypothesis_true)     ; derived: all hypotheses are verified

(:derived (all_hypothesis_true)
    (forall (?h - hypothesis) (true ?h)))

(:derived (hypothesis_related ?x)
    (exists (?h - hypothesis) (and (related ?h ?x) (not (true ?h)))))
\end{lstlisting}

\subsubsection{Cost-Based Action Splitting}
Since FastDownward does not support conditional action costs, we implement a workaround by splitting each action into two variants: a normal version for verified objects and a hypothesis-related version for objects with unverified hypotheses. The hypothesis-related version incurs a higher cost, biasing the planner to operate on verified objects when possible while still allowing exploration of uncertain hypotheses.

\begin{lstlisting}[language=Lisp, caption={Action Splitting for Cost Differentiation}, label=lst:action_splitting]
; Normal pick action - low cost (1)
(:action pick
  :parameters (?r ?o ?l)
  :precondition (and (Robot ?r) (Obj ?o) (Location ?l)
                     (HandEmpty ?r) (At ?r ?l) (IsOpen ?l) (At ?o ?l)
                     (not (hypothesis_related ?o)))
  :effect (and (Holding ?r ?o) (CanMove ?r)
               (not (HandEmpty ?r)) (not (At ?o ?l))
               (increase (total-cost) 1)))

; Hypothesis-related pick action - high cost (10)
(:action pick-hypothesis
  :parameters (?r ?o ?l)
  :precondition (and (Robot ?r) (Obj ?o) (Location ?l)
                     (HandEmpty ?r) (At ?r ?l) (IsOpen ?l) (At ?o ?l)
                     (hypothesis_related ?o))
  :effect (and (Holding ?r ?o) (CanMove ?r)
               (not (HandEmpty ?r)) (not (At ?o ?l))
               (increase (total-cost) 10)))
\end{lstlisting}

This pattern is applied to all relevant actions: \texttt{navigate}, \texttt{pick}, \texttt{place}, \texttt{open}, \texttt{close}, and \texttt{trigger}. The cost ratio (1:10) encourages the planner to:
\begin{enumerate}
    \item Verify hypotheses early when they lie on the critical path
    \item Avoid unnecessary manipulation of unverified objects
\end{enumerate}

\subsubsection{Dynamic Verification Action Generation}
For each hypothesis, we generate a corresponding verification action that can only be executed when the appropriate verification conditions are met (e.g., holding the object for attribute verification, or being at the location for existence verification):

\begin{lstlisting}
(:action verify_h
  :parameters (?r - Robot ?o - Obj ?l - Location)
  :precondition (and (At ?r ?l) ((*@$\neg$@*)True ?h) (Related ?h ?o) (*@$\phi_{cond}$@*) (*@$\phi_{verify}$@*))
  :effect (and (True ?h))
)
\end{lstlisting}

\subsubsection{Augmented Problem Construction}
When hypotheses are generated, the problem file is augmented with:
\begin{itemize}
    \item Hypothesis constants declared in the domain
    \item Initial state facts for hypothesized objects and their attributes
    \item \texttt{related} predicates linking objects to their hypotheses
    \item Goal augmented with \texttt{(all\_hypothesis\_true)} to enforce verification (Optional)
\end{itemize}

\begin{lstlisting}[language=Lisp, caption={Example Augmented Problem}, label=lst:augmented_problem]
(define (problem realworld)
  (:domain mobile_manipulation)
  (:objects 
    table1 kitchenisland1 trashcan1 wallmountedcontrolpanel1
    livingroom1 kitchen1 fetch bowl1 switch1 trigger
  )
  (:init
    ; Hypothesized object existence
    (Type bowl) (Obj bowl1) (Is bowl1 bowl) 
    (At bowl1 kitchenisland1) (related h1 bowl1)
    ; Hypothesized object attribute
    (blue-floral bowl1) (related h2 bowl1)
    ; Hypothesized switch location
    (Type switch) (Obj switch1) (Is switch1 switch)
    (At switch1 wallmountedcontrolpanel1) (related h3 switch1)
    ; Environment and robot state...
  )
  (:goal (and 
    (exists (?o ?l) (and (Is ?o bowl) (blue-floral ?o) 
                         (Is ?l trashcan) (At ?o ?l)))
    (light-off livingroom1)
    (all_hypothesis_true)))
)
\end{lstlisting}

\subsection{Prompts Used in Experiments and Demonstrations}
\label{sec:appendix_prompts}
This section documents the prompts used for LLM-based hypothesis generation, verification, and planning across different experimental scenarios. These prompts evolved through iterative refinement to improve reliability and consistency.
We mainly illustrate the prompts used in mobile manipulation experiments, and the prompts used in block world experiments are similar, except for the action, predicate, and some examples.

\subsubsection{Classification Prompt}
Before hypothesis generation, predicates are classified into attribute predicates (fixed object properties like \texttt{plastic}, \texttt{glass}) and effect predicates (states that can change through actions like \texttt{toasted}, \texttt{washed}):

\begin{lstlisting}[language=Python, caption={Predicate Classification Prompt}, label=lst:classification_prompt]
SYSTEM_PROMPT_FOR_CLASSIFICATION = '''
Classify the type of predicate into attribute predicates 
(fixed attribute of an object, e.g. plastic, glass, etc.) 
or action effect predicates (a state of an object that can be 
affected by an action, e.g. warm, dry, contain_soup, cooked etc.):
Q: { 'predicate1', 'predicate2', 'predicate3'}
A: {'attribute': {'predicate1'}, 'effect': {'predicate2'}}

Example:
Q: { 'plastic', 'contains_soup'}
A: {'attribute': {'plastic'}, 'effect': {'contains_soup'}}
'''
\end{lstlisting}

\subsubsection{Hypothesis Generation Prompt}
The hypothesis generation prompt instructs the LLM to propose hypotheses about object locations, attributes, and action effects based on commonsense reasoning:

\begin{lstlisting}[language=Python, caption={Hypothesis Generation Prompt (Mobile Manipulation)}, label=lst:hypothesis_prompt]
SYSTEM_PROMPT_FOR_HYPOTHESIS = '''
You are given a planning problem and need to generate 
hypothesis to solve the plan.

Robot description: The robot can manipulate objects in the 
environment, and can only observe an object when at the same 
location, observe attributes when holding the object.

Given the input of missing things and the existing observation list.
Output should be a list of hypotheses. Only generate top guesses.

Each hypothesis contains: 
- id: unique id (h1, h2, etc.)
- type: object_existance, object_attribute, or action_effect
- description: brief description
- content: PDDL facts to add if true
- condition: dependent hypothesis id (if any)
- verification condition: state constraint for verification

Example 1:
Goal: Take a thick towel to a bed.
Formal Goal: (exists (?o ?l) (and (Is ?o towel) (thick ?o) (Is ?l bed) (At ?o ?l)))
Current known observation list: XXX
Extracted unknown objects/location: towel
Extracted unknown attributes: [effect attributes:] [visual attributes: thick]
```json
[
    {
        "id": "h1",
        "type": "object_existance",
        "description": "There is a towel at the towelrack1",
        "content": add_fact(
            "towel1",
            ["(Obj towel1)", "(Is towel1 towel)", "(At towel1 towelrack1)"]
        ),
        "condition": "",
        "verification condition": "(At ?r towelrack1) (IsOpen towelrack1)"
    },
    {
        "id": "h2",
        "type": "object_attribute",
        "description": "The towel1 is thick",
        "content": add_fact(
            "towel1",
            ["(thick towel1)"]
        ),
        "condition": "h1",
        "verification condition": "(Holding ?r towel1)"
    }
]
```

Example 2:
Goal: Take a toasted whole wheat breadslice to a glass desk.
Formal Goal: (exists (?o ?l) (and (Is ?o breadslice) (toasted ?o) (wholewheat ?o) (Is ?l desk) (glass ?l) (At ?o ?l)))
Current known observation list: XXX
Extracted unknown objects/location: breadslice
Extracted unknown attributes: [effect attributes: toasted] [visual attributes: wholewheat, glass]
Justification: Given the current known observation list and unknown information, I will guess that a breadslice may be located on diningtable1 in kitchen1, that the bread slice may be whole-wheat, and that desk1 may be a glass desk. In addition, we hypothesize that triggering toaster1 causes the object placed on it to become toasted. 
```json
[
    {
        "id": "h1",
        "type": "object_existance",
        "description": "There maybe a breadslice on the diningtable1 in kitchen1",
        "content": add_fact("breadslice1", ["(Obj breadslice1)", "(Is breadslice1 breadslice)", "(At breadslice1 diningtable1)"]),
        "condition": "",
        "verification condition": "(At ?r diningtable1) (IsOpen diningtable1)"
    },
    {
        "id": "h2",
        "type": "object_attribute",
        "description": "The breadslice1 is whole-wheat",
        "content": add_fact("breadslice1", ["(wholewheat breadslice1)"]),
        "condition": "h1",
        "verification condition": "(Holding ?r breadslice1)"
    },
    {
        "id": "h3",
        "type": "object_attribute",
        "description": "The desk1 is glass",
        "content": add_fact("desk1", ["(glass desk1)"]),
        "condition": "",
        "verification condition": "(At ?r desk1)"
    },
    {
        "id": "h4",
        "type": "action_effect",
        "description": "Triggering the toaster causes the object placed on top of it to become toasted",
        "content": add_effect("trigger", "(when (Is ?l toaster) (and (toasted ?o) (related h4 ?o)))"),
        "condition": "",
        "verification condition": "" # hypothesis is assumed to be true without requiring explicit verification.
    }
]
```

[more examples]

Update Example 1:
Goal: Take an orange plate to the countertop1.
Formal Goal: (exists (?o) (and (Is ?o plate) (orange ?o) (At ?o countertop1)))

Justification: Given the there is no plate on the diningtable1, I will guess that plate1 may be on the diningtable2 based on common sense, and that plate might be orange.

```json
[
    {
        "id": "h1",
        "type": "object_existance",
        "description": "There maybe a plate on the diningtable2",
        "content": add_fact("plate1", ["(Obj plate1)", "(Is plate1 plate)", "(At plate1 diningtable2)"]),
        "condition": "",
        "verification condition": "(At ?r diningtable2) (IsOpen diningtable2)"
    },
    {
        "id": "h2",
        "type": "object_attribute",
        "description": "The plate1 is orange",
        "content": add_fact("plate1", ["(orange plate1)"]),
        "condition": "h1",
        "verification condition": "(Holding ?r plate1)"
    }
]
```
[more update examples]
'''
\end{lstlisting}

The hypothesis generation prompt for the block processing world is slightly different from the one used in the mobile manipulation setting. In the block processing world, all objects are already known to the robot, while the effects of the processors, or regions, remain unknown. Therefore, the main objective is to guide the LLM to infer missing action effects for regions that can process objects placed on top of them. In this setting, the robot can rearrange objects, trigger regions, and observe an object’s state only when holding the object.

\begin{lstlisting}[language=Python, caption={Hypothesis Generation Prompt (Block Processing World)}, label=lst:hypothesis_prompt_block_world]
PROMPT_FOR_HYPOTHESIS = '''
You are given a planning problem and need to extend the effect of an action to solve the plan.

Robot description: The robot can rearrange objects in the environment, trigger a region to apply an effect to the object placed on top of it, and observe the state of an object only when holding it.

Given the missing effects, existing objects, and the actions that can be extended, output a list of single action-effect hypotheses. The number of hypotheses must be equal to the number of missing effects.

Each hypothesis contains:
- id: a unique id for the hypothesis
- type: the type of the hypothesis
- description: a description of the hypothesis
- content: the content of the hypothesis that can be written in the planning domain
- verification condition: a state constraint that can verify the hypothesis

Please ensure that:
- the number of hypotheses is equal to the number of missing effects
- each missing effect corresponds to exactly one region

Example output:

Justification: Given the existing objects and region names, I will guess the following effects for the missing effects.

```json
[
    {
        "id": "h1",
        "type": "action_effect",
        "description": "Triggering the r_toaster causes the object placed on top of it to gain the effect 'toasted'",
        "content": add_effect("trigger", "(when (r_toaster ?x) (and (toasted ?y) (related h1 ?y)))"
        ),
        "verification condition": "(holding ?y)"
    },
    {
        "id": "h2",
        "type": "action_effect",
        "description": "Triggering the r_1 causes the object placed on top of it to gain the effect 'cool'",
        "content": add_effect("trigger", "(when (r_1 ?x) (and (cool ?y) (related h2 ?y)))"),
        "verification condition": "(holding ?y)"
    }
]'''
\end{lstlisting}

\subsubsection{Verification Prompt}
The verification prompt enables the LLM to reason about whether a hypothesis is confirmed or refuted based on robot observations. We provide a simple example prompt for the mobile manipulation setting.:

\begin{lstlisting}[language=Python, caption={Verification Prompt}, label=lst:verification_prompt]
SYSTEM_PROMPT_FOR_VERIFICATION = '''
You are trying to verify whether the hypothesis is true or false.
The observation and hypothesis content will be given.

The output should be structured as follows:
Reason: One sentence reasoning about whether the hypothesis is true 
        (e.g., whether the card is plastic or not).
Result: True or False
'''
\end{lstlisting}

\subsubsection{Language Model Planning Prompts}
We provide the prompts for the most complex mobile manipulation domain. The prompts for the block world follow a similar structure but contain reduced content.
Three variants of planning prompts are used depending on the ablation condition:

\paragraph{H+V (Hypothesis + Verification):} Includes hypothesis-aware planning with explicit verification actions:
\begin{lstlisting}[language=Python, caption={Planning Prompt with Hypothesis and Verification}, label=lst:planning_hv]
SYSTEM_PROMPT_FOR_LLM_H_V = '''
You are task planner that can generate a plan to solve the given problem.
Robot description: The robot is called fetch, it can manipulate the objects 
in the environment, and the robot can only observe object when the robot 
in the same location of the object, observe the attribute and state when 
it holds the object. The action trigger may have different effect when 
apply to different objects.

Actions available:
navigate (room/location): The robot moves to the specified room/location. 
    Precondition: The room/location exists in the environment.
    Effect: The robot is at the room/location.
pick (object): The robot picks up the specified object.
    Precondition: The robot is at the location of the object. The hand is 
                  empty. The location is not a room.
    Effect: The robot is holding the object.
place (location): The robot places the held object at the location. The 
                  robot cannot place an object at a closed location. And 
                  robot should at the location.
    Precondition: The robot is at the location. The location is not closed.
    Effect: The hand is empty. The object is at the location.
open (location): The robot opens a location.
    Precondition: The robot is at the location.
    Effect: The location is opened.
close (location): The robot closes a location. 
    Precondition: The robot is at the location.
    Effect: The location is closed.
trigger (location): The robot activates an appliance with an object on/in 
                    it, there will be corresponding effect on that object.
    Precondition: The robot is at the location. The robot is not holding 
                  any object.
    Effect: The appliance is triggered.
verify (hypothesis): The robot verifies the hypothesis after executing 
                     some action.
    Precondition: The hypothesis's verification condition is satisfied.
    Effect: The hypothesis is verified.

Constraints:
- The robot has a single arm and can hold only one target at a time.
- The action sequence must achieve the specified goal.
- The robot will be given an observation scene graph that partially 
  represents the environment; there may be objects not in the current 
  observation. Plan can include objects that are not in the current 
  observation.
- When you want to manipulate objects, you need to move to the location 
  instead of the room.
- There is a hypothesis set H, you need to generate a plan based on the 
  hypothesis set. The plan should contain the verify action.
- Keep mind you can only verify when verification condition is satisfied 
  (e.g. at same location, holding the object, etc.). And each hypothesis 
  should be verified only once.

Given the Initial state: initial_state
And the goal: goal_state
Hypothesis set: List_of_hypotheses
Output the plan directly.

Example:
Goal: Take a thick towel to a bed.
Formal Goal: (exists (?o ?l) (and (Is ?o towel) (thick ?o) 
             (Is ?l bed) (At ?o ?l)))
Hypotheses: List_of_hypotheses
Output:
(navigate towelrack1)
(verify_h1 towel1)
(pick towel1)
(verify_h2 towel1)
(navigate bed1)
(place bed1)

[more examples]

'''


\end{lstlisting}

\paragraph{Hypothesis Only (H):} Uses hypothesis for planning but omits verification:
\begin{lstlisting}[language=Python, caption={Planning Prompt without Verification}, label=lst:planning_h]
SYSTEM_PROMPT_FOR_LLM_H_nV = '''
You are task planner that can generate a plan to solve the given problem.
Robot description: The robot is called fetch, it can manipulate the objects 
in the environment, and the robot can only observe object when the robot 
in the same location of the object, observe the attribute and state when 
it holds the object. The action trigger may have different effect when 
apply to different objects.

Actions available:
navigate (room/location): The robot moves to the specified room/location. 
    Precondition: The room/location exists in the environment.
    Effect: The robot is at the room/location.
pick (object): The robot picks up the specified object.
    Precondition: The robot is at the location of the object. The hand is 
                  empty. The location is not a room.
    Effect: The robot is holding the object.
place (location): The robot places the held object at the location. The 
                  robot cannot place an object at a closed location. And 
                  robot should at the location.
    Precondition: The robot is at the location. The location is not closed.
    Effect: The hand is empty. The object is at the location.
open (location): The robot opens a location.
    Precondition: The robot is at the location.
    Effect: The location is opened.
close (location): The robot closes a location. 
    Precondition: The robot is at the location.
    Effect: The location is closed.
trigger (location): The robot activates an appliance with an object on/in 
                    it, there will be corresponding effect on that object.
    Precondition: The robot is at the location. The robot is not holding 
                  any object.
    Effect: The appliance is triggered.

Constraints:
- The robot has a single arm and can hold only one target at a time.
- The action sequence must achieve the specified goal.
- The robot will be given an observation scene graph that partially 
  represents the environment; there may be objects not in the current 
  observation. Plan can include objects that are not in the current 
  observation.
- When you want to manipulate objects, you need to move to the location 
  instead of the room.
- There is a hypothesis set H, you need to generate a plan based on the 
  hypothesis set.

Given the Initial state: initial_state
And the goal: goal_state
Hypothesis set: List_of_hypotheses
Output the plan directly.

Example:
Goal: Take a thick towel to a bed.
Formal Goal: (exists (?o ?l) (and (Is ?o towel) (thick ?o) 
             (Is ?l bed) (At ?o ?l)))
Hypotheses: List_of_hypotheses
Output:
(navigate towelrack1)
(pick towel1)
(navigate bed1)
(place bed1)

[more examples]
'''
\end{lstlisting}

\paragraph{No Hypothesis (Baseline):} Plans without any hypothesis augmentation:
\begin{lstlisting}[language=Python, caption={Baseline Planning Prompt}, label=lst:planning_baseline]
SYSTEM_PROMPT_FOR_LLM_nH_nV = '''
You are a task planner who can generate a plan to solve the given problem.
Robot description: The robot is called Fetch. It can manipulate the objects 
in the environment, and it can only observe objects when the robot is 
in the same location as the object, observe the attributes and state when 
it holds the object. The action trigger may have a different effect when 
applied to different objects.

Actions available:
navigate (room/location): The robot moves to the specified room/location. 
    Precondition: The room/location exists in the environment.
    Effect: The robot is in the room/location.
pick (object): The robot picks up the specified object.
    Precondition: The robot is at the location of the object. The hand is 
                  empty. The location is not a room.
    Effect: The robot is holding the object.
place (location): The robot places the held object at the location. The 
                  robot cannot place an object at a closed location. And 
                  the robot should be at the location.
    Precondition: The robot is at the location. The location is not closed.
    Effect: The hand is empty. The object is at the location.
open (location): The robot opens a location.
    Precondition: The robot is at the location.
    Effect: The location is opened.
close (location): The robot closes a location. 
    Precondition: The robot is at the location.
    Effect: The location is closed.
trigger (location): The robot activates an appliance with an object on/in 
                    it, and there will be a corresponding effect on that object.
    Precondition: The robot is at the location. The robot is not holding 
                  any object.
    Effect: The appliance is triggered.

Constraints:
- The robot has a single arm and can hold only one target at a time.
- The action sequence must achieve the specified goal.
- The robot will be given an observation scene graph that partially 
  represents the environment; there may be objects not in the current 
  observation. The plan can include objects that are not in the current 
  observation.
- When you want to manipulate objects, you need to move to the location 
  instead of the room.

Given the Initial state: initial_state
And the goal: goal_state
Output the plan directly.

Example:
Goal: Take a thick towel to a bed.
Formal Goal: (exists (?o ?l) (and (Is ?o towel) (thick ?o) 
             (Is ?l bed) (At ?o ?l)))
Output:
(navigate towelrack1)
(pick towel1)
(navigate bed1)
(place bed1)

[more examples]
'''
\end{lstlisting}

\subsubsection{Appliance Control Prompts}
For the microwave panel experiments, specialized prompts handle button effect hypotheses:

\begin{lstlisting}[language=Python, caption={Appliance Hypothesis Generation}, label=lst:appliance_hypothesis]
SYSTEM_PROMPT_FOR_APPLIANCE_HYPOTHESIS = '''
You are given a planning problem for controlling an appliance 
and need to extend button press effects.

Appliance description: Mode buttons change modes. 
Increase/decrease buttons adjust values BASED ON current mode.

IMPORTANT RULES:
1. Missing FUNCTIONS are adjusted by press-increase-button
2. Effect should be CONDITIONAL on current mode
3. Use: (when (MODE ?m) (increase (FUNCTION ?m) STEP))

Example:
{
  "id": "h1",
  "type": "action_effect", 
  "description": "Level button in set-waiting-time mode 
                  increases waiting time by 10 seconds",
  "content": {"press-level-button": 
    ["(when (set-waiting-time-mode ?m) 
           (increase (waiting-time ?m) 10))"]}
}
'''
\end{lstlisting}

\subsubsection{Goal Parsing Prompt}
Natural language goals are converted to PDDL goal specifications:

\begin{lstlisting}[language=Python, caption={Goal Parsing Prompt}, label=lst:goal_prompt]
SYSTEM_PROMPT_FOR_GOAL = '''
Convert natural language goal to PDDL goal format.

Input: Bring an ivory-white towel to the countertop
Output: (exists (?o ?l) (and (Is ?o towel) (ivory-white ?o) 
                              (Is ?l countertop) (At ?o ?l)))

Input: Place a toasted bread slice on a plastic coffee table  
Output: (exists (?o ?l) (and (Is ?o breadslice) (toasted ?o)
                              (Is ?l coffeetable) (plastic ?l) 
                              (At ?o ?l)))

Input: Make the kitchen light-on and take a pillow to bed
Output: (light-on kitchen1) 
        (exists (?o ?l) (and (Is ?o pillow) (Is ?l bed) 
                              (At ?o ?l)))
'''
\end{lstlisting}
\section{Insights, Limitations, and Future Work}

\subsection{Insights Highlight: Bridging Planning Reasoning and Open-World Knowledge from LLMs}

Our experiments demonstrate that \textbf{uncertainty-aware hypothesis expansion} provides a principled mechanism for bridging structured symbolic planning with the open-world knowledge embedded in large language models. This integration directly addresses a fundamental tension in robot autonomy: Task and Motion Planning (TAMP) systems are logically rigorous and constraint-consistent, yet brittle under model incompleteness, whereas LLMs are flexible and knowledge-rich, but prone to hallucination and unable to enforce strict logical or geometric constraints.
The central insight of this work is that LLMs should not be used as planners that replace formal reasoning, but rather as generators of \emph{model expansions}. These expansions are treated as uncertain hypotheses, whose validity is explicitly reasoned about and tested through planning and execution. In this way, the burden of uncertainty management is shifted from the language model to the planning system, which is better suited to reason about dependencies, preconditions, and action feasibility.

We find it useful to interpret this integration from three complementary perspectives.

\paragraph{Injecting compositional planning structure into unstructured open-world knowledge}
LLMs encode rich commonsense knowledge about objects, attributes, and action effects, but this knowledge is implicit and unstructured. By translating LLM-generated knowledge into object-centric hypotheses that participate in symbolic planning, we impose compositional structure on otherwise unstructured predictions. This allows the planner to reason about how hypothesized facts interact across time, actions, and constraints, rather than treating them as isolated suggestions. In practice, this significantly reduces failure modes such as premature grasping, unreachable placements, or resource conflicts, which are difficult to prevent with purely reactive or LLM-only planners.

\paragraph{Automated domain and problem specification with iterative refinement}
From a planning perspective, hypothesis generation can be viewed as an automated, task-driven process of domain and problem specification. Instead of requiring a complete symbolic model upfront, the robot incrementally proposes missing objects, predicates, or action effects only when they are needed to achieve the task. Verification actions then serve as a mechanism for grounding and correcting these specifications through interaction. This closes the loop between abstract model construction and embodied execution, enabling continual refinement of both domain knowledge and problem instances without manual intervention.

\paragraph{A Bayes-adaptive view of planning with open knowledge}
Conceptually, our framework aligns with a Bayes-adaptive interpretation of planning, in which uncertainty arises not only from partial observability of the environment, but also from uncertainty in the world model itself. Hypotheses represent latent variables over missing model components, and planning explicitly trades off task progress against information gathering. While our current implementation adopts a simplified uncertainty representation, the results suggest that even coarse uncertainty awareness is sufficient to induce qualitatively different and more robust behaviors, such as proactively verifying assumptions before committing to irreversible actions.

\subsection{Limitations and Equivalent Problems}

Despite its effectiveness, the proposed framework makes several simplifying assumptions that limit its generality and highlight important open challenges.

\paragraph{Choice of uncertainty representation and planning under uncertainty}
In the current system, hypotheses are generated as the most likely missing knowledge and represented using a ternary belief state (\texttt{true}, \texttt{false}, \texttt{unknown}). Verification actions are determinized optimistically, and negative outcomes are handled through replanning. While this design keeps planning tractable and enables real-time performance, it limits the robot’s ability to reason about varying degrees of risk or confidence. This choice is motivated by two practical considerations: (1) current foundation models do not reliably produce calibrated probability distributions over generated knowledge, making probabilistic modeling fragile; and (2) planning under richer uncertainty representations remains computationally challenging in long-horizon, integrated task-and-motion settings. As a result, the current system may behave overly optimistically in scenarios where risk-sensitive decision-making would be preferable.

\paragraph{Hypothesis verification conditions}
We assume that language models can propose verification conditions that are sufficient for determining the truth of a hypothesis, and that these conditions can be enforced by a formal planner. However, many real-world verification conditions depend on geometric details, long-horizon interactions, or complex perceptual reasoning that are difficult to infer from language alone. In such cases, determining whether a hypothesis is verifiable may itself constitute a planning or active perception problem. This limitation closely parallels challenges in active perception, where deciding \emph{how} to obtain information is as important as deciding \emph{what} information is missing. A principled integration of language-predicted symbolic conditions with geometry-aware and perception-driven reasoning remains an open problem.

\paragraph{Assumptions on hypothesis generation}
We further assume that the language model can generate relevant hypotheses within a finite number of attempts and translate them into a formal representation compatible with the planner. In this work, hypotheses are limited to relatively simple object-centric facts, such as existence, attributes, and localized action effects. It remains unclear whether current foundation models can reliably generate and maintain more complex hypotheses, such as abstract mechanisms, long-range causal dependencies, or the internal structure of unfamiliar devices. Scaling hypothesis generation to richer relational or procedural knowledge is, therefore, a significant open challenge.

\paragraph{Source of abstraction.}
Our current implementation requires the task instruction to define the abstraction of missing knowledge. This means that missing object concepts and predicates are explicitly encoded in the task goals. We make this assumption to ensure methodological completeness and to keep hypothesis generation task-oriented. In future work, we aim to explore failure reasoning modules \cite{liu2023reflect} as an additional source of abstraction, using execution feedback to identify missing concepts and formulate new hypotheses that are not directly specified by the task goals.

\subsection{Future Work}

Building on these limitations, we identify three promising directions for future research toward long-term autonomous robot deployment.

\paragraph{Richer uncertainty representations and planning algorithms}
A natural extension is to move beyond ternary hypothesis states toward probabilistic or belief-based representations, enabling risk-aware decision making and more nuanced trade-offs between exploration and exploitation. Integrating the proposed hypothesis framework with advanced planning under uncertainty algorithms, including belief-space task and motion planning, is a key direction.

\paragraph{Generalizing verification condition prediction}
Future work should investigate more general mechanisms for predicting and reasoning about verification conditions, combining language models with geometric reasoning, perception-driven affordance estimation, and active sensing strategies. This would allow the robot to autonomously determine not only which hypotheses to test, but also how to test them reliably.

\paragraph{Multi-source hypothesis generation}
Finally, hypothesis generation need not rely solely on language models. Incorporating additional sources, such as human queries, manuals, prior demonstrations, or long-term robot memory, could significantly improve robustness and reduce reliance on any single model. Developing principled methods for fusing and arbitrating between multiple hypothesis sources remains an important direction.

\subsection{Frequent Questions}

\noindent \textit{\textbf{Guarantees and Failures Beyond Assumptions.}} 
We consider a language model to be ``perfect'' if it can generate correct hypotheses within a finite number of attempts, propose verification conditions, and evaluate hypotheses based on observations, as discussed in Section IV.F. In principle, these capabilities allow the system to progressively reduce the gap in the underlying world model within a finite number of iterations.
However, some cases remain inherently unsolvable. Rather than introducing additional assumptions, we view such cases as failures arising from the design of the hypothesis representation and determinization process.
For example, verification may unexpectedly lead to a state from which the goal is no longer reachable, such as entering an unknown door that cannot be reopened from the other side.
Handling such risk-sensitive scenarios requires more sophisticated planning under uncertainty, which we leave for future work.

\noindent \textit{\textbf{Open-Ended Exploration in Open-World Settings.}} 
One possible strategy is to allow the LLM to terminate hypothesis generation when it judges a task to be impossible. We avoid this choice because it may lead to overly pessimistic early termination. Instead, our task scope assumes that the task is achievable in the underlying environment and focuses on optimistically uncovering the missing knowledge needed to solve it.
In practice, termination is bounded by the maximum number of hypothesis-generation attempts and replanning steps.
More broadly, proving impossibility in open-world settings is generally intractable, a challenge shared across the open-world planning literature.
Extending the framework to incorporate impossibility reasoning, such as \textit{belief-space pruning} or \textit{contradiction detection}, is an interesting direction for future work, which we will highlight in the revised version.

\noindent \textit{\textbf{Low-Level Geometric Challenges in Open Worlds.}} 
Our current framework focuses primarily on semantic hypotheses. An interesting direction for future work is to extend it to also reason about geometric hypotheses, potentially using video world models combined with physics-based verification.

\noindent \textit{\textbf{Runtime Complexity.}} 
We clarify that the hypothesis-driven approach does not introduce additional computational burden beyond the inherent complexity of the problem.
\textit{From a complexity perspective}, we decompose sampling from a high-dimensional plan distribution into sampling low-dimensional hypotheses and solving deterministic planning problems.
Let $m$ denote the number of unknown object-centric facts and $k$ the number of possible values for each fact. Direct planning under uncertainty has complexity
$
O\big((b \cdot k^m)^d\big) = O\big(b^d \cdot k^{md}\big),
$
where $b$ is the action branching factor and $d$ is the planning horizon.
In contrast, the hypothesis-driven approach samples hypotheses and solves a deterministic planning problem for each, yielding complexity
$
O\big(k^m \cdot b^d\big),
$
thereby avoiding the exponential coupling between uncertainty and planning depth.
\textit{From a practical perspective}, our approach offloads long-horizon, constraint-sensitive planning from the language model, reducing LLM queries by mitigating constraint violations and inconsistencies.
In the open-world Blocks World long-horizon setting, the hypothesis-driven approach reduces replanning triggers by approximately 38.5\% compared to pure LLM-based methods, while incurring comparable per-loop latency: 7.6s for hypothesis generation and update plus 0.25s for planning, compared with 7.0s for direct plan generation. This leads to improved overall efficiency.
\textit{From a real-world deployment perspective}, execution time dominates planning time. On real robots, planning accounts for only 7.6\% of the total task time on average.

\IEEEpeerreviewmaketitle





\end{document}